\documentclass{egpubl}
\usepackage{eg2019}

\ConferencePaper        
 \electronicVersion 

\ifpdf \usepackage[pdftex]{graphicx} \pdfcompresslevel=9
\else \usepackage[dvips]{graphicx} \fi

\PrintedOrElectronic

\usepackage{t1enc,dfadobe}

\usepackage{egweblnk}
\usepackage{cite}

\usepackage{amsmath}
\usepackage{lineno}
\usepackage{graphicx}
\usepackage{subcaption}
\usepackage{calc}
\usepackage{booktabs}
\usepackage{tabularx}
\usepackage[english]{babel}
\usepackage{multirow}
\usepackage{tikz}
\usetikzlibrary{shapes,arrows, fit, positioning}
\usepackage{makecell}

\usepackage{mathrsfs}
\usepackage{mathbbol}
\usepackage{bm}
\usepackage{wrapfig}
\usepackage{overpic}

\usepackage{color}
\definecolor{green}{rgb}{0.125,0.65,0.125}
\definecolor{reddish}{rgb}{0.7,0.,0.}
\definecolor{orange}{rgb}{1.0,0.3,0.1}
\definecolor{water}{rgb}{0.25,0.35,0.75}
\definecolor{fluorescentpink}{rgb}{1.0, 0.08, 0.58}
\definecolor{neongreen}{rgb}{0.22, 0.88, 0.08}
\definecolor{orange2}{rgb}{0.5,0.15,0.05}

\newcommand{\myrefeq}[1]{Eq.~\eqref{#1}}
\newcommand{\myreffig}[1]{Fig.~\ref{#1}}
\newcommand{\myreftab}[1]{Table~\ref{#1}}
\newcommand{\myrefsec}[1]{Sec.~\ref{#1}}

\newcommand{\myrefapp}[1]{App.~\ref{#1}}

\newcommand{\funcR}{\mathbb{R}}
\newcommand{\funcN}{\mathbb{N}}

\renewcommand{\vec}[1]{\mathbf{#1}}
\newcommand{\cs}{\vec{c}}
\newcommand{\ct}{\vec{d}}
\newcommand{\vel}{\vec{u}}

\newcommand{\newNils}[1]{#1}
\newcommand{\todoNils}[1]{}

\newcommand{\mytitleshort}{Latent Space Physics}
\newcommand{\mytitle}{Latent Space Physics: Towards Learning the Temporal Evolution of Fluid Flow}

\pdfoutput=1

\title[\mytitleshort]      {\mytitle}

\author[Steffen Wiewel, Moritz Becher, Nils Thuerey]
{
    \parbox{\textwidth}
    {
        \centering S. Wiewel$^{1}$,         M. Becher$^{1}$,         N. Thuerey$^{1}$
        \thanks{ This work was funded by the ERC Starting Grant {\em realFlow} (StG-2015-637014). }
    }
    \\
        {
        \parbox{\textwidth}
        {
            \centering $^1$Technical University of Munich\\
                    }
    }
}

\begin{document}

\teaser{
    \includegraphics[width=\linewidth, trim=0 0 0 100]{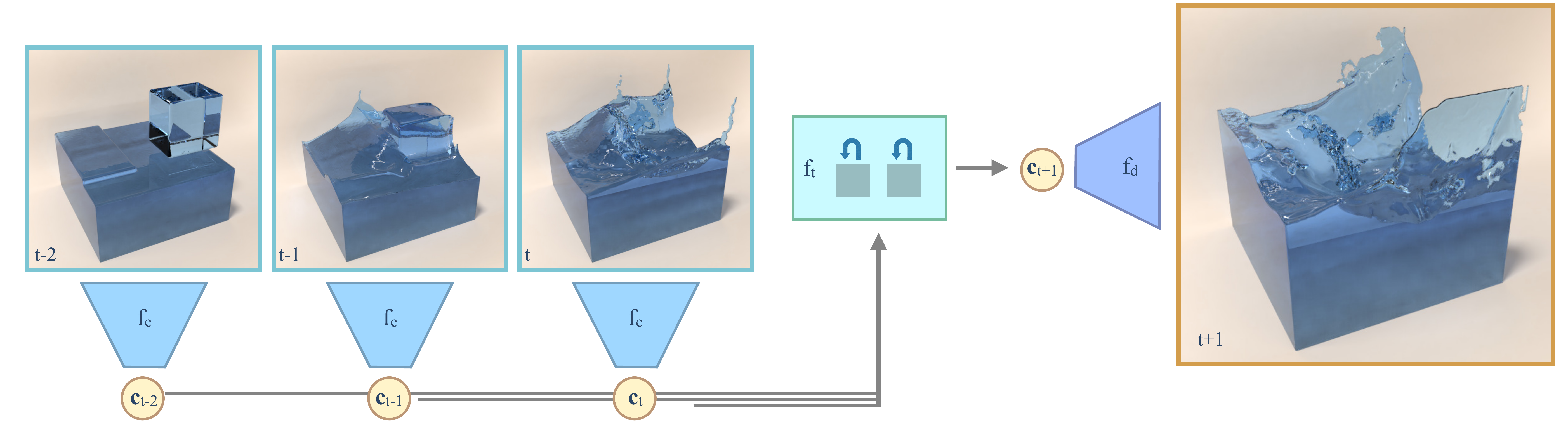}
    \centering
    \caption{Our method encodes multiple steps of a simulation field, typically pressure, into a reduced latent representation with a convolutional neural network. A
    second neural network with LSTM units then predicts the latent space code for one or more future time steps, yielding large reductions in runtime compared
    to regular solvers.}
    \label{fig:teaser}
}

\maketitle
\begin{abstract}
    We propose a method for the data-driven inference of temporal evolutions 
    of physical functions with deep learning. More specifically, we target fluid flows, i.e. Navier-Stokes problems,
    and we propose a novel LSTM-based approach to predict the changes of pressure fields over time.
    The central challenge in this context is the high dimensionality of Eulerian space-time data sets.
    We demonstrate for the first time that dense 3D+time functions of physics system can be predicted
    within the latent spaces of neural networks, and we arrive at a neural-network based simulation
    algorithm with significant practical speed-ups. We highlight the capabilities of our method with
    a series of complex liquid simulations, and with a set of single-phase buoyancy simulations.
    With a set of trained networks, our method is more than two orders of magnitudes faster than a
    traditional pressure solver. Additionally, we present and discuss a series of detailed evaluations
    for the different components of our algorithm.\\

    
\end{abstract}

\section{Introduction} \label{sec:intro}

The variables we use to describe real world physical systems typically 
take the form of complex functions with high dimensionality. 
Especially for transient numerical simulations, we usually employ
continuous models to describe how these functions evolve over time.
For such models, the field of computational methods has been highly
successful at developing powerful numerical algorithms
that accurately and efficiently predict how the 
natural phenomena under consideration will behave.
In the following, we take a different view on this problem: 
instead of relying on analytic expressions, we
use a deep learning approach to 
infer physical functions based on data. More specifically, we will focus on the temporal evolution 
of complex functions that arise in the context of fluid flows. Fluids encompass
a large and important class of materials in human environments, and as such
they're particularly interesting candidates for learning models.

While other works have demonstrated that machine learning methods
are highly competitive alternatives to traditional methods, e.g., for 
computing local interactions of particle based liquids~\cite{ladicky2015data},
to perform divergence free projections for a single point in time~\cite{tompson2016accelerating},
or for adversarial training of high resolution flows~\cite{xie2018tempogan},
few works exist that target temporal evolutions of physical systems.
While first works have considered predictions of Lagrangian objects
such as rigid bodies~\cite{watters2017visual}, and control of two dimensional 
interactions~\cite{ma2018fluid},
the question whether neural networks (NNs) can predict the evolution
of complex three-dimensional functions such as pressure fields of fluids
has not previously been addressed.
We believe that this is a particularly interesting challenge, as it not only
can lead to faster forward simulations, as we will demonstrate below, but also
could be useful for giving NNs predictive capabilities for complex inverse problems.

The complexity of nature at human scales makes it necessary to 
finely discretize both space and time for traditional numerical 
methods, in turn leading to a large number of degrees of freedom. 
Key to our method is reducing the dimensionality of the
problem using {\em  convolutional neural networks} (CNNs)
with respect to both time and space. Our
method first learns to map the original, three-dimensional problem
into a much smaller spatial {\em latent space}, at the same time
learning the inverse mapping. We then train a 
second network that maps a collection of reduced representations
into an encoding of the temporal evolution. This reduced
temporal state is then used to output a sequence of spatial
latent space representations, which are decoded to yield
the full spatial data set for a point in time. A key advantage of CNNs in this
context is that they give us a natural way to compute accurate
and highly efficient non-linear representations. We will later on 
demonstrate that the setup for computing this reduced representation
strongly influences how well the time network can predict
changes over time, and we will demonstrate the generality of
our approach with several liquid and single-phase problems.
The specific contributions of this work are: 
\begin{itemize}
\item a first LSTM architecture to predict temporal evolutions of dense, physical 3D functions in learned latent spaces,
\item an efficient encoder and decoder architecture, which by means of a strong compression, yields a very fast simulation algorithm,
\item in addition to a detailed evaluation of training modalities. \end{itemize}

\section{Related Work and Background}

Despite being a research topic for a long time \cite{rumelhart1988learning},
the interest in neural network algorithms is a relatively
new phenomenon, triggered by seminal works such as {\em ImageNet} \cite{krizhevsky2012imagenet}.
In computer graphics, such approaches have led to impressive results, e.g.,
for synthesizing novel viewpoints of natural scenes \cite{flynn2016deepstereo},
to generate photorealistic face textures \cite{saito2016dlfaces},
and to robustly transfer image styles between photographs \cite{luan2017deep},
to name just a few examples.
The underlying optimization 
approximates an unknown function $f^*(x)=y$, by minimizing 
an associated loss function $L$ 
such that $f(x,\theta) \approx y$.
Here, $\theta$ denotes the degrees of freedom of the chosen representation
for $f$. For our algorithm, we will consider deep neural networks. With the right choice of $L$, e.g.,
an $L_2$ norm in the simplest case, such a  neural network will approximate the original function $f^*$ as closely
as possible given its internal structure. A single layer $l$ of an NN
can be written as $a^{l} = \sigma( W_l a^{l-1} + b_l )$, where $a^{i}$ is the output of the i'th
layer, $\sigma$ represents an activation function, and $W_l,b_l$ denote weight matrix and bias, respectively.
 In the following, we collect the weights $W_l,b_l$ for all layers $l$ in $\theta$.

The latent spaces of generative NNs were shown to be powerful tools in image processing and synthesis
\cite{RadfordMC15,wu2016learning}.
They provide a non-linear representation that is closely tied to the given data distribution, and
our approach leverages such a latent space to predict the evolution of dense physical functions.
While others have demonstrated that trained feature spaces likewise pose
very suitable environments for high-level operations with impressive results 
\cite{upchurch2016deep}, we will focus on latent spaces of autoencoder networks in the following.
The sequence-to-sequence methods which we will use for time prediction
have so far predominantly found applications
in the area of natural language processing, e.g.,
 for tasks such as machine translation \cite{sutskever2014seqtoseq}.
These recurrent networks are especially popular for 
control tasks in reinforcement learning environments \cite{mnih2016asynchronous}.
Recently, impressive results were also achieved for tasks such as automatic 
video captioning \cite{xu2017learning}.

Using neural networks in the context of physics problems is a new area within the field of deep learning
methods. Several works have targeted predictions of Lagrangian objects based on image data.
E.g., Battaglia et al. \shortcite{battaglia2016interaction} predict
two-dimensional physics, a method that was subsequently extended to videos \cite{watters2017visual}.
Another line of work proposed a specialized architecture for two-dimensional
rigid bodies physics \cite{chang2016compositional},
while others have targeted predictions of liquid motions for robotic control \cite{schenck2017reasoning}
Farimani et al. \cite{farimani2017} proposed an adversarial training approach to infer solutions
for two-dimensional physics problems, such as heat diffusion and lid driven cavity flows. 
Other researchers have proposed networks to learn PDEs \cite{long2017pde} by encoding
the unknown differential operators with convolutions. 
To model the evolution of a subsurface multiphase flow others are using proper orthogonal decomposition in combination with recurrent neural networks (RNNs)~\cite{DBLP:journals/corr/abs-1810-10422}.
In the field of weather prediction, short-term prediction architectures evolved that internally also make use of RNN layers~\cite{DBLP:journals/corr/ShiCWYWW15}.
Other works create velocity field predictions by learning parameters for Reynolds Averaged Navier Stokes models from simulation data~\cite{osti_1333570}.
In addition, learned Koopman operators~\cite{morton2018deep} were proposed for representing temporal changes, whereas Lusch et al. are searching for representations of Koopman eigenfunctions to globally linearize dynamical systems using deep learning~\cite{Lusch2018DeepLF}.
While most of these works share our goal to infer
Eulerian functions for physical models, they are limited to relatively simple, two dimensional problems.
In contrast, we will demonstrate that our reduced latent space representation can work with
complex functions with up to several million degrees of freedom.

We focus on flow physics, for which we employ the well established
{\em Navier-Stokes} (NS) model. Its incompressible form is given by
\begin{equation}
	\label{eq:navierstokes}
	\partial\vel/\partial{t}  + \vel \cdot \nabla \vel =
	- 1/\rho \nabla{p} + \nu\nabla^2\vel + \vec{g}
	\ , \ \  \nabla\cdot\vel = 0 ,
\end{equation}
where the most important quantities are flow velocity $\vec{u}$ and pressure $p$. 
The other parameters $\rho,\nu,\vec{g}$ denote density, kinematic viscosity and external forces, respectively.
For liquids, we will assume that a signed-distance function $\phi$ is either advected with the flow, or reconstructed
from a set of advected particles.

In the area of visual effects, Kass and Miller were the first to employ height field models \cite{cg:kass:1990},
while Foster and Metaxas employed a first three-dimensional NS solver \cite{Foster:1996:RAL}.
After Jos Stam proposed an unconditionally stable advection and time integration scheme \shortcite{stam1999}, 
it led to powerful liquid solvers based on the particle levelset \cite{Foster01Practical}, 
in conjunction with accurate free surface boundary conditions \cite{enright03}.
Since then, the {\em fluid implicit particle} (FLIP) method has been especially
popular for detailed liquid simulations \cite{Zhu2005}, and we will use it to generate our training data.
Solvers based on these algorithms have subsequently been extended 
with accurate boundary handling \cite{Batty2007} synthetic turbulence \cite{kim:2008:wlt},
or narrow band algorithms \cite{ferstl2016narrow}, to name just a few examples.
A good overview of fluid simulations for computer animation can be found in the 
book by R. Bridson \shortcite{bridson2015}.
Aiming for more complex systems, other works have targeted
coupled reduced order models, or sub-grid coupling effects \cite{teng2016eulerian,fei2017multi}.
While we do not target coupled fluid solvers in our work,
these directions of course represent interesting future topics.

Beyond these primarily grid-based techniques, {\em smoothed particle hydrodynamics} (SPH)
are a popular Lagrangian alternative \cite{muller2003particle,macklin2014unified}. However, we will
focus on Eulerian solvers in the following, as CNNs are particularly amenable to grid-based
discretizations.
The pressure function has received special attention, as the underlying iterative solver
is often the most expensive component in an algorithm. E.g., techniques for dimensionality
reduction \cite{lentine2010novel,ando2015dimension}, and fast solvers \cite{McAdams:2010:PMP,ihmsen2014iisph}
have been proposed to diminish its runtime impact.

In the context of fluid simulations and machine learning for animation, a regression forest based
approach for learning SPH interactions has been proposed by Ladicky et al. \shortcite{ladicky2015data}.
Other graphics works have targeted learning flow descriptors with CNNs \cite{chu2017cnnpatch}, or
learning the statistics of splash formation \cite{um2017mlflip}, and
two-dimensional control problems \cite{ma2018fluid}.
While pressure projection algorithms with CNNs \cite{tompson2016accelerating,yang2016data}
shares similarities with our work on first sight, they are largely orthogonal. 
Instead of targeting divergence freeness
for a single instance in time, our work aims for learning its temporal evolution over the course of many
time steps. An important difference
is also that the CNN-based projection so far has only been demonstrated for smoke simulations,
similar to other model-reduced simulation approaches \cite{Treuille:2006:MRF}. For all of these methods,
handling the strongly varying free surface boundary conditions remains an open challenge, and we demonstrate
below that our approach works especially well for liquids.

\section{Method}

The central goal of our method is to predict future states of a physical function $\vec{x}$. While previous works often consider
low dimensional Lagrangian states such as center of mass positions, $\vec{x}$ takes the form of a dense Eulerian function in 
our setting. E.g., it can represent a spatio-temporal pressure function, or a velocity field. Thus, we consider 
$\vec{x}: \funcR^{3} \times \funcR^{+} \rightarrow \funcR^{d}$,
with $d=1$ for scalar functions such as pressure, and $d=3$ for vectorial functions such as velocity fields. As a consequence,
$\vec{x}$ has very high dimensionality when discretized, typically on the order of millions of spatial degrees of freedom.

Given a set of parameters $\theta$ and a functional representation $f_t$
our goal is to predict the $o$ future states $\vec{x}(t+h)$ to $\vec{x}(t+oh)$ as closely as possible given a current state and a series of $n$ previous states, i.e.,
\begin{equation} \label{eq:ft_def}
	f_t \big( \vec{x}(t-nh),...,\vec{x}(t-h), \vec{x}(t) \big) \approx \big[\vec{x}(t+h), \dots, \vec{x}(t+oh)\big].
\end{equation}
To provide better visual clarity the set of weights, i.e. learnable parameters, $\theta$ is omitted in the function definitions.

Due to the high dimensionality of $\vec{x}$, directly solving \myrefeq{eq:ft_def} would be very costly. Thus, we employ two additional functions $f_{d}$ and $f_{e}$, that compute a low dimensional encoding. The encoder $f_{e}$ maps into an $m_s$ dimensional space $\cs \in \funcR^{m_s}$ with $f_{e}(\vec{x}(t))=\cs^t$, whereas the decoding function $f_d$ reverses the mapping with $f_{d}(\cs^t)=\vec{x}(t)$.
Thus, $f_{d}$ and $f_{e}$ here represent spatial decoder and encoder models, respectively. 
Given such an en- and decoder, we rephrase the problem above as
\begin{equation}
	\begin{split}
		\tilde{f_t} \big( f_{e}( \vec{x}(t-nh) ),..., f_{e}( \vec{x}(t)) \big) &\approx \big[\cs^{t+h}, \dots, \cs^{t+oh}\big] \\
		f_{d} \big( \big[\cs^{t+h}, \dots, \cs^{t+oh}\big] \big) &\approx \big[\vec{x}(t+h), \dots, \vec{x}(t+oh)\big] \\
		f_{d}\big( \tilde{f_t}(f_{e}( \vec{x}(t-nh) ),..., f_{e}( \vec{x}(t))) \big) &\approx f_t( \vec{x}(t-nh),..., \vec{x}(t) )
	\end{split}
\end{equation}

We will use CNNs for $f_{d}$ and $f_{e}$, and thus the space $\cs$ is given
by their learned {\em latent space}. We choose its dimensionality $m_s$ such that the temporal
prediction problem above becomes feasible for dense three dimensional samples.

Our prediction network that models the function $\tilde{f_t}$ will likewise employ an encoder-decoder structure, and turns the temporal stack of encoded data points $f_{e}(\vec{x})$ into a reduced representation $\ct$, which we will refer to as {\em temporal context} below. 
The architectures of the spatial and temporal networks are described in the following sections \myrefsec{sec:spacenet} and \myrefsec{sec:timenet}, respectively.

\subsection{Reducing Dimensionality}
\label{sec:spacenet}

In order to reduce the spatial dimensionality of our inference problem, we employ a fully convolutional autoencoder architecture \cite{masci2011stacked}. 
Our autoencoder (AE) consists of the aforementioned encoding and decoding functions $f_e,f_d$ and is trained to reconstruct the quantity $\vec{x}$ as accurately as possible w.r.t. an $L_2$ norm, i.e. 
\begin{equation}
	\underset{\theta_d, \theta_e}{\text{min}} | f_{d}( f_{e}( \vec{x}(t) )) - \vec{x}(t) |_2,
\end{equation}
where $\theta_d, \theta_e$ represent the parameters of the decoder and encoder, respectively.
We use a series of convolutional layers activated by leaky rectified linear units (LeakyReLU) \cite{maas2013rectifier} for encoder and decoder, with a bottleneck layer of dimensionality $m_s$. 
This layer yields the latent space encoding that we use to predict the temporal evolution.
Both encoder and decoder consist of 6 convolutional layers that increase / decrease the number of features by a factor of 2.
In total, this yields a reduction factor of 256.

\begin{figure}[h]
	\centering
	\includegraphics[width = 0.4\textwidth]{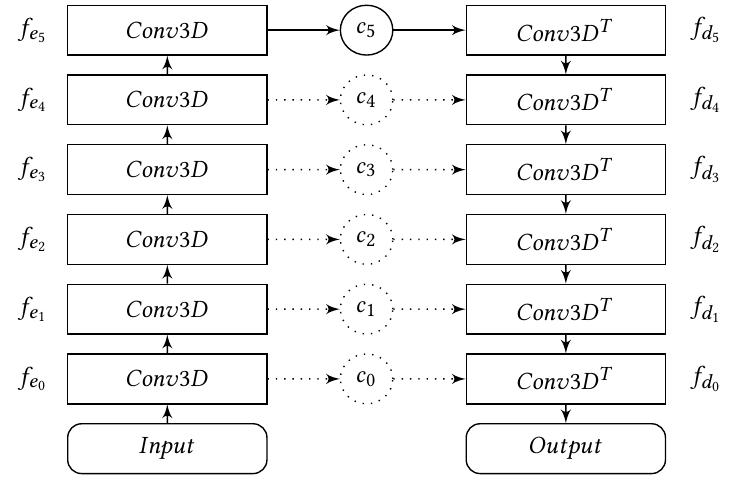}
	\medskip
	\captionsetup{belowskip=0.0em,aboveskip=0em}	
	\caption{\small Overview of the autoencoder network architecture}
	\caption*{\small See \myreftab{tab:autoencoder_architecture} for a detailed description of the layers. 
	Note that the paths from $f_{e_k}$ over $c_k$ to $f_{d_k}$ are only active in pretraining stage $k$. 
	After pretraining only the path $f_{e_5}$ over $c_5$ to $f_{d_5}$ remains active. 	} \label{fig:autoencoder_architecture}
\end{figure}

In the following, we will explain additional details of the autoencoder pre-training, and layer setup.
We denote layers in the encoder and decoder stack as $f_{e_i}$ and $f_{d_j}$, where $i, j \in [0, l]$, 
with $i,j$ being integers, denote the depth from the input and output layers, $l$ being the depth of the latent space layer.
In our network architecture, encoder and decoder layers with $i = j$ have to match, i.e., the output shape of $f_{e_i}$ has to be identical to that of $f_{d_j}$ and vice versa. 
This setup allows for a greedy, layer-wise pretraining of the autoencoder, as proposed by Bengio et al. \cite{bengio2007greedy}, where beginning from a shallow single layer deep autoencoder, additional layers are added to the model forming a series of deeper models for each stage.
The optimization problem of such a stacked autoencoder in pretraining is therefore formulated as
\begin{equation} \label{eq:aeMin2}
	\underset{\theta_{e_{0...k}}, \theta_{d_{0...k}}}{\text{min}} | f_{d_0} \circ f_{d_1} \circ ... \circ f_{d_k}( f_{e_k} \circ f_{e_{k-1}} \circ ... \circ f_{e_0}( \vec{x}(t) )) - \vec{x}(t) |_2 , 
\end{equation}
with $\theta_{e_{0...k}}, \theta_{d_{0...k}}$ denoting the parameters of the sub-stack for pretraining stage $k$, and $\circ$ denoting composition of functions.
For our final models, we typically use a depth $l=5$, and thus perform 6 runs of pretraining before training the complete model.
This is illustrated in \myreffig{fig:autoencoder_architecture}, where the paths from $f_{e_k}$ over $c_k$ to $f_{d_k}$ are only active in pretraining stage $k$. 
After pretraining only the path $f_{e_5}$ over $c_5$ to $f_{d_5}$ remains active.

\begin{table}[h!]
	\centering 
	\footnotesize
		\begin{tabular}{ c c c c l c c c}
			& Layer 	& Kernel 	& Stride 	& Activation 		& Output 			& Features\\ 
			\hline \hline
			& $Input$	& 			& 			&					& $\mathbf{r}/1$ 	& $d$\\
			& $f_{e_0}$ & 4 		& 2 		& Linear 			& $\mathbf{r}/2$ 	& $32$ \\
			& $f_{e_1}$ & 2 		& 2 		& LeakyReLU 		& $\mathbf{r}/4$ 	& $64$\\
			& $f_{e_2}$ & 2 		& 2 		& LeakyReLU 		& $\mathbf{r}/8$ 	& $128$\\
			& $f_{e_3}$ & 2 		& 2 		& LeakyReLU 		& $\mathbf{r}/16$ 	& $256$\\
			& $f_{e_4}$ & 2 		& 2 		& LeakyReLU 		& $\mathbf{r}/32$ 	& $512$\\
			& $f_{e_5}$ & 2 		& 2 		& LeakyReLU 		& $\mathbf{r}/64$ 	& $1024$\\ 
			\hline	 			 
			&  $c_5$ 	& 			& 			&					& $\mathbf{r}/64$ 	& $1024$\\ 
			\hline 
			& $f_{d_5}$ & 2 		& 2 		& LeakyReLU 		& $\mathbf{r}/32$ 	& $512$\\
			& $f_{d_4}$ & 2 		& 2 		& LeakyReLU 		& $\mathbf{r}/16$ 	& $256$\\
			& $f_{d_3}$ & 2 		& 2 		& LeakyReLU 		& $\mathbf{r}/8$	& $128$\\
			& $f_{d_2}$ & 2 		& 2 		& LeakyReLU 		& $\mathbf{r}/4$	& $64$\\
			& $f_{d_1}$ & 2 		& 2 		& LeakyReLU 		& $\mathbf{r}/2$	& $32$\\
			& $f_{d_0}$ & 4 		& 2 		& Linear 			& $\mathbf{r}/1$	& $d$\\ 
		\end{tabular}
		\captionsetup{belowskip=0.0em,aboveskip=0em}
	\vspace{0.3cm}
		\caption{ Parameters of the autoencoder layers. Here, $\mathbf{r} \in \funcN^3$ denotes the resolution of the data, and $d \in \funcN$ its dimensionality. 
			} \label{tab:autoencoder_architecture}
\end{table}

In addition, our autoencoder does not use any pooling layers, but instead only relies on strided convolutional layers. 
This means we apply convolutions with a stride of $s$, skipping $s-1$ entries when applying the convolutional kernel.
We assume the input is padded, and hence for $s=1$ the output size matches the input, while choosing a stride $s > 1$ results in a downsampled output \cite{odena2016deconv}.
Equivalently the decoder network employs strided transposed convolutions, where strided application increases the output dimensions by a factor of $s$. 
The details of the network architecture, with corresponding strides and kernel sizes can be found in \myreftab{tab:autoencoder_architecture}.

In addition to this basic architecture, we will also evaluate a {\em variational} autoencoder~\cite{rezende2014stochastic} in \myrefsec{sec:eval} that enforces a normalization on the latent space while keeping the presented AE layout identical.
As no pre-trained models for physics problems are available, we found greedy pre-training of the autoencoder stack to be crucial for a stable and feasible training process.

\subsection{Prediction of Future States}
\label{sec:timenet}

\begin{figure*}[hbt!] \centering
	\tikzstyle{io} = [rectangle, draw, text width=1em, rounded corners, text badly centered, inner sep=5pt, minimum height=2em, minimum width=2em]
\tikzstyle{layer} = [rectangle, draw, text width=1em, text badly centered, inner sep=0pt, minimum height=4em]
\tikzstyle{arrow} = [draw, -latex']
\tikzstyle{line} = [draw]	
\tikzstyle{description} = [text width=16em, text badly centered, inner sep=0pt, minimum height=2em, node distance=4em]
\begin{tikzpicture}[node distance=4em, auto]
        \node [io, text width=2em] (input) { $c^t$\\$c^{t-1}$\\$\cdots$\\$c^{t-n}$ };
        \node [io, node distance=5em,  right of=input] (enc_input) {$c$};
    \node [layer, color=orange!90, fill=orange!5, text width=4em, right of=enc_input] (enc_lstm_1) {LSTM};
    \node [io, node distance=4.5em, right of=enc_lstm_1] (enc_output) {$d$};
    \node [draw, dashed, inner xsep=1em, inner ysep=2em, fit=(enc_input) (enc_lstm_1) (enc_output)] (encoder) {};
    \node [description, node distance=3.5em, below of=encoder] (encoder_desc) {\emph{(n+1)-iterations}};
    \node [description, node distance=5em, above of=encoder] (encoder_title) {$\tilde{f_{t_e}}$};
    
        \node [io, node distance=4.5em, text width=1em, right of=enc_output] (context_input) {$d^n$};
        \node [io, node distance=8.0em, text width=2em, right of=context_input] (context_output) {$d^0$\\$d^1$\\$\cdots$\\$d^{o-1}$};
    \node [inner xsep=1em, inner ysep=2em, fit=(context_input)(context_output)] (context) {};
    \node [description, node distance=5em, above of=context] (context_title) {Context};

        \node [io, node distance=4.5em,  right of=context_output] (dec_input) {$d$};
    \node [layer, color=orange!90, fill=orange!5, text width=4em, right of=dec_input] (dec_lstm_1) {LSTM};
    \node [layer, color=green!90, fill=green!5, node distance=5.5em, text width=4em, right of=dec_lstm_1] (dec_fc_1) {Conv1D};
    \node [io, node distance=4em, right of=dec_fc_1] (dec_output) {$c$};	
    \node [draw, dashed, inner xsep=1em, inner ysep=2em, fit=(dec_input) (dec_lstm_1) (dec_fc_1) (dec_output)] (decoder) {};
                \node [description, node distance=3.5em, below of=decoder] (decoder_desc) {\emph{o-iterations}};
    \node [description, node distance=5em, above of=decoder] (decoder_title) {$\tilde{f_{t_d}}$ (Time Convolution)};
    
        \node [io, node distance=5em, text width=2em, right of=dec_output] (output) {$c^{t+1}$\\$c^{t+2}$\\$\cdots$\\$c^{t+o}$};
    
        \path [arrow, dotted] (input) -- (encoder);
    \path [arrow, dotted] (encoder) -- (context_input);
    \path [arrow] (context_input) -- (context_output) node [midway, above=0.0em] {$Repeat(o)$};
            \path [arrow, dotted] (context_output) -- (decoder);
    \path [arrow, dotted] (decoder) -- (output);
    
        \path [arrow] (enc_input) -- (enc_lstm_1);
    \path [arrow] (enc_lstm_1) -- (enc_output);
        \path [arrow] (enc_lstm_1) -- ++(2.5em, 0em) |- ++(0em, 3em) -| (enc_lstm_1);

        \path [arrow] (dec_input) -- (dec_lstm_1);
    \path [arrow] (dec_lstm_1) -- (dec_fc_1);
    \path [arrow] (dec_fc_1) -- (dec_output);
            \path [arrow] (dec_lstm_1) -- ++(2.5em, 0em) |- ++(0em, 3em) -| (dec_lstm_1);
    
\end{tikzpicture}
		\captionsetup{belowskip=0.0em,aboveskip=0em}
	\vspace{-0.3cm}
	\caption{Architecture overview of our LSTM prediction network. The dashed boxes indicate an iterative evaluation. Layer details can be found in the supplemental document. }
	\label{fig:seq-to-seq}
\end{figure*}
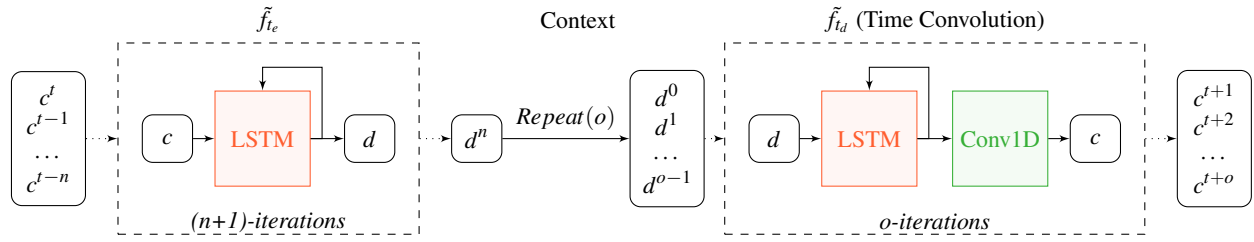

The prediction network, that models $\tilde{f_t}$, transforms a sequence of $n+1$ chronological, encoded input states $X = (\cs^{t-nh}, \dots, \cs^{t-h}, \cs^{t})$ 
into a consecutive list of $o$ predicted future states $Y = (\cs^{t+h}, \dots, \cs^{t+oh})$.
The minimization problem solved during training thus aims for minimizing
the mean absolute error between the $o$ predicted and ground truth states
with an $L_1$ norm:
\begin{equation} \label{eq:lstmMinMultiOutTS}
	\begin{gathered}
		\underset{\theta_t}{\text{min}} \big| \tilde{f_t} \big( \cs^{t-nh}, \dots, \cs^{t-h}, \cs^{t} \big) - \big[ \cs^{t+h}, \dots, \cs^{t+oh} \big] \big|_1  
		\ .
	\end{gathered}
\end{equation}
Here $\theta_t$ denotes the parameters of the prediction network, and $[\cdot,\cdot]$ denotes concatenation of the $\cs$ vectors.

In contrast to the spatial reduction network above, which receives the full spatial input at once and infers a latent space coordinate without any data internal to the network, the prediction network uses a recurrent architecture for predicting the evolution over time. 
It receives a series of inputs one by one, and computes its output iteratively with the help of an internal network state.
In contrast to the spatial reduction, which very heavily relies on convolutions, we cannot employ similar convolutions for the time data sets.
While it is a valid assumption that each entry of a latent space vector $\cs$ varies smoothly in time, the order of the entries is arbitrary and we cannot make any assumptions about local neighborhoods within $\cs$.
As such, convolving $\cs$ with a filter along the latent space entries typically does not give meaningful results.
Instead, our prediction network will use convolutions to translate the LSTM state into the latent space, in addition to fully connected layers of LSTM units.

The prediction network approximates the desired function $\tilde{f_t}$ with the help of an internal temporal context of dimension $m_t$, which we will denote as $\ct$.
Thus, the first part of our prediction network represents a recurrent encoder module, transforming $n+1$ latent space points into a temporal context $\ct$.
The time decoder module has a similar structure, and is likewise realized with layers of LSTM units.
This module takes a context $\ct$ as input, and outputs a series of future, spatial latent space representations.
By means of its internal state, the time decoder is trained to predict $o$ future states when receiving the same context $\ct$ repeatedly.  
We use {\em tanh} activations for all LSTM layers, and hard sigmoid functions for efficient, internal activations. 

Note that the iterative nature is shared by encoder and decoder module of the prediction network,
i.e., the encoder actually internally produces $n+1$ contexts, the first $n$ of which are intermediate contexts.
These intermediate contexts are only required for the feedback loop internal to the corresponding LSTM layer, and are discarded afterwards.
We only keep the very last context in order to pass it to the decoder part of the network.
This context is repeated $o$ times, in order for the decoder LSTM to infer the desired future states.

\begin{figure}[h]
	\centering
	\tikzstyle{io} = [rectangle, draw, text width=1em, rounded corners, text badly centered, inner sep=5pt, minimum height=2em, minimum width=2em]
\tikzstyle{layer} = [rectangle, draw, text width=1em, text badly centered, inner sep=0pt, minimum height=4em]
\tikzstyle{arrow} = [draw, -latex']
\tikzstyle{line} = [draw]	
\tikzstyle{description} = [text width=16em, text badly centered, inner sep=0pt, minimum height=2em, node distance=4em]
\begin{tikzpicture}[node distance=4em, auto]
    \node [io, node distance=4.5em, text width=2em] (context_output) {$d^0$\\$d^1$\\$...$\\$d^{o-1}$};

        \node [io, node distance=4.5em,  right of=context_output] (dec_input) {$d$};
    \node [layer, color=orange!90, fill=orange!5, text width=4em, right of=dec_input] (dec_lstm_1) {LSTM};
    \node [layer, color=orange!90, fill=orange!5, node distance=5.5em, text width=4em, right of=dec_lstm_1] (dec_lstm_2) {LSTM};
    \node [io, node distance=4.5em, right of=dec_lstm_2] (dec_output) {$c$};		
    \node [draw, dashed, inner xsep=1em, inner ysep=2em, fit=(dec_input) (dec_lstm_1) (dec_lstm_2) (dec_output)] (decoder) {};
    \node [description, node distance=3.5em, below of=decoder] (decoder_desc) {\emph{o-iterations}};
    \node [description, node distance=5em, above of=decoder] (decoder_title) {$\tilde{f_{t_d}}$ (Fully Recurrent)};
    
        \node [io, node distance=5em, text width=2em, right of=dec_output] (output) {$c^{t+1}$\\$c^{t+2}$\\$...$\\$c^{t+o}$};
    
        \path [arrow, dotted] (context_output) -- (decoder);
    \path [arrow, dotted] (decoder) -- (output);
    
        \path [arrow] (dec_input) -- (dec_lstm_1);
    \path [arrow] (dec_lstm_1) -- (dec_lstm_2);
    \path [arrow] (dec_lstm_2) -- (dec_output);
    \path [arrow] (dec_lstm_1) -- ++(2.5em, 0em) |- ++(0em, 3em) -| (dec_lstm_1);
    \path [arrow] (dec_lstm_2) -- ++(2.5em, 0em) |- ++(0em, 3em) -| (dec_lstm_2);

\end{tikzpicture}
	\captionsetup{belowskip=0.0em,aboveskip=0em}
	\caption{Alternative to the time convolution sequence to sequence decoder (fully recurrent)}
	\label{fig:fullyrecurrent}
\end{figure}
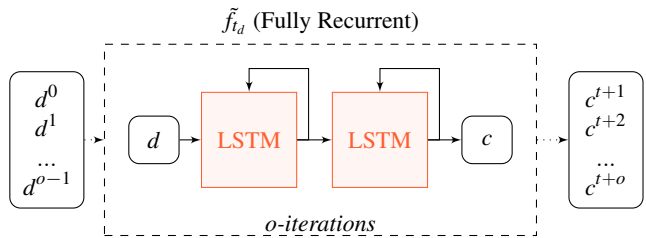

Despite the spatial dimensionality reduction with an autoencoder, the number of weights in LSTM layers can quickly grow due to their inherent internal feedback loops (typically equivalent to four fully connected layers). 
To prevent overfitting from exploding weight numbers in the LSTM layers, we propose a hybrid structure of LSTM units and convolutions as shown in \myreffig{fig:seq-to-seq} that is used instead of the fully recurrent approach presented in \myreffig{fig:fullyrecurrent}.

For our prediction network we use two LSTM layers that infer an internal temporal representation of the data, followed by a final linear, convolutional network that translates the temporal representation into the corresponding latent space point.
This convolution effectively represents a translation of the context information from the LSTM layer into latent space points that is constant for all output steps.
This architecture effectively prevents overfitting, and ensures a high quality temporal prediction, as we will demonstrate below.
In particular, we will show that this hybrid network outperforms networks purely based on LSTM layers, and significantly reduces the weight footprint.
Additionally the prediction network architecture can be extended by applying multiple stacked convolution layers after the final LSTM layer.
We found this hybrid architecture crucial for inferring the high-dimensional outputs of physical simulations.

While the autoencoder, thanks to its fully convolutional architecture, could be applied to inputs of varying size, the prediction network is trained for fixed latent space inputs, and internal context sizes.
Correspondingly, when the latent space size $m_s$ changes, it influences the size of the prediction network's layers. 
Hence, the prediction network has to be re-trained from scratch when the latent space size is changed. 
We have not found this critical in practice, because the prediction network takes significantly less time to train than the autoencoder, as we will discuss in \myrefsec{sec:eval}.

\begin{figure*}[t!]
	\centering \footnotesize
		\begin{overpic}[width = 0.999\textwidth]{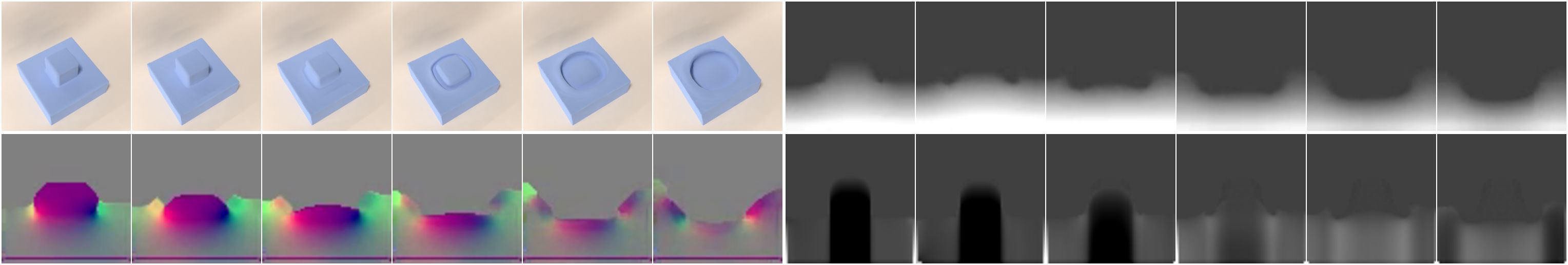}
			\put( 1.0, 15.2 ){\tiny \color{black}{Surface}} 
			\put( 1.0, 6.7 ){\tiny \color{white}{$\vec{u}$}} 
			\put( 51.0, 15.2 ){\tiny \color{white}{$p_t$}} 
			\put( 51.0, 6.7 ){\tiny \color{white}{$p_s$}} 
		\end{overpic}
	\caption{
	Example sequences of different quantities under consideration, all ground truth reference data (center slices).
	The surface shown top left illustrates the 3D configuration of the liquid, but is not used for inference.
	The three velocity components of $\vec{u}$ are shown as RGB, whereas the total pressure $p_t$ and the static part of the split pressure $p_s$ are shown as grayscale images.
	}
	\label{fig:data_fields_raw}
\end{figure*}
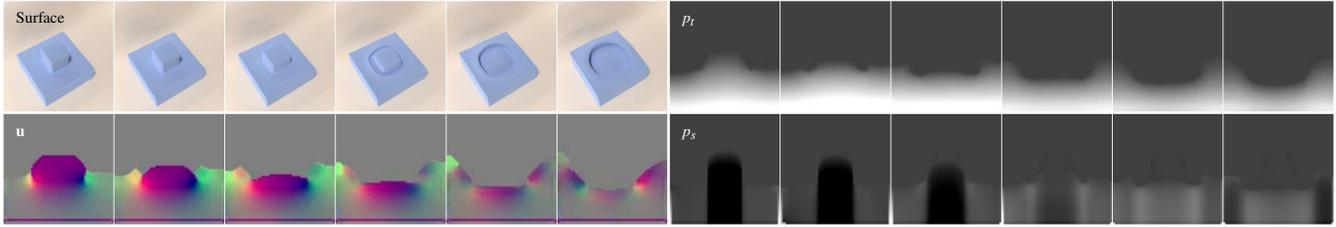

\section{Fluid Flow Data}\label{sec:fluiddata}

To generate fluid data sets for training we rely on a NS solver with operator splitting~\cite{bridson2015}
to calculate the training data at discrete points in space and time. On a high level,
the solver contains the following steps: computing motion of the fluid
with the help of transport, i.e. {\em advection} steps, for the velocity $\vec{u}$, evaluating external forces, and then computing
the harmonic pressure function $p$. 
In addition, a visible, passive quantity such as smoke density $\rho$, or a level-set representation $\phi$
for free surface flows is often advected in parallel to the velocity itself.
Calculating the pressure typically involves solving an elliptic second-order PDE,
and the gradient of the resulting pressure is used to make the flow divergence free. 

In addition, we consider a {\em split} pressure, which represents a residual quantity. 
Assuming a fluid at rest on a ground with height $z_g$,
a hydrostatic pressure $p_s$ for cell at height $z$,
can be calculated as
$p_s(z) = p(z_0) + \frac{1}{A} \int_z^{z_0}  \iint_A \vec{g} \rho(h) ~  \text{d}x  \text{d}y ~  \text{d}h$,
with $z_0$, $p_0$, $A$ denoting
surface height, surface pressure, and cell area, respectively. 
As density and gravity can be treated as constant in our setting, this further simplifies to 
$p_s = \rho \vec{g} (z - z_0)$.
While this form can be evaluated very efficiently, it has the drawback that it is only
valid for fluids in hydrostatic equilibrium, and typically cannot be used for dynamic simulations in 3D.
Given a data-driven method to predict pressure fields, we can incorporate the hydrostatic pressure 
into a 3D liquid simulation by decomposing the regular pressure field into hydrostatic and dynamic 
components $p_t = p_s + p_d$, such that our autoencoder separately receives and encodes the 
two fields $p_s$ and $p_d$.
To differentiate between the classic pressure field created by the simulation and our extracted split pressure components $p_s$ and $p_d$, the classic pressure field is called total pressure $p_t$ in the following.
With this split pressure, the autoencoder could potentially put more emphasis on the small scale 
fluctuations $p_d$ from the hydrostatic pressure gradient.

Overall,
these physical data sets differ significantly from data sets such as natural images that are 
targeted with other learning approaches. They are typically well structured, and less ambiguous due to a lack
of projections, which motivates our goal to use learned models.
At the same time they exhibit strong temporal changes, as is visible in \myreffig{fig:data_fields_raw}, which make the temporal inference problem a non-trivial task.

Depending on the choice of physical quantity to infer with our framework, 
different simulation algorithms emerge. We will focus on velocity $\vec{u}$ and the two pressure variants, total $p_t$ and split ($p_s$ and $p_d$) in the following. 
When targeting $\vec{u}$ with our method, this means that we can omit velocity advection as well as pressure solve,
while the inference of pressure means that we only omit the pressure solve,
but still need to perform advection and velocity correction with the pressure gradient.
While this pressure inference requires more computations, 
the pressure solve is typically the most time consuming part with a super-linear complexity, and as such
both options have comparable runtimes. 
When predicting the pressure field, we also use a boundary condition alignment step 
for the free surface \cite{ando2015dimension}.
It takes the form of three Jacobi iterations in a narrow band at the liquid surface
in order to align the Dirichlet boundary conditions with the current position of the interface. 
This step is important for liquids, as it incorporates small scale dynamics, leaving
the large-scale dynamics to a learned model.

\subsection{Interval Prediction}\label{sec:interval_pred}
A variant for both of these simulation algorithm classes is to only rely on the network prediction for a
limited time interval of $i_p$ time steps, and then perform a single full simulation
step without any network calculations,
i.e., for $i_p=0$ the network is not used at all, while $i_p=\infty$ is identical to
the full network prediction described in the previous paragraph.
We will investigate prediction intervals on the order of $4$ to $14$ steps.
This simulation variant represents a joint numerical time integration and network prediction, that can have advantages to prevent drift from the learned predictions.
We will denote such versions as {\em interval predictions} below.

\subsection{Data Sets}\label{sec:data_gen}

To demonstrate that our approach is applicable to a wide range of physics phenomena,
we will show results with three different 3D data sets in the following. 
To ensure a sufficient amount of variance with respect to 
physical motions and dynamics, we use randomized simulation setups.
We target scenes with high complexity, i.e., strong visible splashes and vortices, and large CFL (Courant-Friedrichs-Lewy) numbers (typically around 2-3),
that measure how fast information travels from cell to cell in a complex simulation domain.
For each of our data sets, we generate $n_{s}$ scenes of different initial conditions, for which we discard the first $n_{w}$ time steps, as these
typically contain small and regular, and hence less representative dynamics. 
Afterwards, we store a fixed number of $n_{t}$ time steps 
as training data, resulting in a final size
of $n_{s} n_{t}$ spatial data sets. Each data set content is normalized to the range of [-1,1].

\begin{table}[hbt]
	\centering
		\begin{tabular}{ c | c c c }
									& liquid64 	& liquid128 	& smoke128 	\\ \hline \hline
			Scenes $n_s$ 				& 4000 	 	& 800		  	& 800	 	\\
			Time steps $n_t$ 			& 100 		& 100		  	& 100 		\\
			\makecell{Size}		 	& 419.43GB 	& 671.09GB	  	& 671.09GB 	\\
			\makecell{Size, encoded}	& 1.64GB	& 2.62GB		& 2.62GB	
		\end{tabular} 
	\vspace{0.3cm}
		\caption{ List of the augmented data sets for the total pressure architecture. 
		Compression by a factor of 256 is achieved by the encoder part of the autoencoder $f_e$. }
		\label{tab:datasets}
\end{table}

Two of the three data sets contain liquids, 
while the additional one targets smoke simulations.
The liquid data sets with spatial resolutions of $64^{3}$ and $128^3$ contain randomized sloshing
waves and colliding bodies of liquid. The scene setup consists of a low basin, represented by a large volume of liquid at the bottom of the domain, and a tall but narrow pillar of liquid, that drops into it. Additionally a random amount, ranging from zero to three smaller liquid drops are placed randomly in the domain. For the $128^3$ data set we additionally include complex geometries for the initial liquid bodies, yielding a
larger range of behavior. These data sets will be denoted as {\em liquid64} and {\em liquid128}, respectively. 
In addition, we consider a data set containing single-phase flows with buoyant smoke
which we will denote as  {\em smoke128}. 
We place 4 to 10 inflow regions into an empty
domain at rest, and then simulate the resulting plumes of hot smoke. 
As all setups are invariant w.r.t. rotations around the axis of gravity (Y in our setups),
we augment the data sets by mirroring along XY and YZ.
This leads to sizes of the data sets from 80k to 400k entries, and the $128^3$ data sets have a total
size of 671GB.
Rendered examples from all data sets can be found in \myreffig{fig:dataset_renderings}, \myreffig{fig:dataset_liq128initial} and \myreffig{fig:dataset_smoke} in the supplemental document, as well as further information about the initial conditions and physical parameters of the fluids.

\section{Evaluation and Training}\label{sec:eval}

\newcommand{\myq}{0.195\textwidth}
\begin{figure*}
	\centering 
	\footnotesize 
		\begin{overpic}[width = \myq]{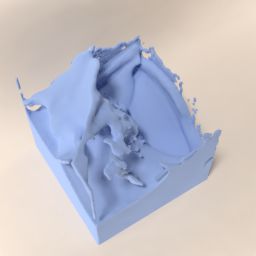} \put( 6, 90 ){\scriptsize \color{black}{Reference (GT)}} \end{overpic}\ 
		\begin{overpic}[width = \myq]{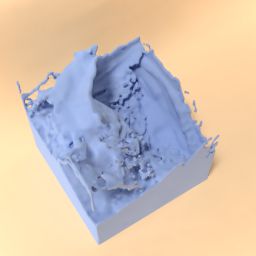} \put( 6, 90 ){\scriptsize \color{black}{$p_t$}} \end{overpic}\ 
		\begin{overpic}[width = \myq]{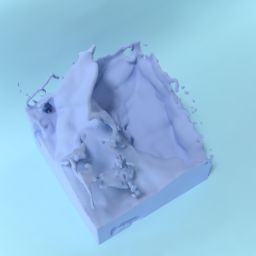} \put( 6, 90 ){\scriptsize \color{black}{$p_s$ and $p_d$}} \end{overpic}\ 
		\begin{overpic}[width = \myq]{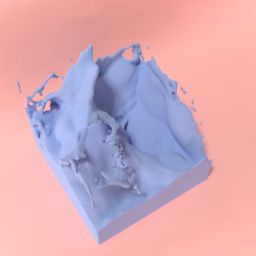} \put( 6, 90 ){\scriptsize \color{black}{VAE $p_s$ and $p_d$}} \end{overpic}\ 
		\begin{overpic}[width = \myq]{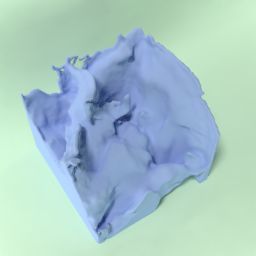} \put( 6, 90 ){\scriptsize \color{black}{$\vec{u}$}} \end{overpic}\ 
	
	\caption{
		Comparison of free surface renderings of simulations driven by a classic fluid simulation working with the different fields that are directly en- and then decoded, i.e. compressed, with a trained autoencoder. Time prediction is not used in this example, thus only the performance of the individual autoencoders for each quantity is evaluated.
		The velocity significantly differs from the reference (i.e., ground truth data, left), while all three pressure variants fare well
		with average PSNRs of 64.81, 64.55, and 62.45 (f.l.t.r.).}
	\label{fig:encdec_renderings}
\end{figure*}

In the following we will evaluate the different options discussed in the previous section with respect to their prediction accuracies. 
In terms of evaluation metrics, we will use PSNR (peak signal-to-noise ratio) as a baseline metric,
in addition to a surface-based Hausdorff distance in order to more accurately compare the position of the liquid interface~\cite{huttenlocher1993comparing,li2015database}.
More specifically, given two signed distance functions $\phi_{r}, \phi_{p}$ representing reference and 
predicted surfaces,	we compute the surface error as	
\begin{equation}
	e_h = \text{max}( 1/|S_p| \sum_{\vec{p}_1 \in S_p} \phi_r(\vec{p}_1), 1/|S_r| \sum_{\vec{p}_2 \in S_r} \phi_p(\vec{p}_2)) / \Delta x.
\end{equation}
Unless otherwise noted, the error measurements start after 50 steps of simulation, and are averaged 
for ten test scenes from the  {\em liquid64} setup.

\subsection{Spatial Encoding}
We first evaluate the accuracy of only the spatial encoding, i.e., the autoencoder network in conjunction with a numerical time integration scheme.
At the end of a fluid solving time step, we encode the physical variable $\vec{x}$ under consideration with $\cs = f_e(\vec{x})$,
and then restore it from its latent space representation  $\vec{x}' = f_d(\cs)$. 
In the following, we will compare flow velocity $\vel$, total pressure $p_t$, and split pressure ($p_s$, $p_d$), all with a latent space size of $m_s=1024$.
We train a new autoencoder for each quantity, and we additionally consider a variational autoencoder for the split pressure. 
Training times for the autoencoders were two days on average, including pre-training.
To train the different autoencoders, we use 6 epochs of pretraining and 25 epochs of training using an Adam optimizer, with
a learning rate of 0.001 and a decay factor of 0.005. For training we used 80\% of the data set, 10\% for validation during training, and another 10\% for testing. 

\begin{figure*}[bt]
	\centering \scriptsize
	\newcommand{\myv}{0.22\textwidth}  	\renewcommand{\arraystretch}{0}
	\begin{tabular}{@{}c c c c@{}}
		\subcaptionbox{\scriptsize AE only, different phys. quantities\label{fig:encdec_hausdorff}}{\includegraphics[width = \myv]{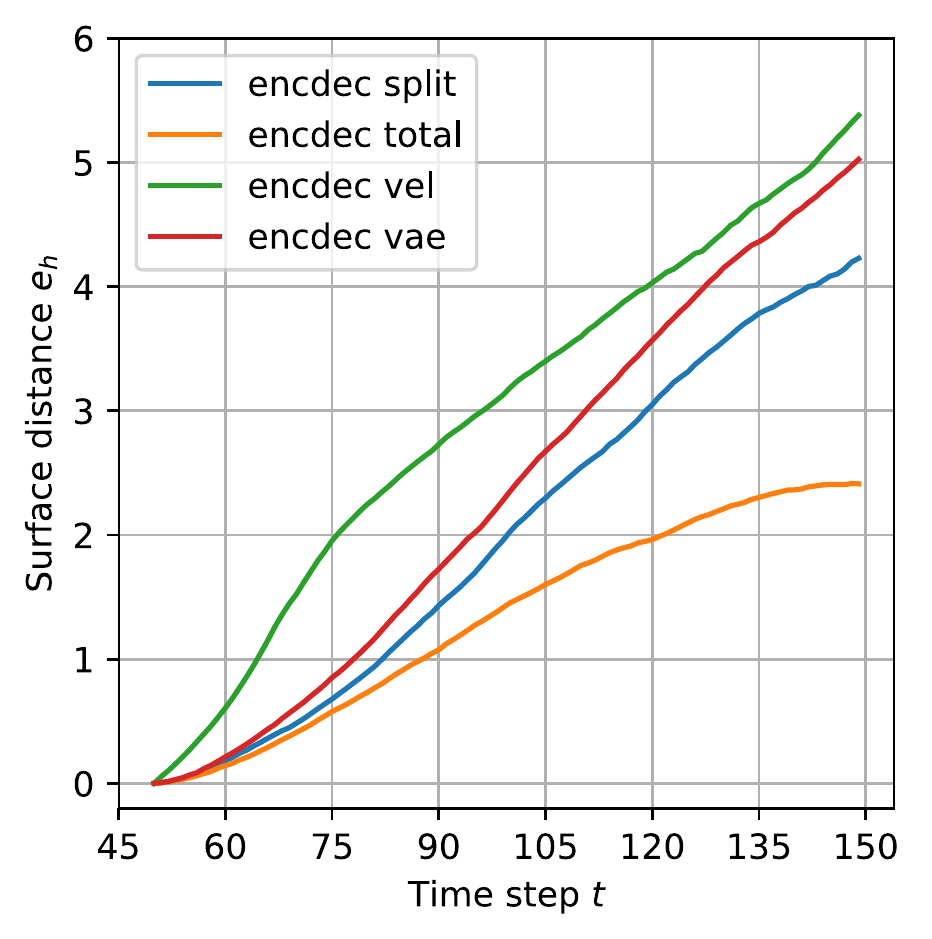}} &
				\subcaptionbox{\scriptsize Full alg., phys. quantities, $i_p=\infty$.\label{fig:basenetworkcomparisoni1000}}{\includegraphics[width = \myv]{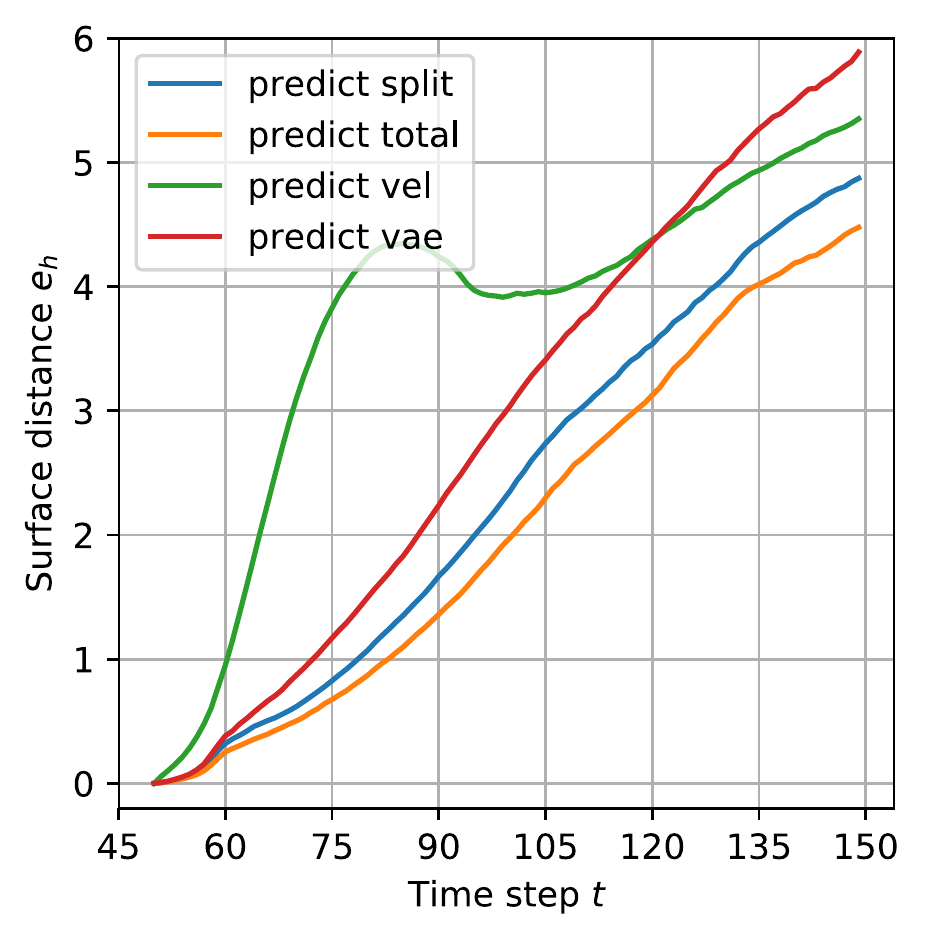}} &
				\subcaptionbox{\scriptsize Full alg., phys. quantities, $i_p=14$.\label{fig:basenetworkcomparisoni15}}{\includegraphics[width = \myv]{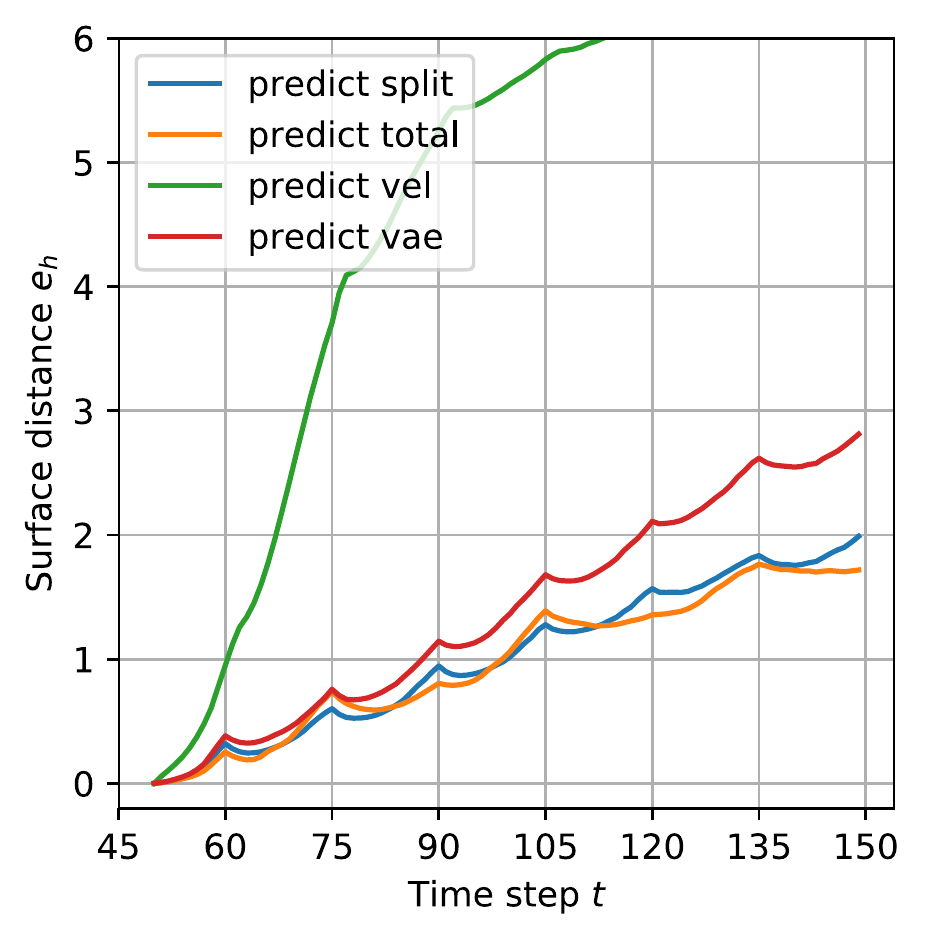}} &
				\subcaptionbox{\scriptsize Pred. intervals $i_p$ for $p_t$, full alg.\label{fig:intervals_comparison_levelset}}{\includegraphics[width = \myv]{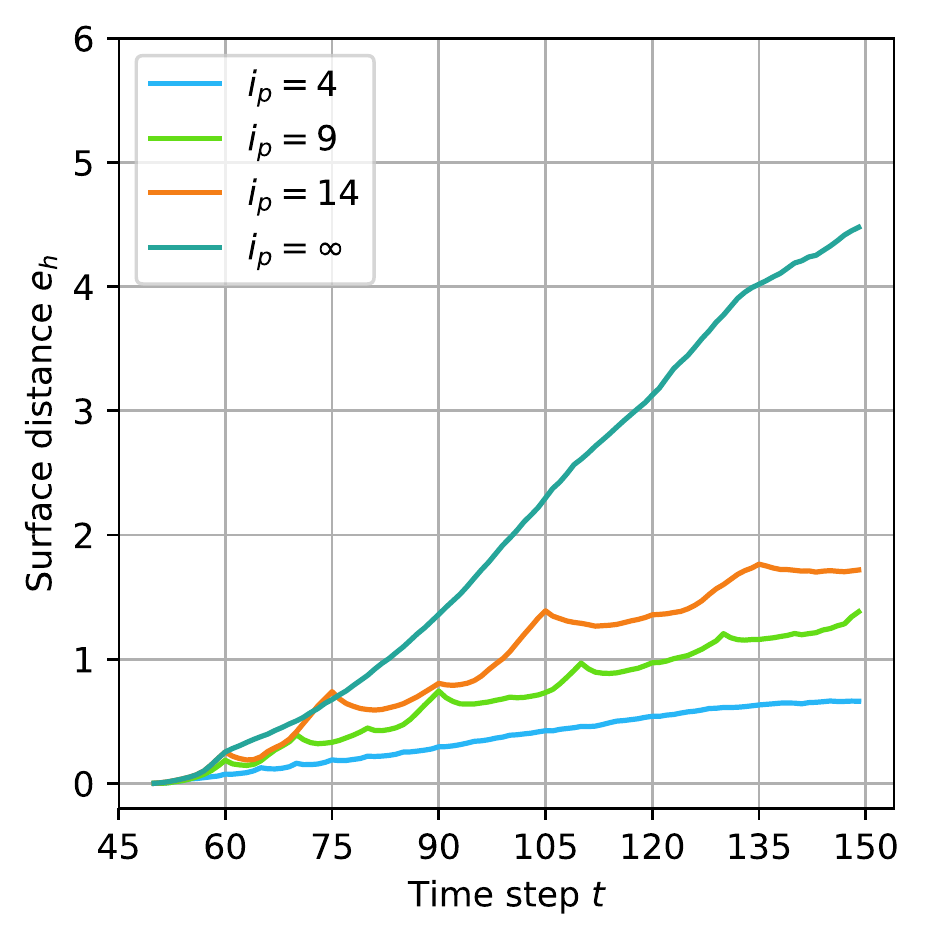}} 
\\
				\subcaptionbox{\scriptsize AE only, pressure PSNR values \label{fig:Mencdec_PSNR}}{\includegraphics[width = \myv]{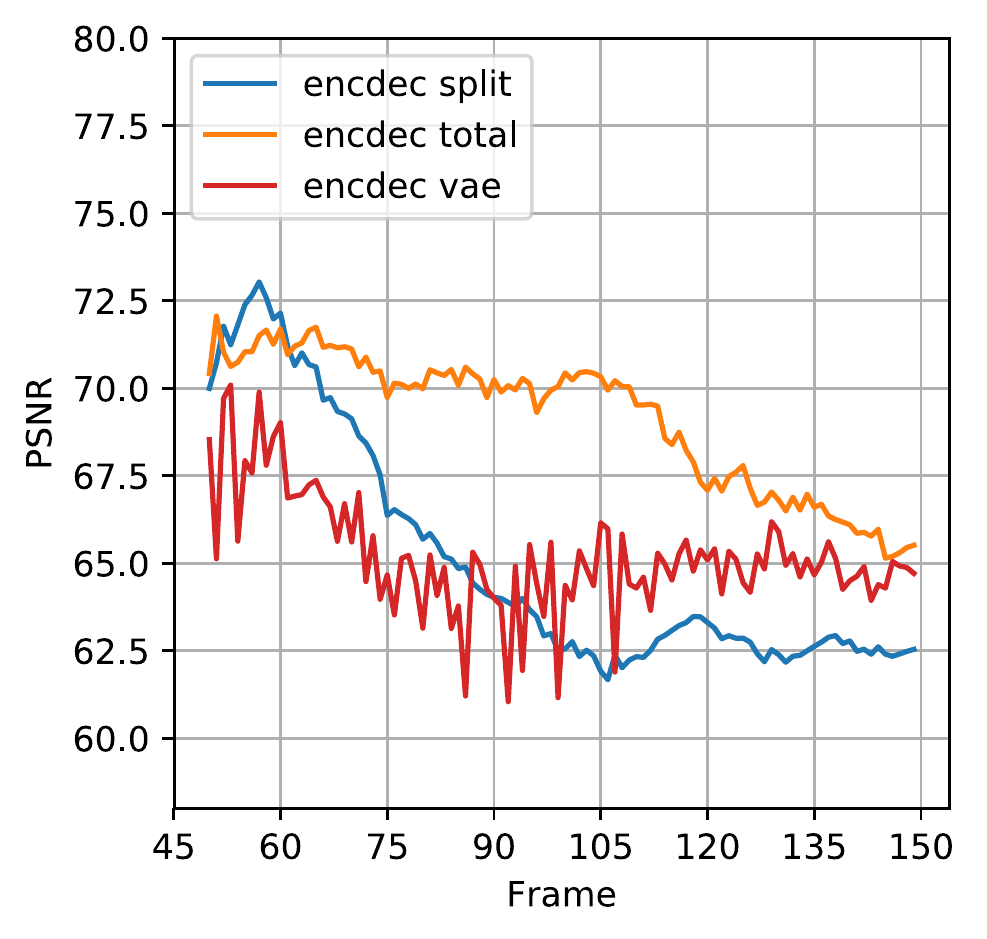}} &
				\subcaptionbox{\scriptsize Full alg, pressure PSNR, with $i_p=\infty$.\label{fig:Mbasenetworkcomparisoni1000_psnr}}{\includegraphics[width = \myv]{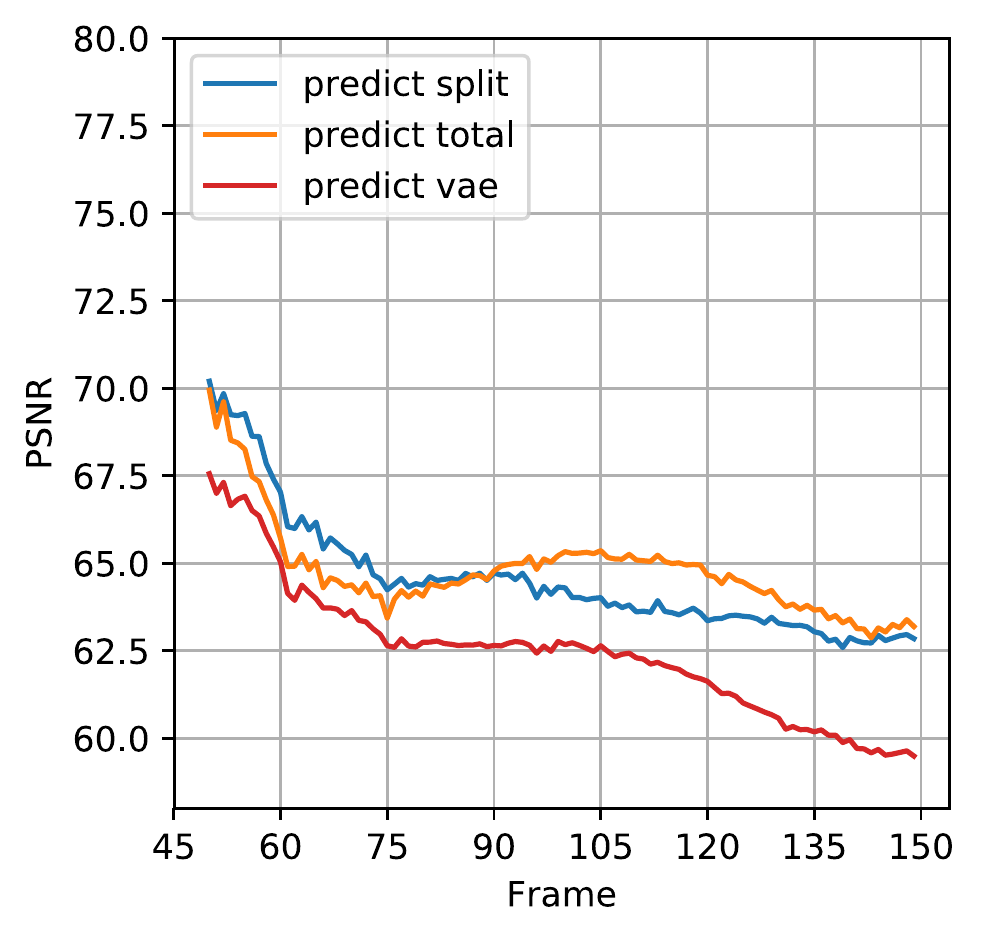}} &
				\subcaptionbox{\scriptsize Architecture variants, $p_t$ PSNR values $i_p=\infty$, $o=5$.\label{fig:Malgcomparisoni1000_psnr}}{\includegraphics[width = \myv]{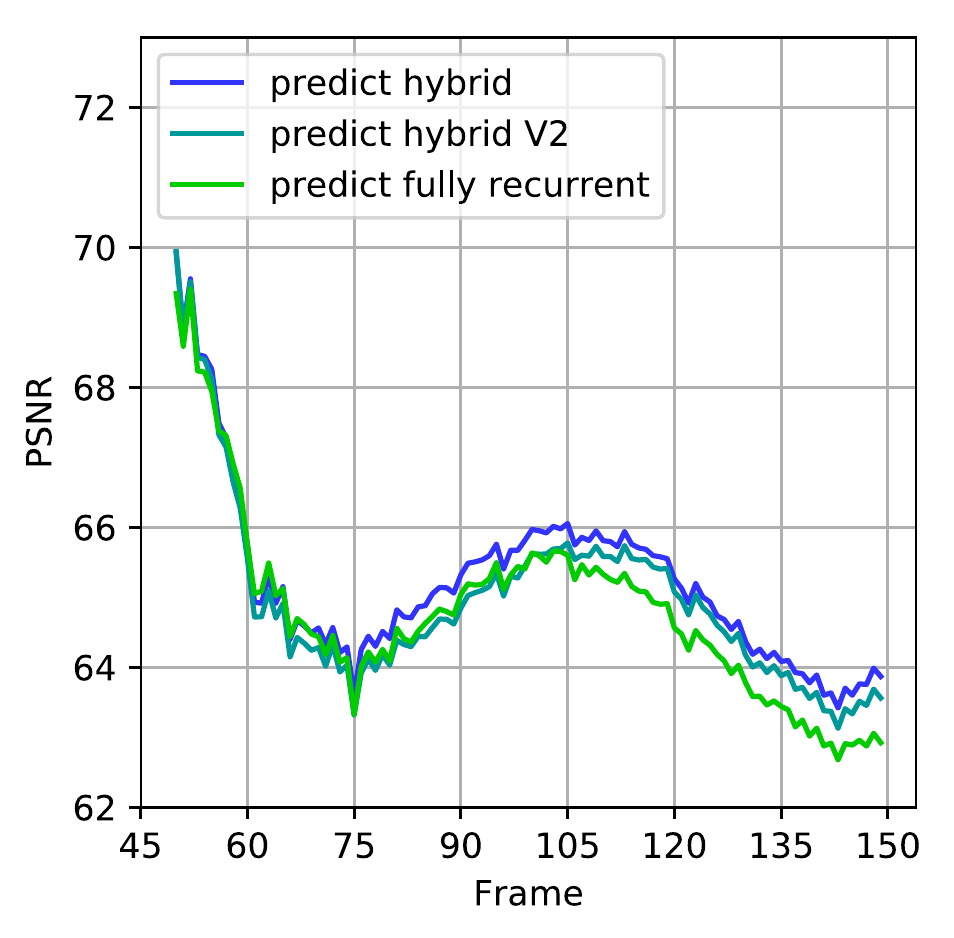}} &
				\subcaptionbox{\scriptsize Different output steps for $p_t$, varied $i_p$ \label{fig:intervals_comparison_levelset_multiout}}{\includegraphics[width = \myv]{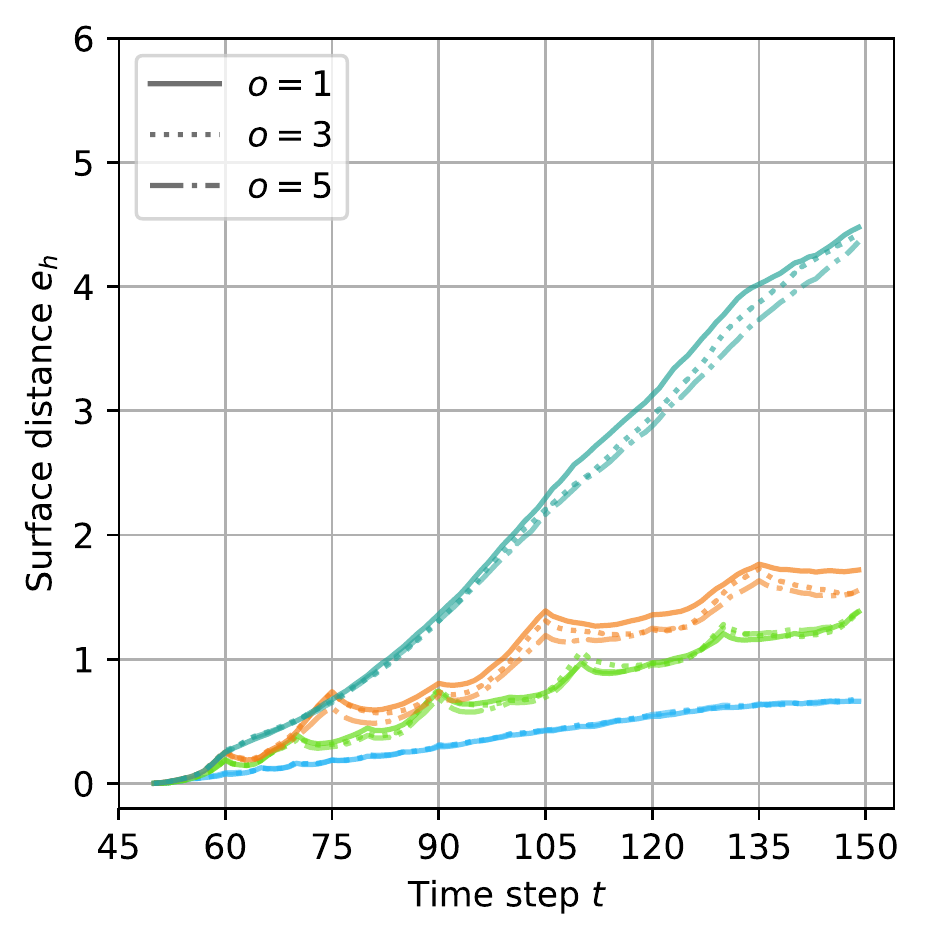}} 

	\end{tabular}
	\caption{Error graphs over time for 100 steps, averaged over ten {\em liquid64} simulations. Note that $o=1$ in \myreffig{fig:intervals_comparison_levelset_multiout}
	corresponds to \myreffig{fig:intervals_comparison_levelset}, and is repeated for comparison.
	}
	\label{fig:comparisons}
\end{figure*}						

\myreffig{fig:encdec_hausdorff} and 
\myreffig{fig:Mencdec_PSNR} show error measurements averaged for 10 simulations from the test data set.
Given the complexity of the data, especially the total pressure variant exhibits very good 
representational capabilities with an average PSNR value of $69.14$.
On the other hand, the velocity encoding introduces significantly larger errors in \myreffig{fig:Mencdec_PSNR}. 
Interestingly, neither the latent space normalization of the VAE, nor the split pressure data increase the reconstruction accuracy,
i.e., the CNN does not benefit from the reduced data range of the pressure splitting approach.
A visual comparison of the results can be found in \myreffig{fig:encdec_renderings}.

\subsection{Temporal Prediction}\label{subsec:temp_pred}

Next, we evaluate reconstruction quality when including the temporal prediction network.
Thus, now a quantity $\vec{x}'$ is inferred based on a series of previous latent space points.
For the following tests, our prediction model uses a history of $6$, and infers the next time step, thus $o=1$,
with a latent space size $m_s=1024$.
For a resolution of $64^3$ the fully recurrent network contains 700, and 1500 units for the first and second LSTM layer of \myreffig{fig:seq-to-seq}, respectively.
Hence, $\vec{d}$ has a dimensionality of 700 for this setup.
The two LSTM layers are followed by a convolutional layer targeting the $m_s$ latent space dimensions for our hybrid architecture,
or alternatively another LSTM layer of size $m_s$ for the fully recurrent version.
A dropout rate of $1.32\cdot10^{-2}$ with a recurrent dropout of $0.385$,
and a learning rate of $1.26\cdot10^{-4}$ with a decay factor of $3.34\cdot10^{-4}$
were used for all trainings of the prediction network.
Training was run for $50$ epochs with RMSProp,
with $319600$ training samples in each epoch, taking 2 hours, on average.
Hyperparameters as well as the length of the time history used as input for the prediction network and the generated output time steps were chosen by utilizing a hyper parameter search, i.e. training multiple configurations of the same network with differing input-/output counts or hyperparameter settings.

\begin{table}[bt]
	\centering
		\footnotesize
		\begin{tabular}{ c  l  l  c }
										& Layer (Type) 		& Activation 	& Output Shape \\ \hline \hline
			\multirow{2}{*}{$\tilde{f_{t_e}}$} 	& Input           	&				& ($n+1$, $m_s$) \\
										   & LSTM     			& tanh			& ($m_t$) \\ \hline
			\multirow{1}{*}{Context}   	& Repeat          	& 				& ($o$, $m_t$) \\ \hline
			\multirow{2}{*}{$\tilde{f_{t_d}}$} 	& LSTM     			& tanh			& ($o$, $m_{t_d}$) \\
										   & Conv1D 			& linear		& ($o$, $m_s$) \\ 
		\end{tabular}
	\vspace{0.3cm}
		\caption{Recurrent prediction network with hybrid architecture (\myreffig{fig:seq-to-seq})}
		\label{tab:recenctimeconvdec}
\end{table}

\begin{table}[bt]
\centering
	\footnotesize
	\begin{tabular}{ c l l c }
									& Layer (Type) 	& Activation 	& Output Shape \\ \hline \hline
		\multirow{2}{*}{$\tilde{f_{t_e}}$} 	& Input         &				& ($n+1$, $m_s$) \\
										& LSTM		   	& tanh			& ($m_t$) \\ \hline
		\multirow{1}{*}{Context}   	& Repeat        &				& ($o$, $m_t$) \\ \hline
		\multirow{2}{*}{$\tilde{f_{t_d}}$} 	& LSTM	   		& tanh			& ($o$, $m_{t_d}$) \\
										& LSTM   		& tanh			& ($o$, $m_s$) \\
	\end{tabular}
\vspace{0.3cm}
	\caption{Fully recurrent network architecture (\myreffig{fig:fullyrecurrent})}
	\label{tab:fullyrecurrent}
\end{table}

\begin{figure*}
	\footnotesize \centering
			\begin{overpic}[width = \myq]{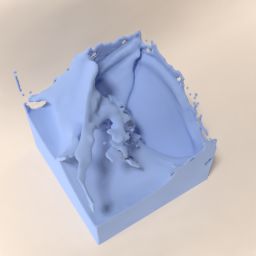} \put( 6, 90 ){\scriptsize \color{black}{Reference (GT)}} \end{overpic}\ 
			\begin{overpic}[width = \myq]{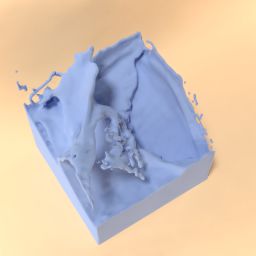} \put( 6, 90 ){\scriptsize \color{black}{$p_t$}} \end{overpic}\ 
			\begin{overpic}[width = \myq]{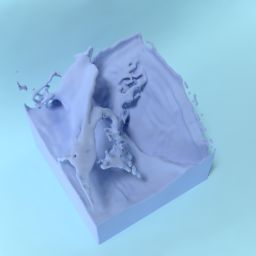}  \put( 6, 90 ){\scriptsize \color{black}{$p_s$ and $p_d$}} \end{overpic}\ 
			\begin{overpic}[width = \myq]{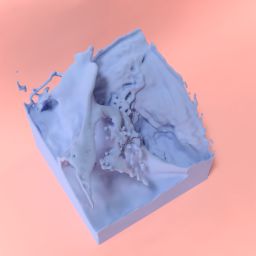}  \put( 6, 90 ){\scriptsize \color{black}{VAE $p_s$ and $p_d$}} \end{overpic}\ 
			\begin{overpic}[width = \myq]{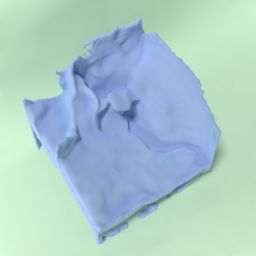}  \put( 6, 90 ){\scriptsize \color{black}{$\vec{u}$}} \end{overpic}
		\caption{Liquid surfaces predicted by different models for 40 steps with $i_p=\infty$.
			While the velocity version (green) leads to large errors in surface position,
		all three pressure versions closely capture the large scale motions. On smaller scales, both split pressure variants ($p_s$ and $p_d$) introduce artifacts.}
		\label{fig:prediction_renderings}
\end{figure*}

The error measurements for simulations predicted by the combination of autoencoder and prediction network are shown in \myreffig{fig:basenetworkcomparisoni1000} and \myreffig{fig:Mbasenetworkcomparisoni1000_psnr}, with a surface visualization in  \myreffig{fig:prediction_renderings}.
Given the autoencoder baseline, the prediction network does very well at predicting future states for the simulation variables. 
The accuracy only slightly decreases compared to \myreffig{fig:encdec_hausdorff} and \ref{fig:Mencdec_PSNR}, with an average PSNR value of $64.80$ (a decrease of only 6.2\% w.r.t. the AE baseline).  
\newNils{Here, it is also worth noting that the LSTM does not benefit from the normalized latent space
of the VAE. On the contrary, the predictions without the regular AE exhibit a lower error.}

\myreffig{fig:basenetworkcomparisoni15} shows an evaluation of the interval prediction scheme explained above.
Here we employ the LSTM for $i_p=14$ consecutive steps, and then perform a single regular simulation step. 
This especially improves the pressure predictions, for which the average surface error after 100 steps is still below two cells.

\begin{figure*}
	\vspace{0.3cm}
	\centering \footnotesize
	{\renewcommand{\arraystretch}{0}
	\newcommand{\mwa}{0.49\textwidth}
	\begin{tabular}{@{}c@{~~~}c@{}c@{}}
\begin{overpic}[width = \mwa]{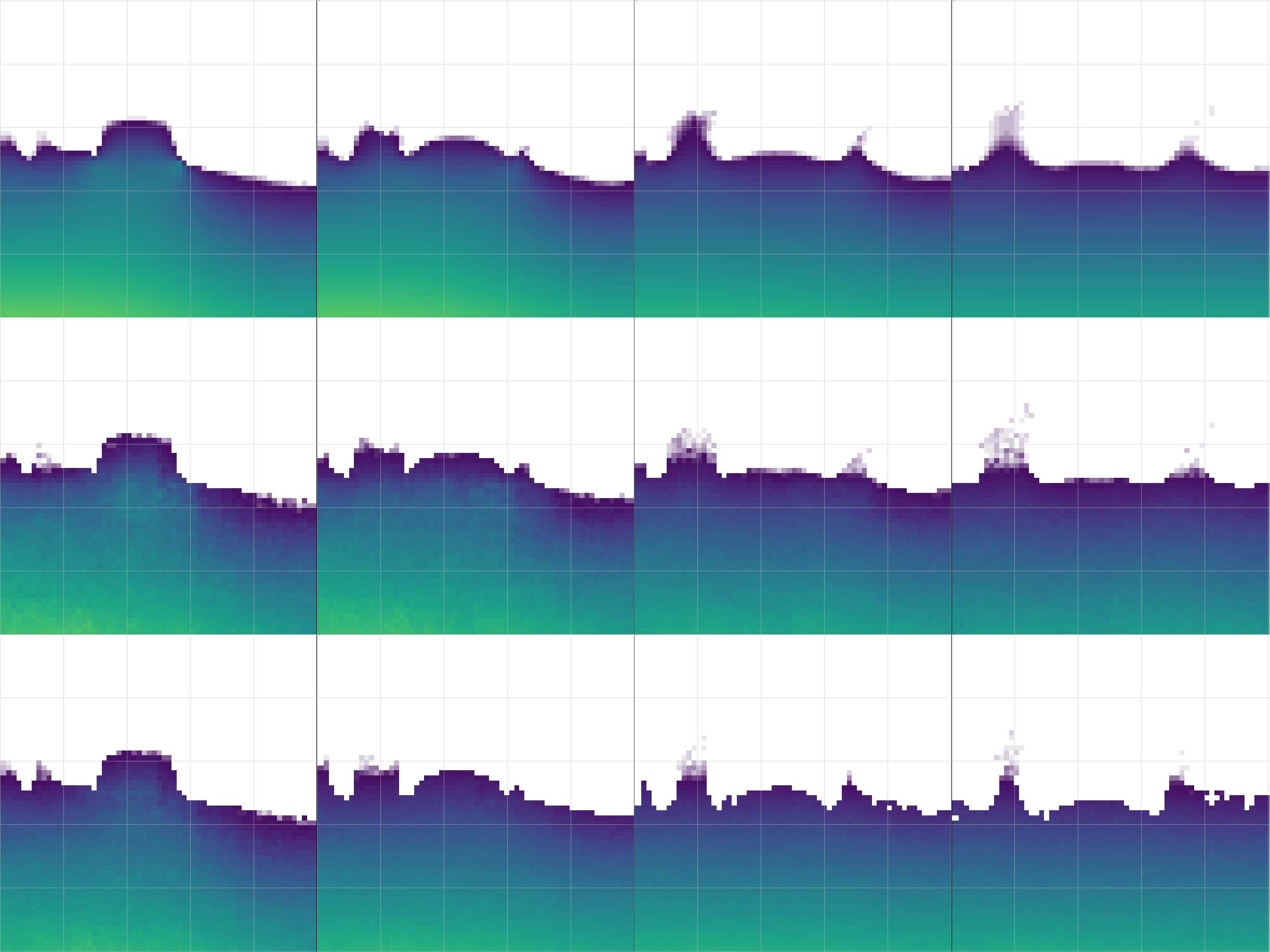}
	\put( 0, 77 ){\scriptsize \color{black}{a)}}
	\put( 1, 70 ){\scriptsize \color{gray}{GT}} 
	\put( 1, 45 ){\scriptsize \color{gray}{AE}} 
	\put( 1, 20 ){\scriptsize \color{gray}{LSTM}} 
\end{overpic}&
\begin{overpic}[width = \mwa]{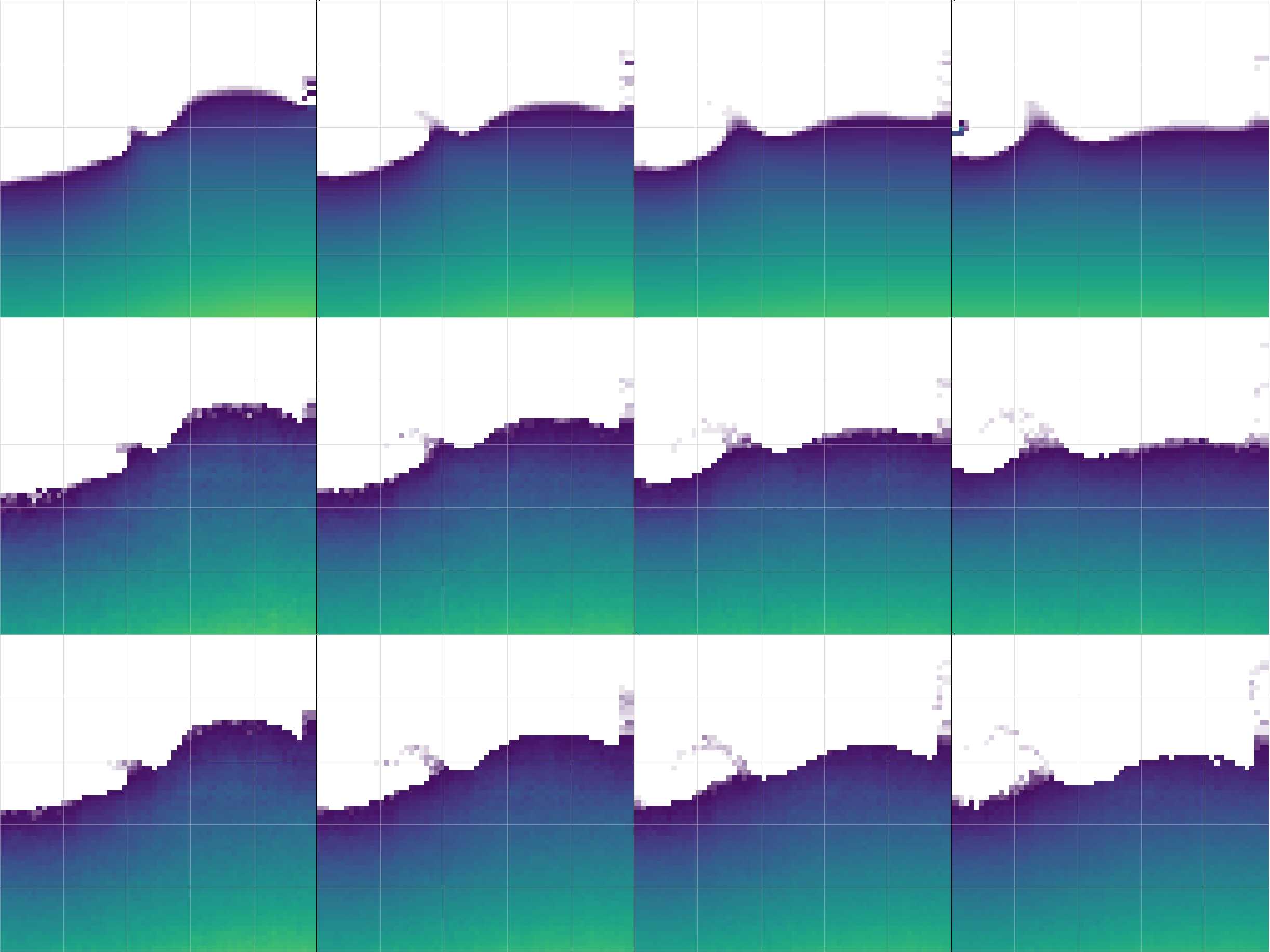}
	\put( 0, 77 ){\scriptsize \color{black}{b)}}
	\put( 1, 70 ){\scriptsize \color{gray}{GT}} 
	\put( 1, 45 ){\scriptsize \color{gray}{AE}} 
	\put( 1, 20 ){\scriptsize \color{gray}{LSTM}} 
\end{overpic}&
\\
	\end{tabular}
	}
	\caption{
	Two examples of ground truth pressure fields (top), the autoencoder baseline (middle),
	and the LSTM predictions (bottom). 
	Both examples have resolutions of $64^2$, 
	and are shown over the course of a long prediction horizon of 30 steps. 
	The network successfully predicts the temporal evolution within the latent space with $i_p=\infty$,
	as shown in the bottom row. }
	\label{fig:data_fields_lstm_noj}
\end{figure*}

\newNils{We also evaluate how 
well our model can predict future states based on a single set of inputs.
For multiple output steps, i.e. $o>1$, our model predicts several latent space points from a single time context $\ct$.
A graph comparing accuracy for 1, 3 and 5 steps of output can be found in \myreffig{fig:intervals_comparison_levelset_multiout}.
It is apparent that the accuracy barely degrades when multiple steps are predicted at once.
However, this case is significantly more efficient for our model.
E.g., the $o=3$ prediction only requires 30\% more time to evaluate, despite generating three times as many predictions (details can be found in \myrefsec{sec:results}).
Thus, the LSTM context successfully captures the state of the temporal latent space evolution, such that the model can predict future states almost as far as the given input history.}

\begin{figure*}
	\newcommand{\mwr}{0.14\textwidth}
\newcommand{\squeeze}{\hspace{-0.16cm}}
\centering
\begin{overpic}[width = \mwr]{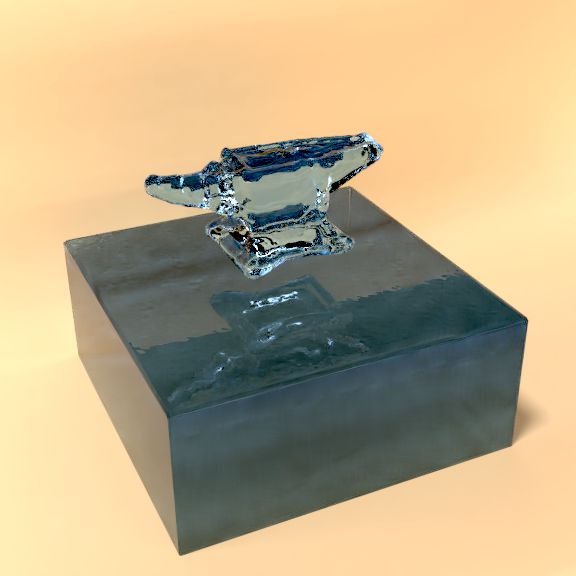} \put( 6, 88 ){\scriptsize \color{black}{a) $t=0$}} \end{overpic}
	\squeeze
\begin{overpic}[width = \mwr]{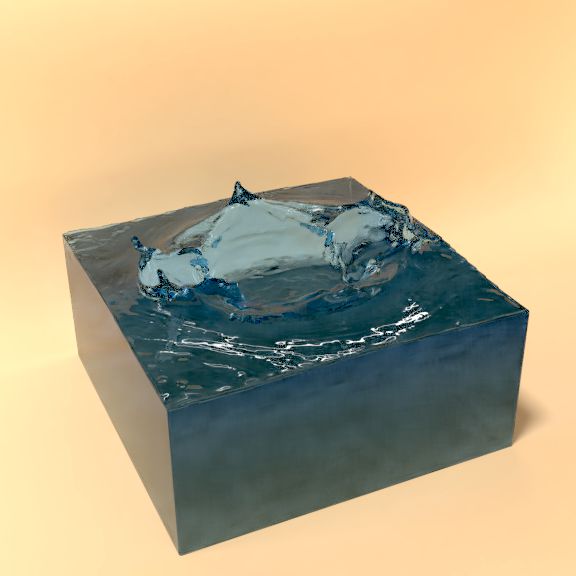} \put( 6, 88 ){\scriptsize \color{black}{$t=100$}} \end{overpic}
	\squeeze
\begin{overpic}[width = \mwr]{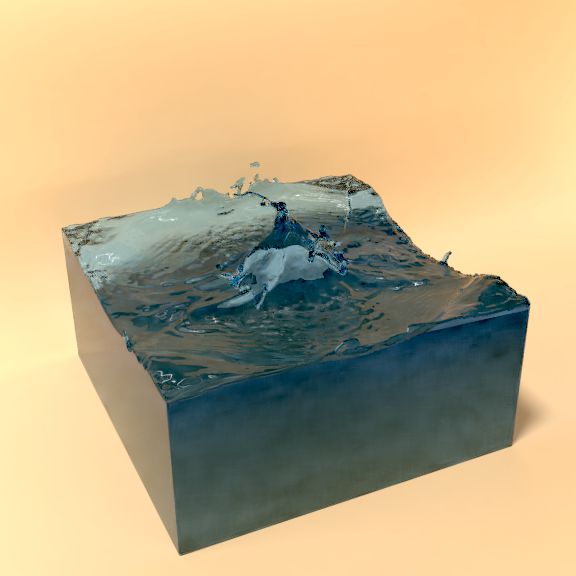} \put( 6, 88 ){\scriptsize \color{black}{$t=200$}} \end{overpic}
	\squeeze
\begin{overpic}[width = \mwr]{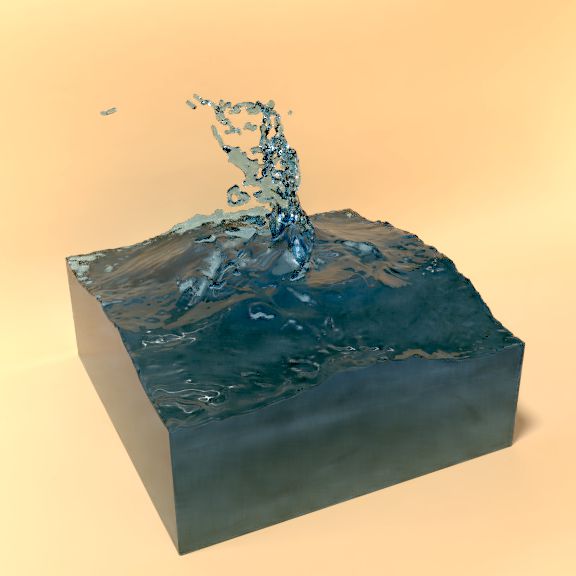} \put( 6, 88 ){\scriptsize \color{black}{$t=300$}} \end{overpic}
\hspace{0.1cm}
\begin{overpic}[width = \mwr]{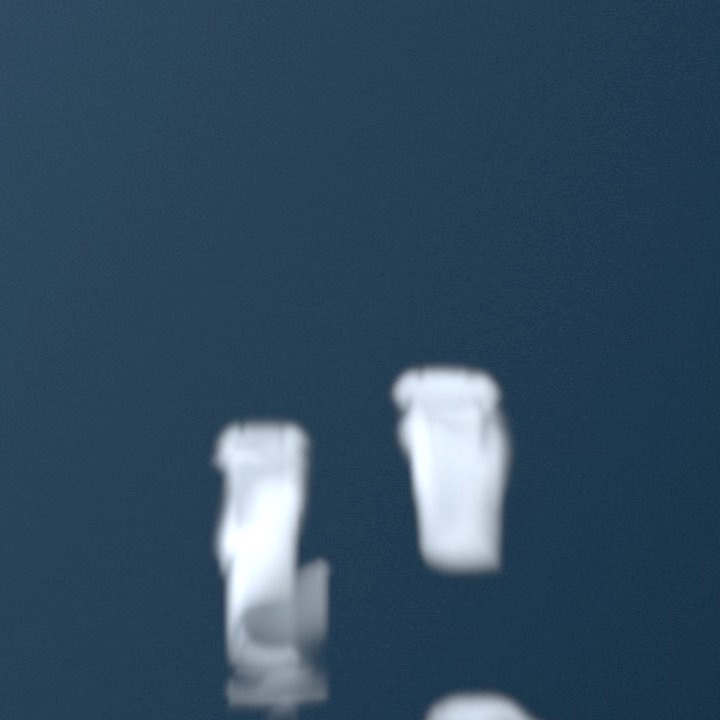} \put( 6, 88 ){\scriptsize \color{white}{b) $t=20$}} \end{overpic}
	\squeeze
\begin{overpic}[width = \mwr]{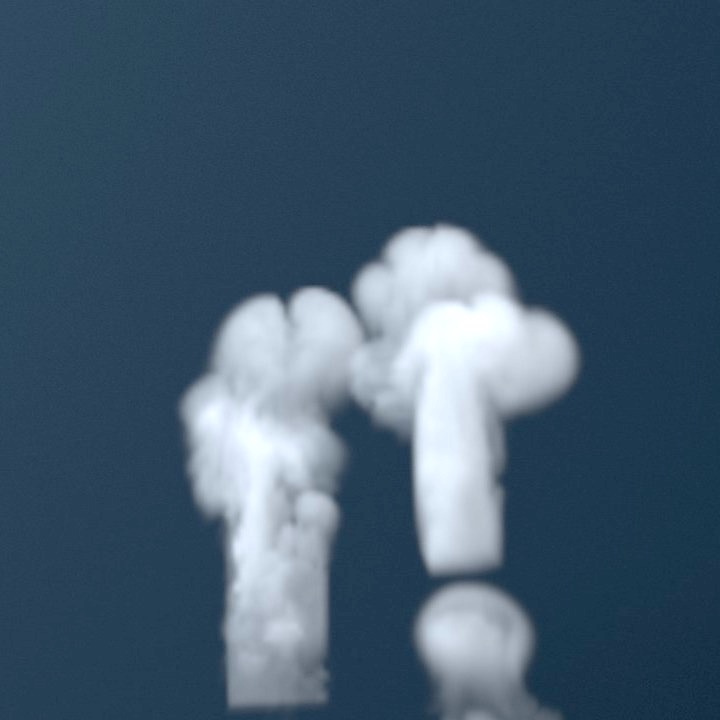} \put( 6, 88 ){\scriptsize \color{white}{$t=53$}} \end{overpic}
	\squeeze
\begin{overpic}[width = \mwr]{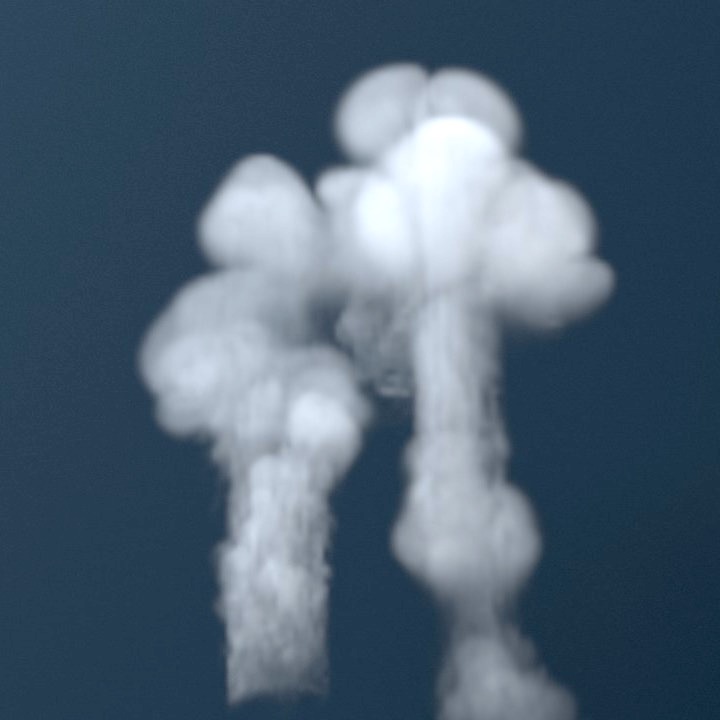} \put( 6, 88 ){\scriptsize \color{white}{$t=86$}} \end{overpic}
\caption{a) A test simulation with our {\em liquid128} model. The initial anvil shape was not part of the training data, but our
model successfully generalizes to unseen shapes such as this one. b) A test simulation configuration for our {\em smoke128} model.}
\label{fig:renderPreview}
\end{figure*}

A comparison of a fully recurrent LSTM with our proposed hybrid alternative can be found in 
\myreffig{fig:Malgcomparisoni1000_psnr}.
In this scene, representative for our other test runs, the hybrid architecture outperforms the 
fully recurrent (FR) version in terms of accuracy, while using 8.9m fewer weights than the latter. 
The full network sizes are 19.5m weights for hybrid, and 28.4m for the FR network.
We additionally evaluate a hybrid architecture with an additional conv. layer of size 4096 with tanh
activation after the LSTM decoder layer (V2 in \myreffig{fig:Malgcomparisoni1000_psnr}). This variant yields
similar error measurements to the original hybrid architecture. We found in general
that additional layers did not significantly improve prediction quality in our setting.
The FR version for the $128^3$ data set below requires 369.4m weights due to its increased
latent space dimensionality, which turned out to be infeasible. Our hybrid variant has 64m weights,
which is still a significant number, but yields accurate predictions and reasonable training times.
Thus, in the following tests, a total pressure inference model with a hybrid LSTM architecture
for $o=1$ will be used unless otherwise noted.

To clearly show the full data sets and their evolution over the course of a temporal prediction, 
we have trained a two-dimensional model,
the details of which are given in \myrefapp{sec:additional_results}.
In \myreffig{fig:data_fields_lstm_noj} sequences of the ground truth data are compared
to the corresponding autoencoder baseline, and the outputs of our prediction network.
Even though the autoencoder produces noise due to the strong compression, the temporal predictions closely match the autoencoder baseline, and the network is able to reproduce the complex behavior of the underlying simulations.
E.g., the two waves forming on the right hand side of the domain in \myreffig{fig:data_fields_lstm_noj}a indicate that the network successfully learned an abstraction of the temporal evolution of the flow.
Further tests of the full prediction network with autoencoder models that utilize gradient losses to circumvent the visual noise, yielded no better prediction capabilities than the presented autoencoder with L2 loss.
Additional 2D examples using the presented autoencoder can be found in \myreffig{fig:data_fields_prediction} of \myrefapp{sec:additional_results}.

\section{Results}\label{sec:results}

We now apply our model to the additional data sets with higher spatial resolutions, 
and we will highlight the resulting performance in more detail. 
First, we demonstrate how our method performs on the {\em liquid128} data set, with its
eight times larger number of degrees of freedom per volume.
Correspondingly, we use a latent space size of $m_s=8192$,
and a prediction network 
with LSTM layers of size 1000 and 1500.
Despite the additional complexity of this data set, our method successfully predicts the temporal
evolution of the pressure fields, with an average PSNR of 44.8. The lower value
compared to the $64^3$ case is most likely caused by the higher intricacy of the $128^3$ data.
\myreffig{fig:renderPreview}a) shows a more realistically rendered simulation for $i_p=4$. 
This setup contains a shape that was not part of any training data simulations.
Our model successfully handles this new configuration, as well as other
situations shown in the accompanying video. 
This indicates that our model generalizes to a broad class of physical behavior.
To evaluate long term stability, we have additionally simulated a scene for 650 time steps which successfully comes to rest. 
This simulation and additional scenes can be found in the supplemental video.

A trained model for the {\em smoke128} data set can be seen in \myreffig{fig:renderPreview}b.
Despite the significantly different physics,
our approach successfully predicts the evolution and motion of the vortex structures.
However, we noticed a tendency to underestimate pressure values, and
to reduce small-scale motions.
Thus, while our model successfully captures a significant part 
of the underlying physics, there is a clear room for improvement for this data set.

\begin{figure*}
	\centering
	\newcommand{\myvv}{0.2\textwidth}
	\renewcommand{\arraystretch}{0}
	\begin{tabular}{@{}c@{}c@{}c@{}c@{}c@{}}
				\subcaptionbox{$t=0$}{\includegraphics[trim={  176px, 0px, 176px, 0px}, clip, width = \myvv]{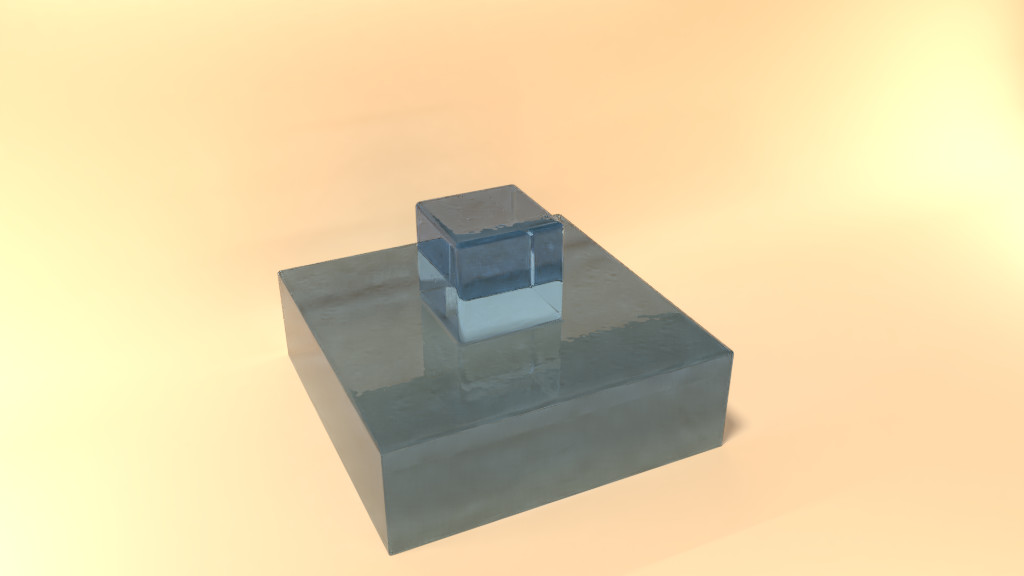}} &
		\subcaptionbox{$t=200$}{\includegraphics[trim={176px, 0px, 176px, 0px}, clip, width = \myvv]{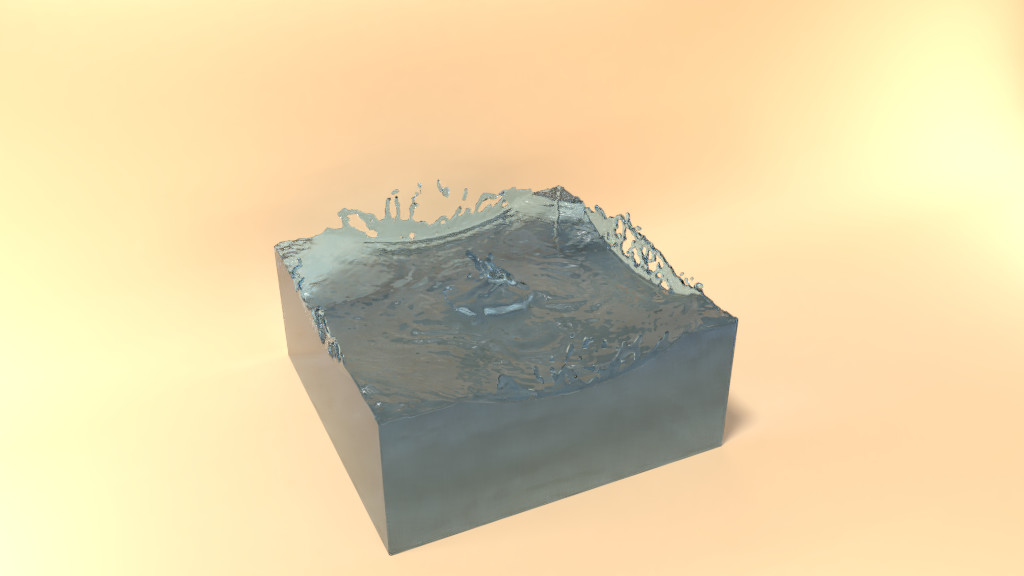}} &
		\subcaptionbox{$t=400$}{\includegraphics[trim={176px, 0px, 176px, 0px}, clip, width = \myvv]{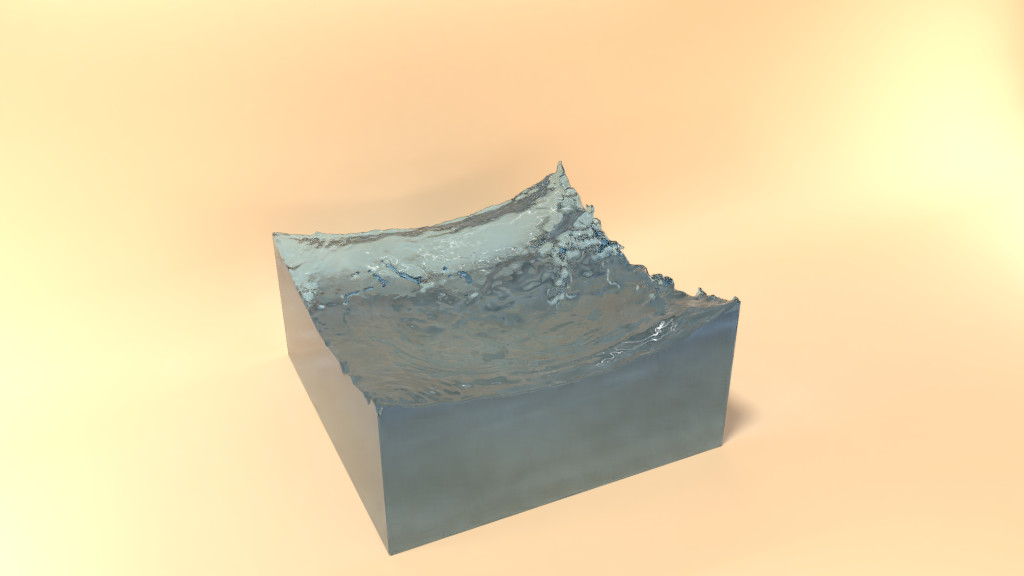}} &
		\subcaptionbox{$t=600$}{\includegraphics[trim={176px, 0px, 176px, 0px}, clip, width = \myvv]{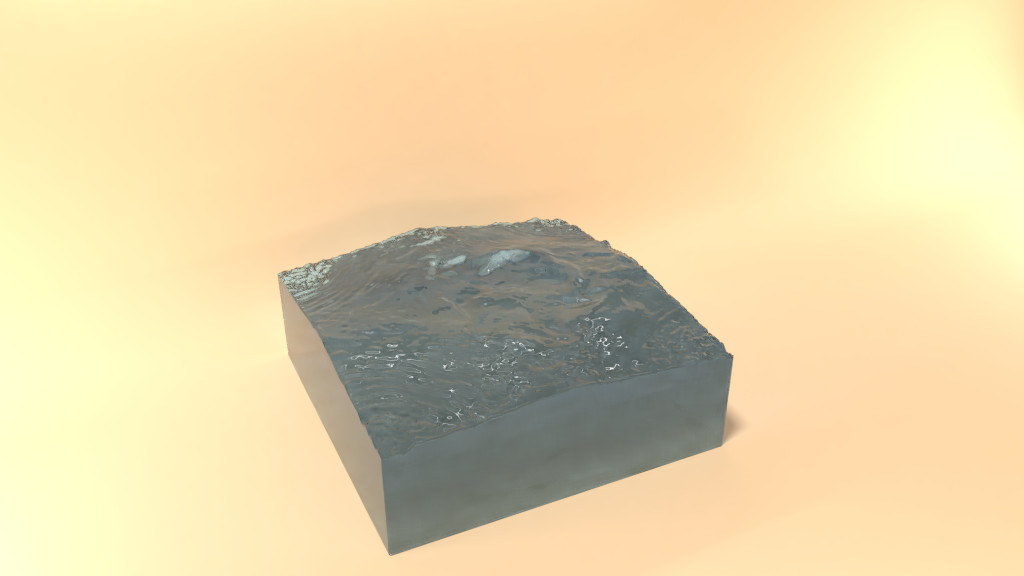}} &
		\subcaptionbox{$t=800$}{\includegraphics[trim={176px, 0px, 176px, 0px}, clip, width = \myvv]{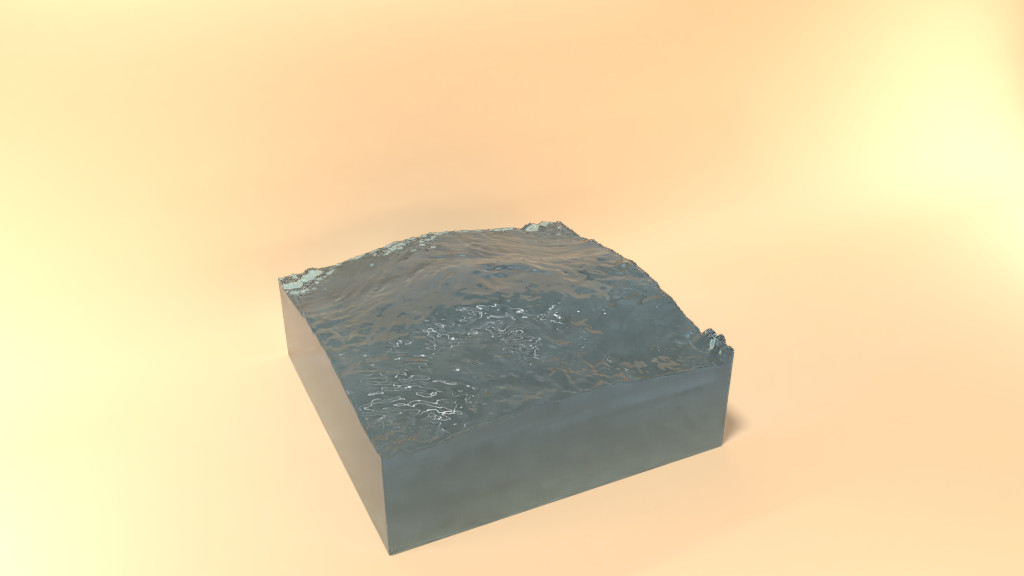}} 
	\end{tabular}
\vspace{0.3cm}
	\caption{Renderings of a long running prediction scene for the {\em liquid128} data set with $i_p=4$. The fluid successfully	comes to rest at the end of the simulation after 800 time steps.}
	\label{fig:resting_renderings}
\end{figure*}

Our method also leads to significant speedups compared to regular pressure solvers, especially for larger volumes.
For example the pressure inference by the prediction network for a $128^3$ volume takes 9.5ms, on average.
Including the times to encode and decode the respective simulation fields of resolution $128^3$ (4.1ms and 3.3ms, respectively) this represents a 155$\times$ speedup compared to a parallelized state-of-the-art iterative MIC-CG pressure solver \cite{bridson2015}, running with eight threads.
While the latter yields a higher overall accuracy, and runs on a CPU instead of a GPU, it also represents a highly optimized numerical method.
We believe the speedup of our LSTM version indicates a huge potential for very fast physics solvers with learned models.

It however, also leads to a degradation of accuracy compared to a regular iterative solver.
The degradation can be controlled by chosing an appropiate prediction interval as described in \myrefsec{subsec:temp_pred} and can therefore be set according to the required accuracy.
Even when taking into account a factor of ca. 10$\times$ for GPUs due to their better memory bandwidth, this leaves a speedup by a factor of more than 15$\times$, pointing to a significant gain in efficiency for our LSTM-based prediction. 
In addition, we measured the speedup for our (not particularly optimized) implementation,
where we include data transfer to and from the GPU for each simulation step.
This very simple implementation already yields practical speedups of 10x for an interval prediction with $i_p=14$.
Details can be found in \myreftab{tab:performance} and \myreftab{tab:performance64}, while \myreftab{tab:datasets} summarizes the sizes of our data sets.
All measurements were created with the {\em tensorflow timeline} tools on Intel i7 6700k (4GHz) and Nvidia Geforce GTX970.

\begin{table}[hbt]
	\centering
		\begin{tabular}{ c c c c }
			Interval $i_p$			& \makecell{Solve} & \makecell{Mean surf. dist} & Speedup  \\ \hline
			Reference 				& 2.629s 						  & 0.0 		& 1.0     \\
			$4$ 					& 0.600s 						  & 0.0187 		& 4.4     \\
			$9$ 					& 0.335s 						  & 0.0300 		& 7.8     \\
			$14$ 					& 0.244s 						  & 0.0365 		& 10.1    \\
			$\infty$				& 0.047s 						  & 0.0479 		& 55.9    \\
			& & & \\
									& Enc/Dec						  & Prediction 	& Speedup \\ \hline
			Core exec. time			& 4.1ms + 3.3ms					  & 9.5ms      	& 155.6
		\end{tabular}
	\vspace{0.3cm}
		\caption{Performance measurement of ten {\em liquid128} example scenes, averaged over 150 simulation steps each.
		The mean surface distance is a measure of deviation from the reference per solve.} 	\label{tab:performance}
\end{table}
\begin{table}[hbt]
	\centering
	\begin{tabular}{ c c c c }
				 					&  \multicolumn{2}{c}{\makecell{Solve}}	& Speedup  \\ \hline
		Reference 					&  \multicolumn{2}{c}{169ms}   						& 1.0 \\
		& & & \\
									& Enc/Dec				& Prediction  	&  \\ \hline
		\makecell{Core exec., $o=1$} & 3.8ms + 3.2ms	& 9.6ms			& 10.2 \\
		\makecell{Core exec., $o=3$} & 3.9ms + $3*3.3$ms	& 12.5ms		& 19.3 \\
	\end{tabular}
\vspace{0.3cm}
	\caption{Performance measurement of ten {\em liquid64} example scenes, averaged over 150 simulation steps each.} 	\label{tab:performance64}
\end{table}

\subsection{Limitations}
While we have shown that our approach leads to large speed-ups and robust simulations
for a significant variety of fluid scenes, there are several areas with room for improvements and follow up work.
First, our LSTM at the moment strongly relies on the AE, which primarily
encodes large scale scale dynamics, while small scale dynamics are 
integrated by the alignment of free surface boundary conditions \cite{ando2015dimension}.
Also, our current, relatively simple AE can introduce a certain amount of noise in the solutions,
which, however, can potentially be alleviated by different network architectures.

Overall, improving the AE network is important in order to improve the quality of the temporal predictions.
Our experiments also show that larger data sets should
directly translate into improved predictions. This is especially important for the latent
space data set, which cannot be easily augmented.

\section{Conclusions}

With this work we arrive at three important conclusions:
first, deep neural network architectures can successfully predict the temporal evolution
of dense physical functions, second, 
learned latent spaces in conjunction with LSTM-CNN hybrids are highly suitable for this task,
and third, they can yield very significant increases in simulation performance.

In this way, we arrive at a data-driven solver that yields practical speed-ups, and at its core is more than 150x faster than a regular pressure solve. We believe that our work represents an important first step towards deep-learning powered simulation algorithms. 
On the other hand, given the complexity of the problem at hand, our approach represents only a first step. There are numerous, highly interesting avenues for future research, ranging from improving the accuracy of the predictions, over performance considerations, to using such physics predictions as priors for inverse problems.

\small
\bibliographystyle{eg-alpha-doi}
\bibliography{fluids}

\appendix

\newpage
\ \\

\begin{centering} \vspace{1cm}  \LARGE Supplemental Document for \mytitle  \\ \vspace{1cm}
\end{centering}

\normalsize

\section{Long-short Term Memory Units and Dimensionality}\label{app:netdetails}

A central challenge for deep learning problems involving fluid flow is the large number of degrees of freedom present in three-dimensional data sets.
This quickly leads to layers with large numbers of nodes -- from hundreds to thousands per layer.
Here, a potentially unexpected side effect of using LSTM nodes is the number of weights they require.

Therefore we briefly summarize the central equations for layers of LSTM units to gain understanding about the required weights.
Below, $f,i,o,g,s$ will denote forget, input, output, update and result connections, respectively.
$h$ denotes the result of the LSTM layer.
Subscripts denote time steps, while $\theta$ and $b$ denote weight and bias.
Below, we assume $\tanh$ as output activation function.
The new state for time step $t$ of an LSTM layer is then given by:
\begin{equation}
\label{eq:lstm_formulas}
	\begin{split}
		f_t &= \sigma( \theta_{xf} x_t + \theta_{hf} h_{t-1} + b_f ) \\
		i_t &= \sigma( \theta_{xi} x_t + \theta_{hi} h_{t-1} + b_i ) \\
		o_t &= \sigma( \theta_{xo} x_t + \theta_{ho} h_{t-1} + b_o ) \\
		g_t &= \tanh( \theta_{xg} x_t + \theta_{hg} h_{t-1} + b_g ) \\
		s_t &= f_t \odot s_{t-1} + i_t \odot g_t \\
		h_t &= o_t \odot \tanh(s_t)
	\end{split}
\end{equation}

The local feedback loops for the gates of an LSTM unit all have trainable weights, and as such induce an $n \times n$ weight matrix for $n$ LSTM units.
E.g., even for a simple network with a one dimensional input and output, and a single hidden layer of 1000 LSTM units, with only $2 \times 1000$ connections and weights between in-, output and middle layer, the LSTM layer internally stores $1000^2$ weights for its temporal feedback loop.
In practice, LSTM units have {\em input, forget and output} gates in addition to the feedback connections, leading to $4 n^2$ internal weights for an LSTM layer of size $n$. 
Correspondingly, the number of weights of such a layer with $n_o$ nodes, i.e., outputs, and $n_i$ inputs is given by $n_{\text{lstm}} = 4 (n_o^2 + n_o (n_i+1))$.
In contrast, the number of weights for the 1D convolutions we propose in the main document is
$n_{\text{conv-1d}} = n_o k (n_i + 1) $, with a kernel size $k=1$.

Keeping the number of weights at a minimum is in general extremely important to prevent overfitting, reduce execution times, and to arrive at networks which are able to generalize.
To prevent the number of weights from exploding due to large LSTM layers, we propose the mixed use of LSTM units and convolutions for our final temporal network architecture. 
Here, we change the decoder part of the network to consist of a single dense LSTM layer that generates a sequence of $o$ vectors of size $m_{t_d}$.
Instead of processing these vectors with another dense LSTM layer as before, we concatenate the outputs into a single tensor, and employ a single one-dimensional convolution translating the intermediate vector dimension into the required $m_s$ dimension for the latent space.
Thus, the 1D convolution works along the vector content, and is applied in the same way to all $o$ outputs.
Unlike the dense LSTM layers, the 1D convolution does not have a quadratic weight footprint, and purely depends on the size of input and output vectors.

\section{Additional Results}\label{sec:additional_results}

\begin{figure}[h]
	\footnotesize \centering
		\newcommand{\myv}{0.11} 	{\renewcommand{\arraystretch}{0}
		\newcolumntype{V}{>{\centering\arraybackslash}m{\myv\textwidth} }
		\begin{tabular}{@{}m{1.5em}@{}V@{}V@{}V@{}V@{}}
			\begin{minipage}{\myv\textwidth} \vspace{\myv\textwidth} \end{minipage} &
			t = 60 &
			t = 70 &
			t = 80 &
			t = 90 \\
			\rotatebox[]{90}{Reference} &
			\includegraphics[width = \myv\textwidth]{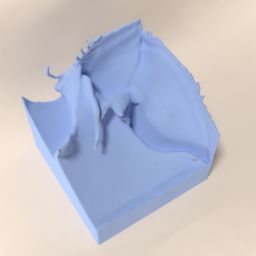} &
			\includegraphics[width = \myv\textwidth]{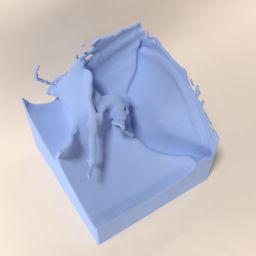} &
			\includegraphics[width = \myv\textwidth]{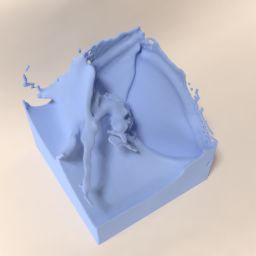} &
			\includegraphics[width = \myv\textwidth]{graphics/prediction/renderings/ref_6/pred_0090} \\		
			\rotatebox[]{90}{$p_s$, $p_d$} &
			\includegraphics[width = \myv\textwidth]{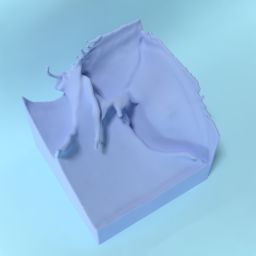} &
			\includegraphics[width = \myv\textwidth]{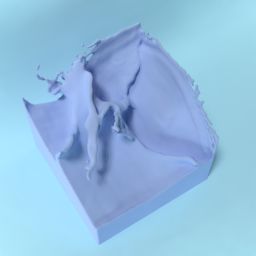} &
			\includegraphics[width = \myv\textwidth]{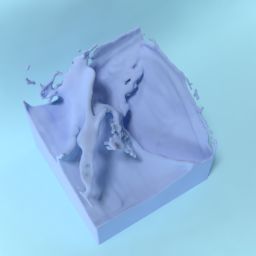} &
			\includegraphics[width = \myv\textwidth]{graphics/prediction/renderings/predict_split_6/pred_0090} \\		
			\rotatebox[]{90}{$p_t$} &
			\includegraphics[width = \myv\textwidth]{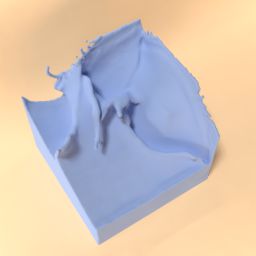} &
			\includegraphics[width = \myv\textwidth]{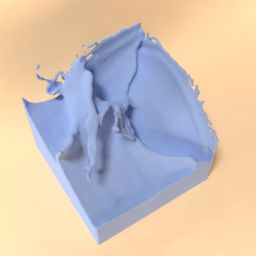} &
			\includegraphics[width = \myv\textwidth]{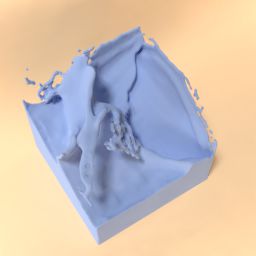} &
			\includegraphics[width = \myv\textwidth]{graphics/prediction/renderings/predict_total_6/pred_0090} \\
			\rotatebox[]{90}{VAE $p_s$, $p_d$} &
			\includegraphics[width = \myv\textwidth]{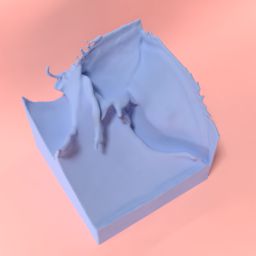} &
			\includegraphics[width = \myv\textwidth]{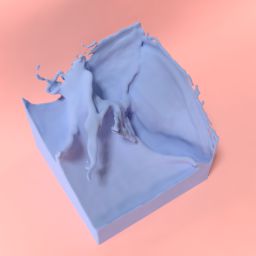} &
			\includegraphics[width = \myv\textwidth]{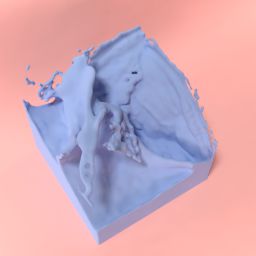} &
			\includegraphics[width = \myv\textwidth]{graphics/prediction/renderings/predict_vae_6/pred_0090} \\
			\rotatebox[]{90}{Velocity} &
			\includegraphics[width = \myv\textwidth]{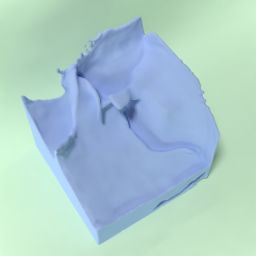} &
			\includegraphics[width = \myv\textwidth]{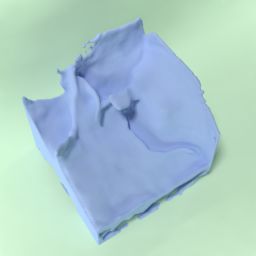} &
			\includegraphics[width = \myv\textwidth]{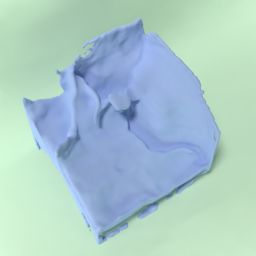} &
			\includegraphics[width = \myv\textwidth]{graphics/prediction/renderings/predict_vel_6/pred_0090} \\
		\end{tabular}
		}
		\caption{Additional comparison of liquid surfaces predicted by different architectures for 40 time steps for $i_p=\infty$, with prediction
		starting at time step 50. While the velocity version (green) leads to large errors in surface position,
		all three pressure versions closely capture the large scale motions. On smaller scales, both split pressure
		and especially VAE introduce artifacts.}
		\label{fig:prediction_renderingsOld}
\end{figure}

\begin{figure}[h]
	\centering
	\newcommand{\myvv}{0.25}
	\renewcommand{\arraystretch}{0}
	\begin{tabular}{@{}c@{}c@{}c@{}c@{}}
		\subcaptionbox*{$t=0$}{\includegraphics[trim={176px, 0px, 176px, 0px}, clip, width = \myvv\columnwidth]{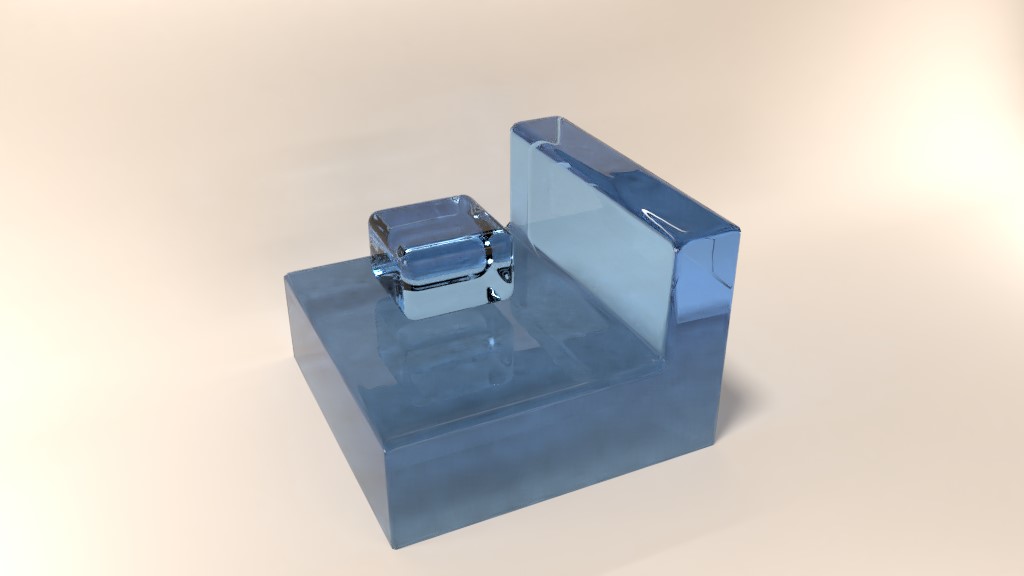}} &
		\subcaptionbox*{$t=50$}{\includegraphics[trim={176px, 0px, 176px, 0px}, clip, width = \myvv\columnwidth]{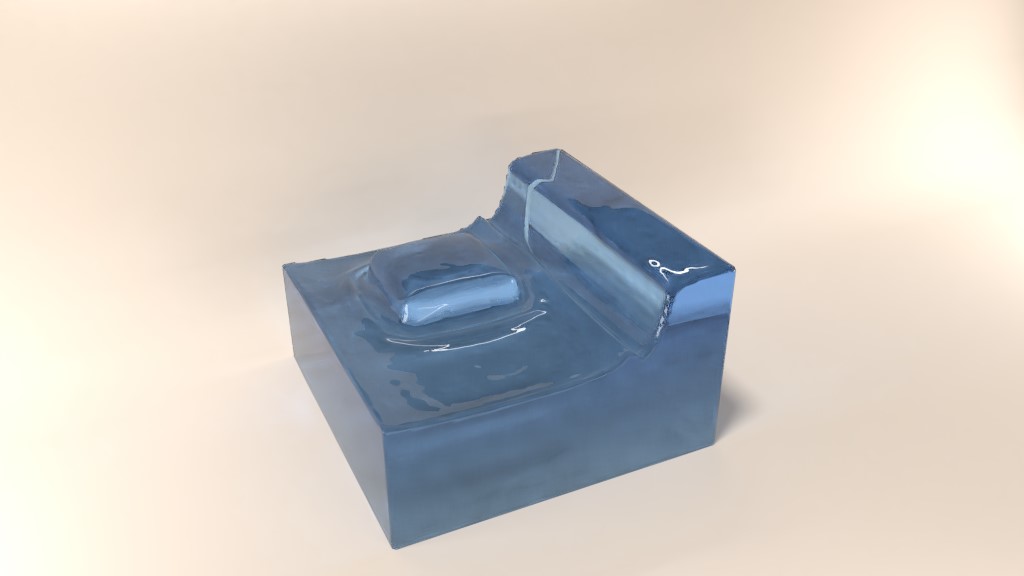}} &
		\subcaptionbox*{$t=100$}{\includegraphics[trim={176px, 0px, 176px, 0px}, clip, width = \myvv\columnwidth]{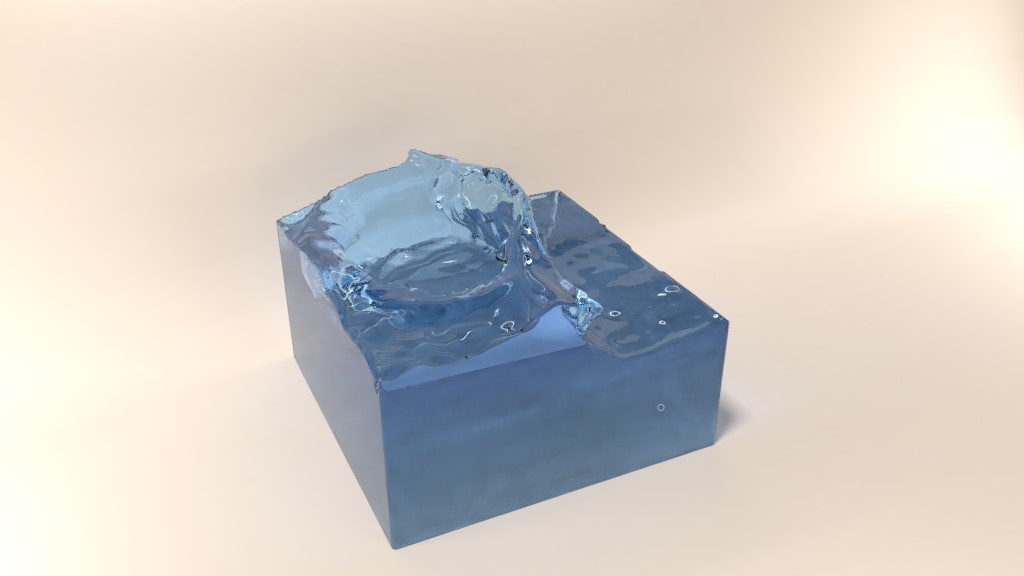}} &
		\subcaptionbox*{$t=150$}{\includegraphics[trim={176px, 0px, 176px, 0px}, clip, width = \myvv\columnwidth]{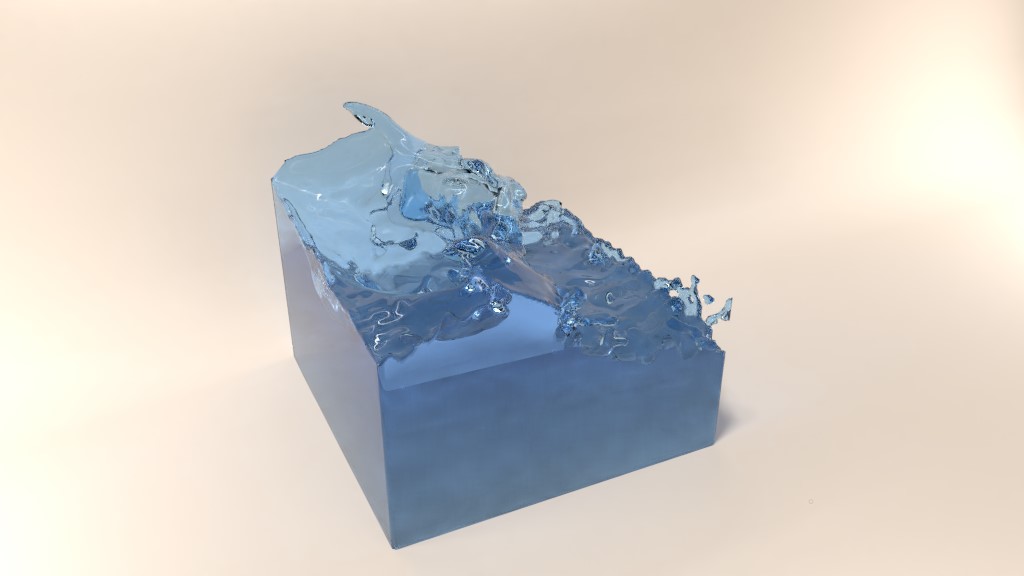}}
	\end{tabular}\vspace{-0.3cm}
	\caption{Renderings at different points in time of a $64^3$ scene predicted with $i_p=4$ by our network. }
	\label{fig:transparent_64}
\end{figure}

\begin{figure}[h]
	\centering
	\newcommand{\myv}{0.25\columnwidth} 
	\renewcommand{\arraystretch}{0}
	\begin{tabular}{@{}c@{}c@{}c@{}c@{}}
		\includegraphics[trim={176px, 0px, 176px, 0px}, clip, width = \myv]{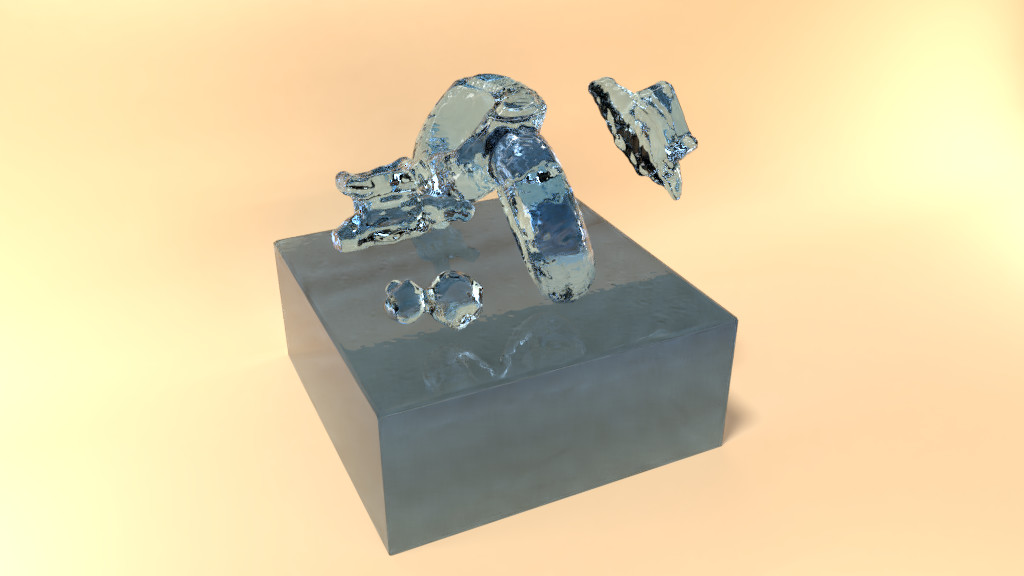} &
		\includegraphics[trim={176px, 0px, 176px, 0px}, clip, width = \myv]{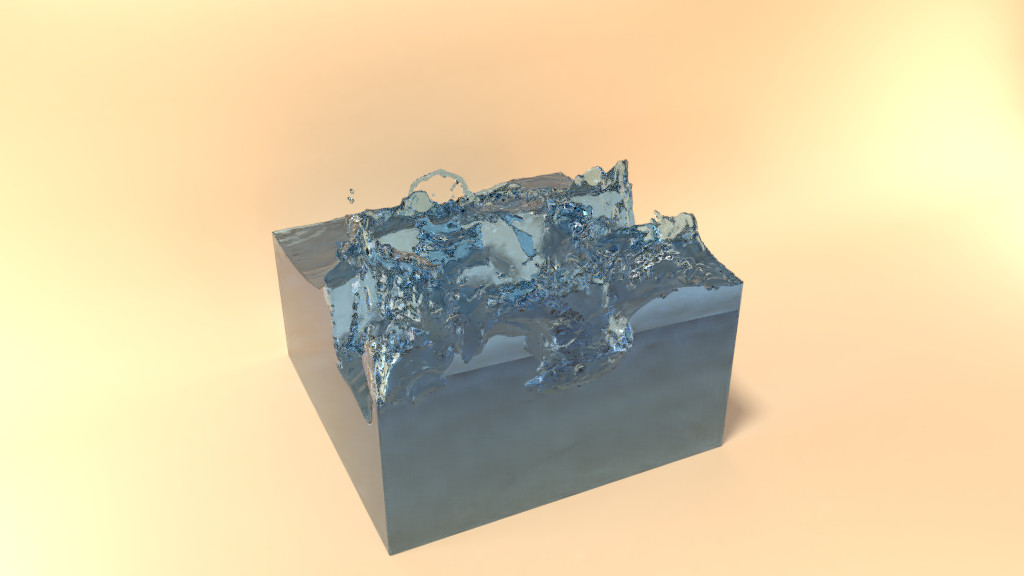} &
		\includegraphics[trim={176px, 0px, 176px, 0px}, clip, width = \myv]{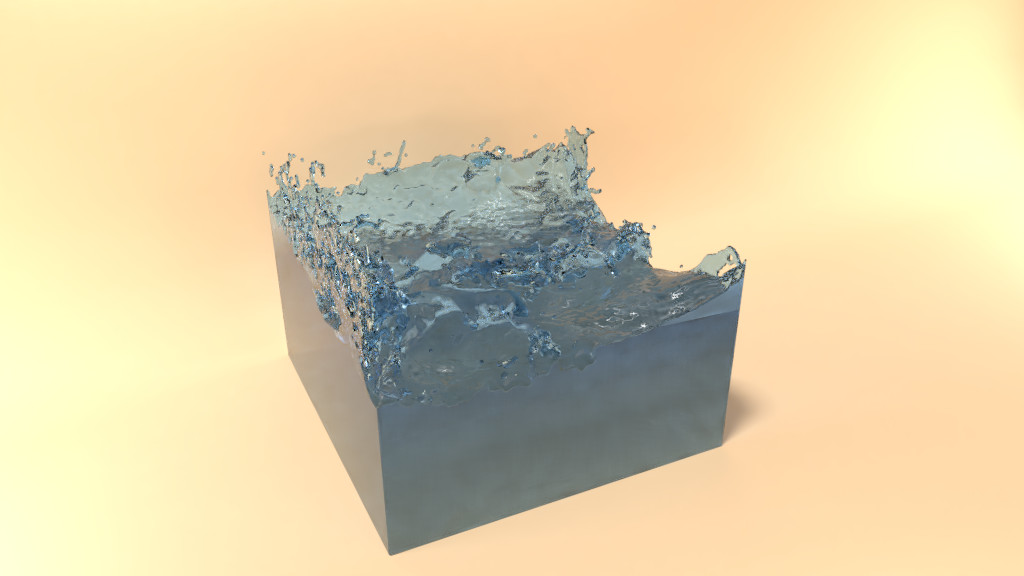} &
		\includegraphics[trim={176px, 0px, 176px, 0px}, clip, width = \myv]{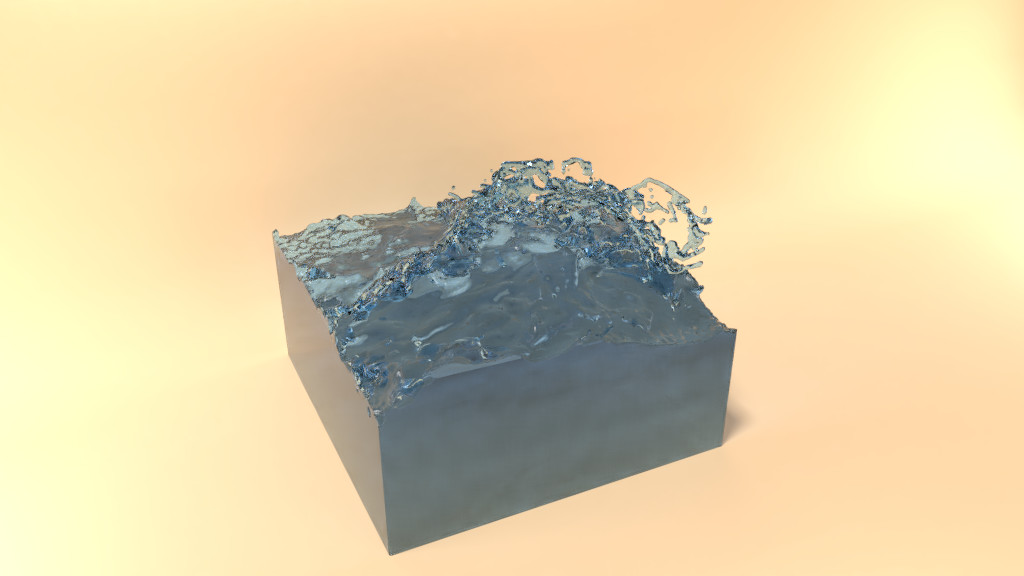} \\
		\includegraphics[trim={176px, 0px, 176px, 0px}, clip, width = \myv]{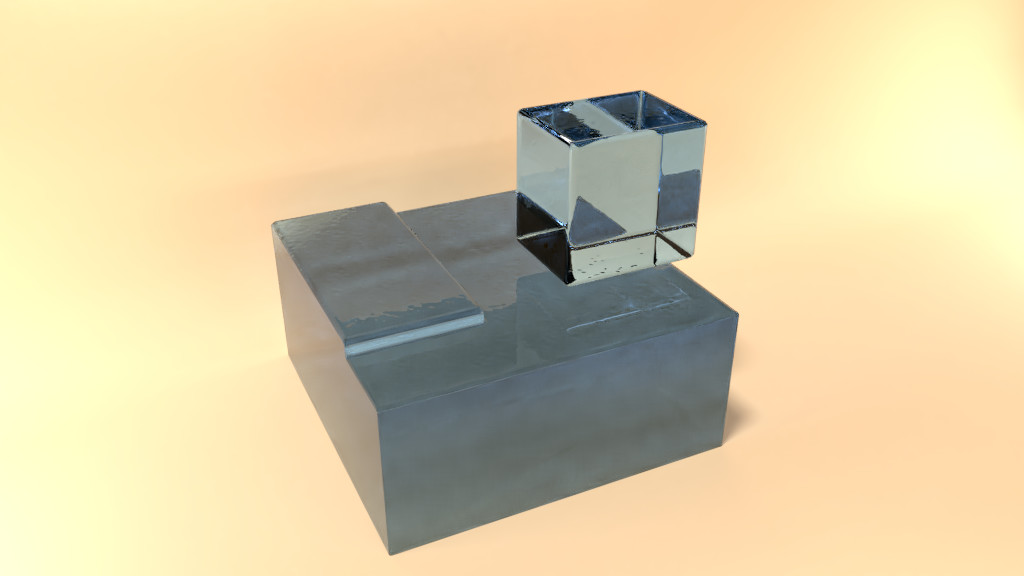} &
		\includegraphics[trim={176px, 0px, 176px, 0px}, clip, width = \myv]{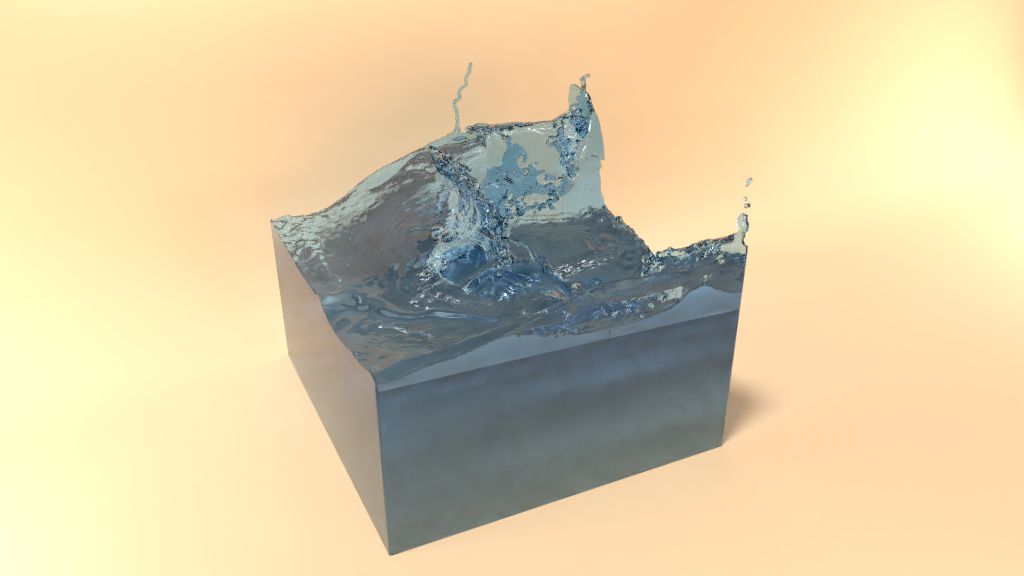} &
		\includegraphics[trim={176px, 0px, 176px, 0px}, clip, width = \myv]{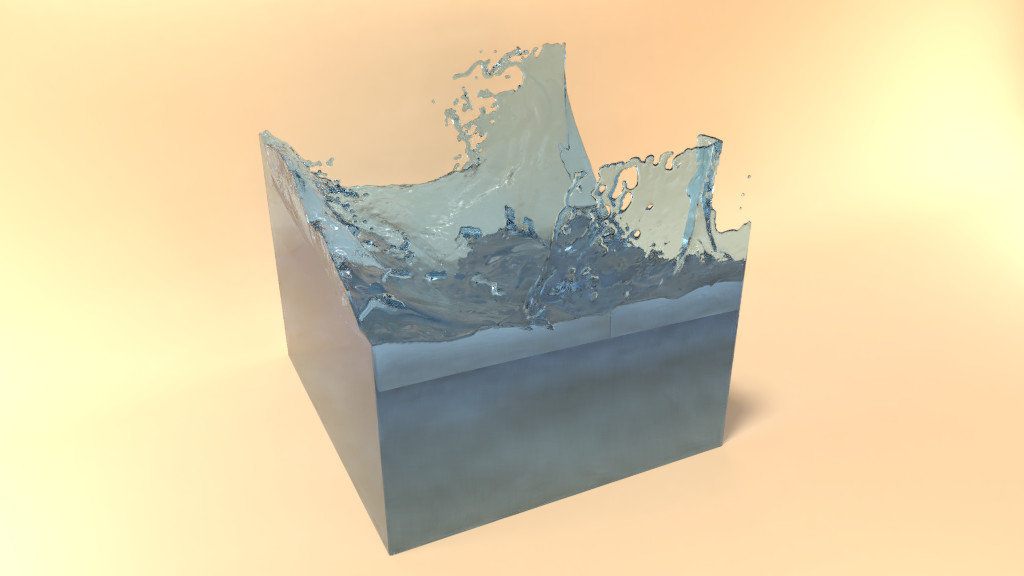} &
		\includegraphics[trim={176px, 0px, 176px, 0px}, clip, width = \myv]{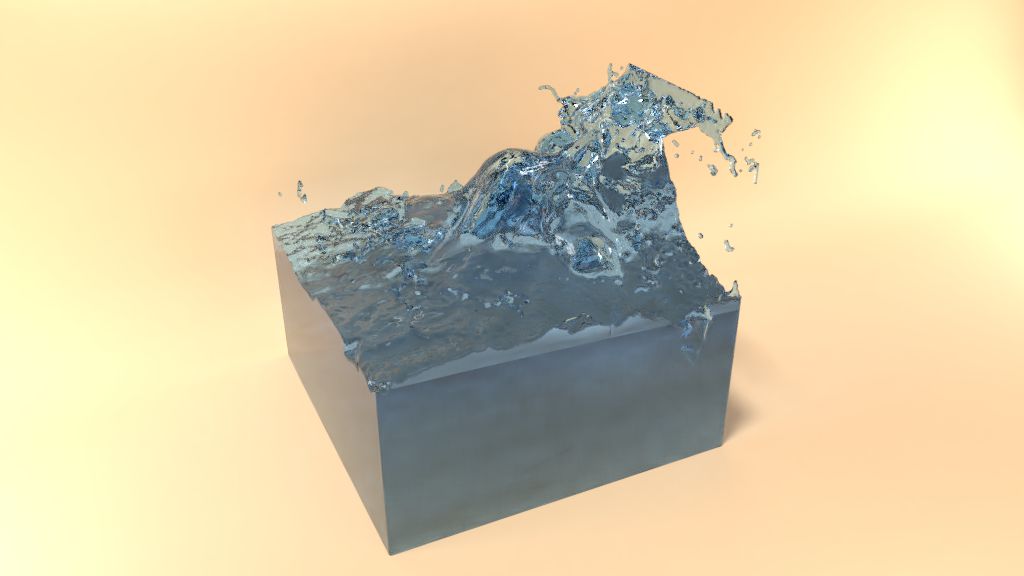} \\
		\subcaptionbox*{$t=0$}{\includegraphics[trim={176px, 0px, 176px, 0px}, clip, width = \myv  ]{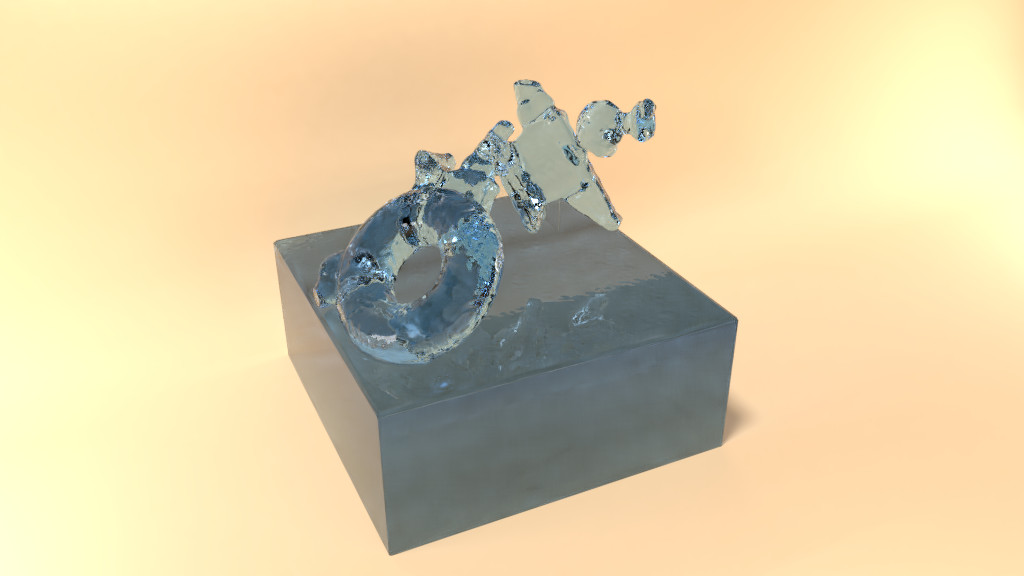}} &
		\subcaptionbox*{$t=100$}{\includegraphics[trim={176px, 0px, 176px, 0px}, clip, width = \myv]{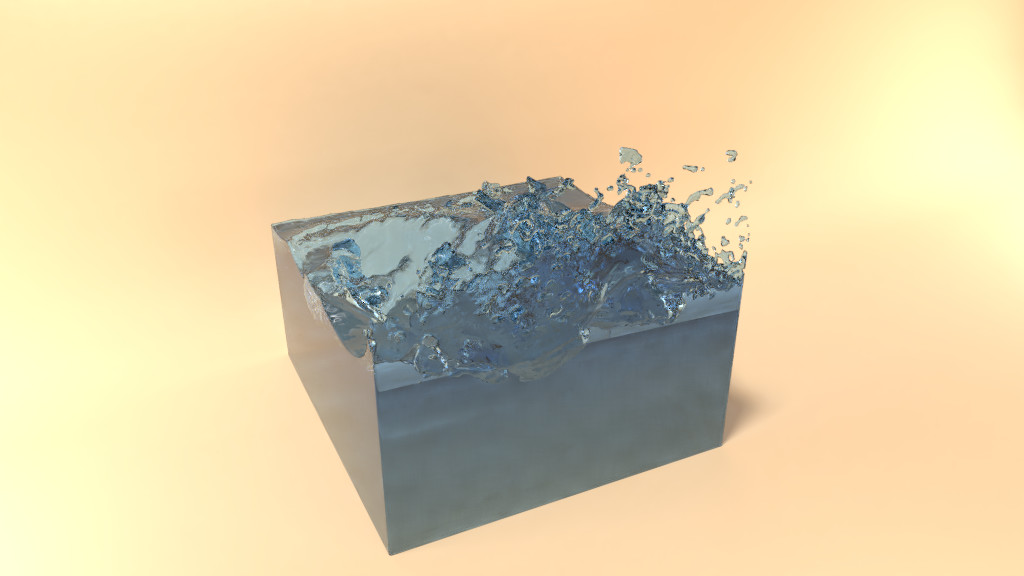}} &
		\subcaptionbox*{$t=200$}{\includegraphics[trim={176px, 0px, 176px, 0px}, clip, width = \myv]{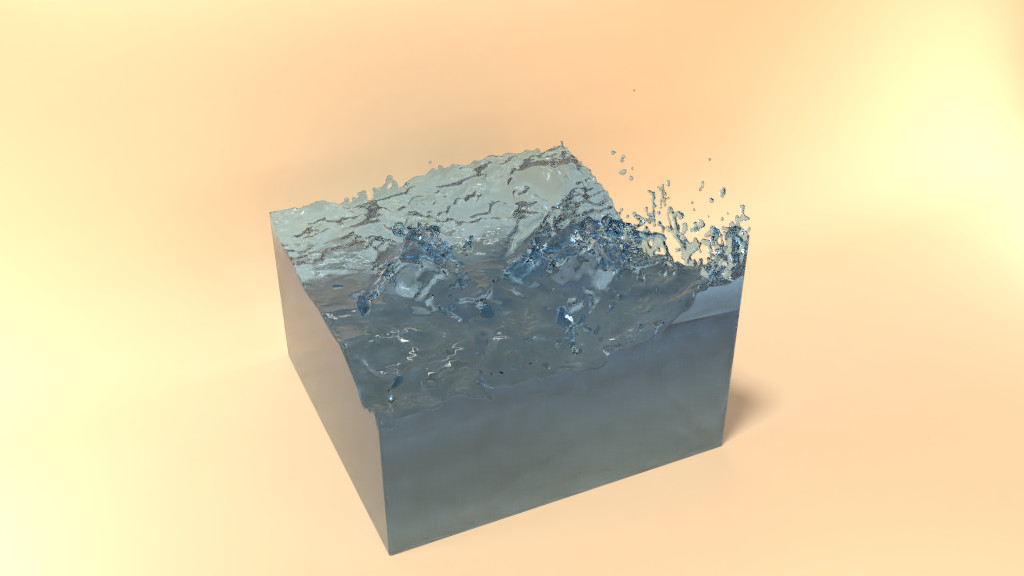}} &
		\subcaptionbox*{$t=300$}{\includegraphics[trim={176px, 0px, 176px, 0px}, clip, width = \myv]{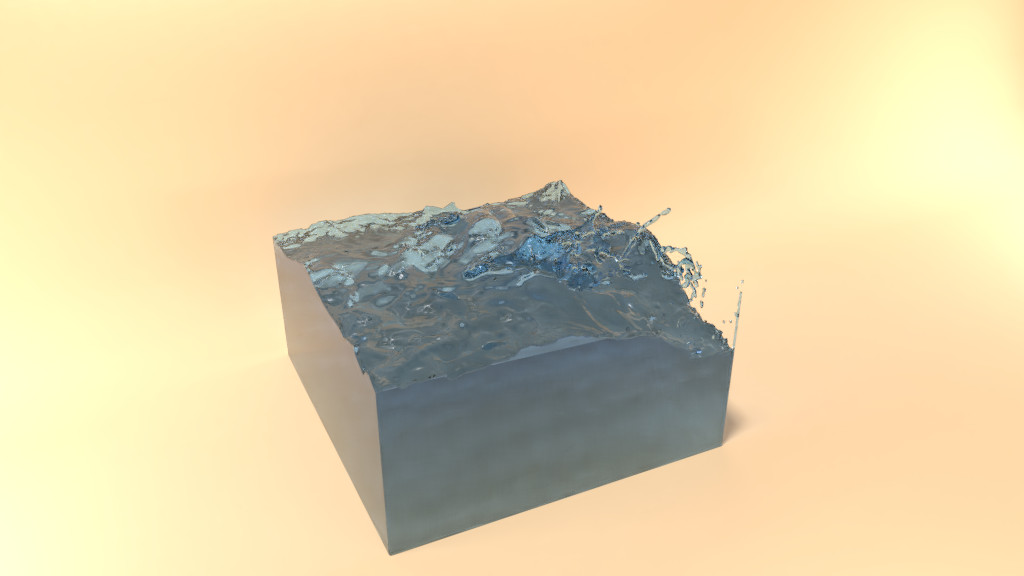}} \\
	\end{tabular} \vspace{-0.03cm}
	\caption{Additional examples of $128^3$ liquid scenes predicted with an interval of $i_p=4$ by our LSTM network.}
	\label{fig:transparent_128}
\end{figure}

\begin{figure}
	\centering
	\newcommand{\myvv}{0.25}
	\renewcommand{\arraystretch}{0}
	\begin{tabular}{@{}c@{}c@{}c@{}c@{}}
		\includegraphics[trim={224px, 0px, 224px, 0px}, clip, width = \myvv\columnwidth]{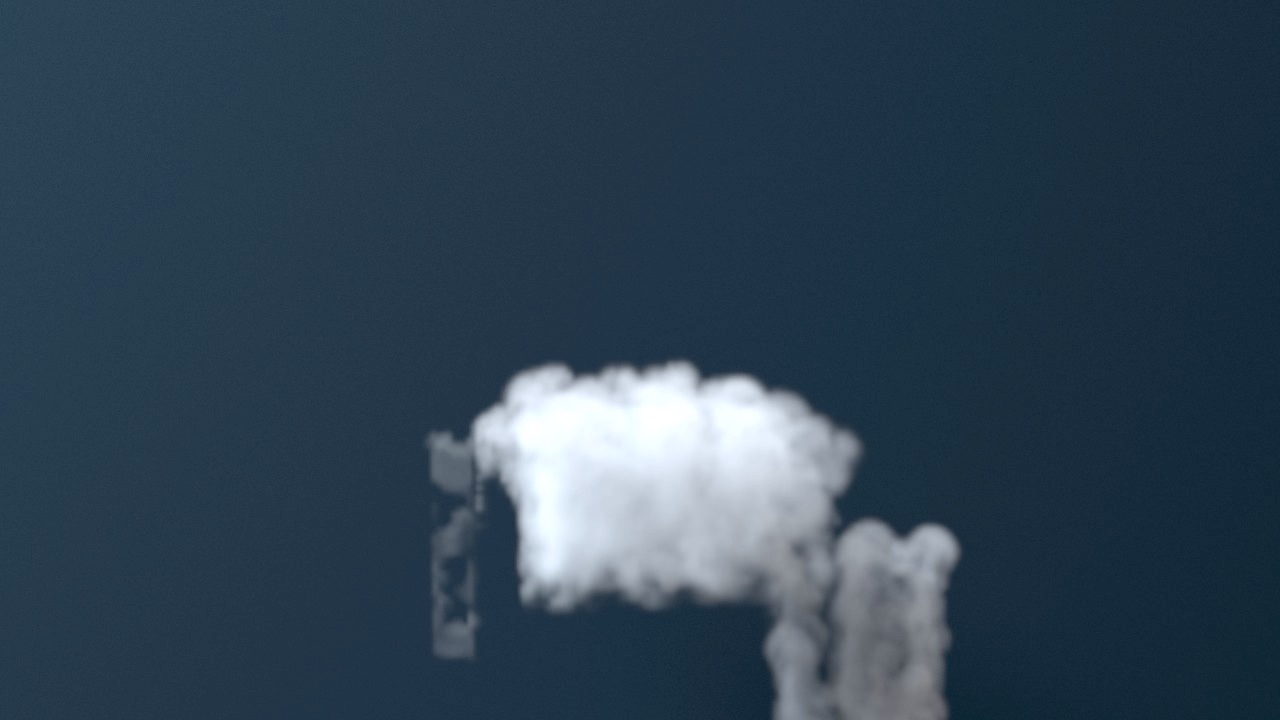} &
		\includegraphics[trim={224px, 0px, 224px, 0px}, clip, width = \myvv\columnwidth]{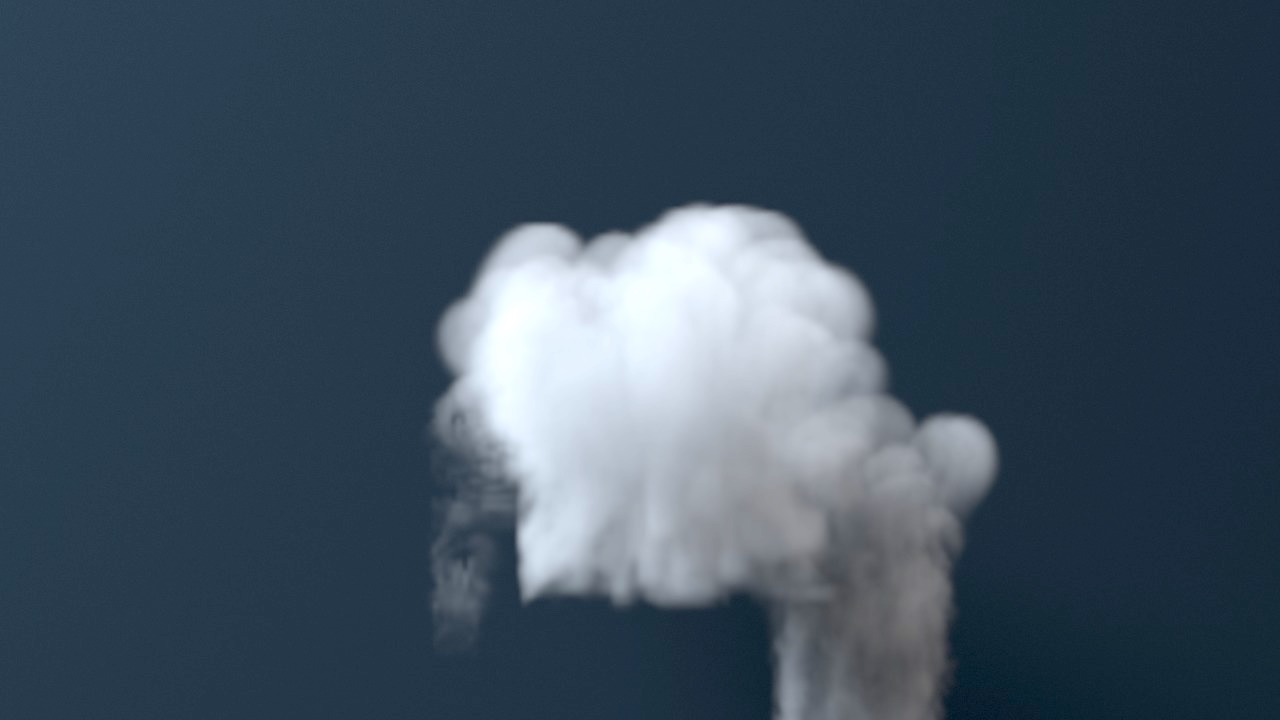} &
		\includegraphics[trim={224px, 0px, 224px, 0px}, clip, width = \myvv\columnwidth]{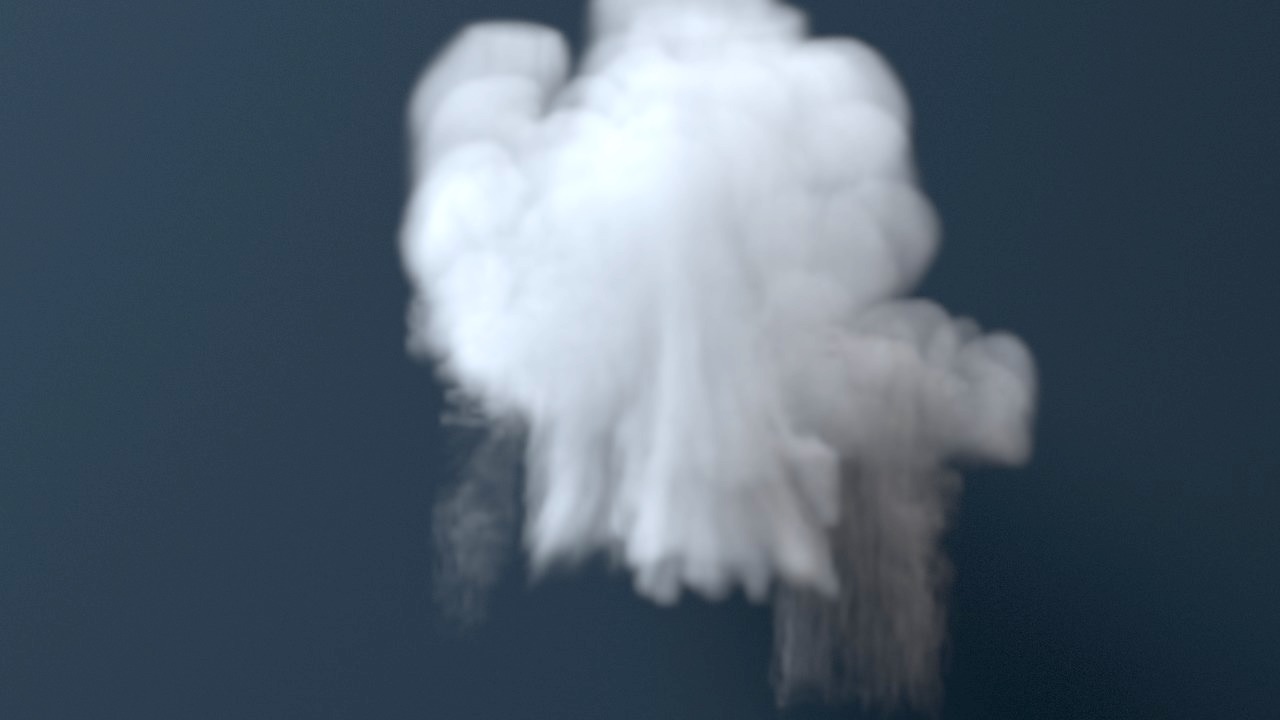} &
		\includegraphics[trim={224px, 0px, 224px, 0px}, clip, width = \myvv\columnwidth]{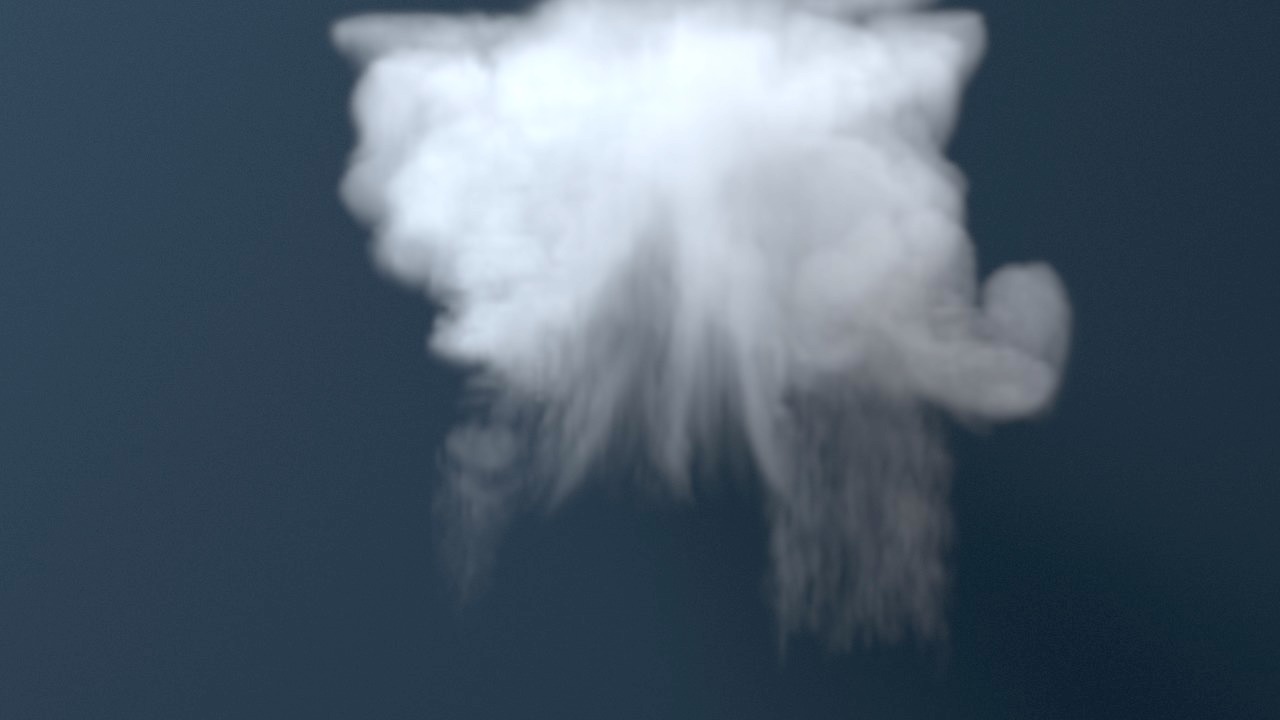} \\
		\includegraphics[trim={224px, 0px, 224px, 0px}, clip, width = \myvv\columnwidth]{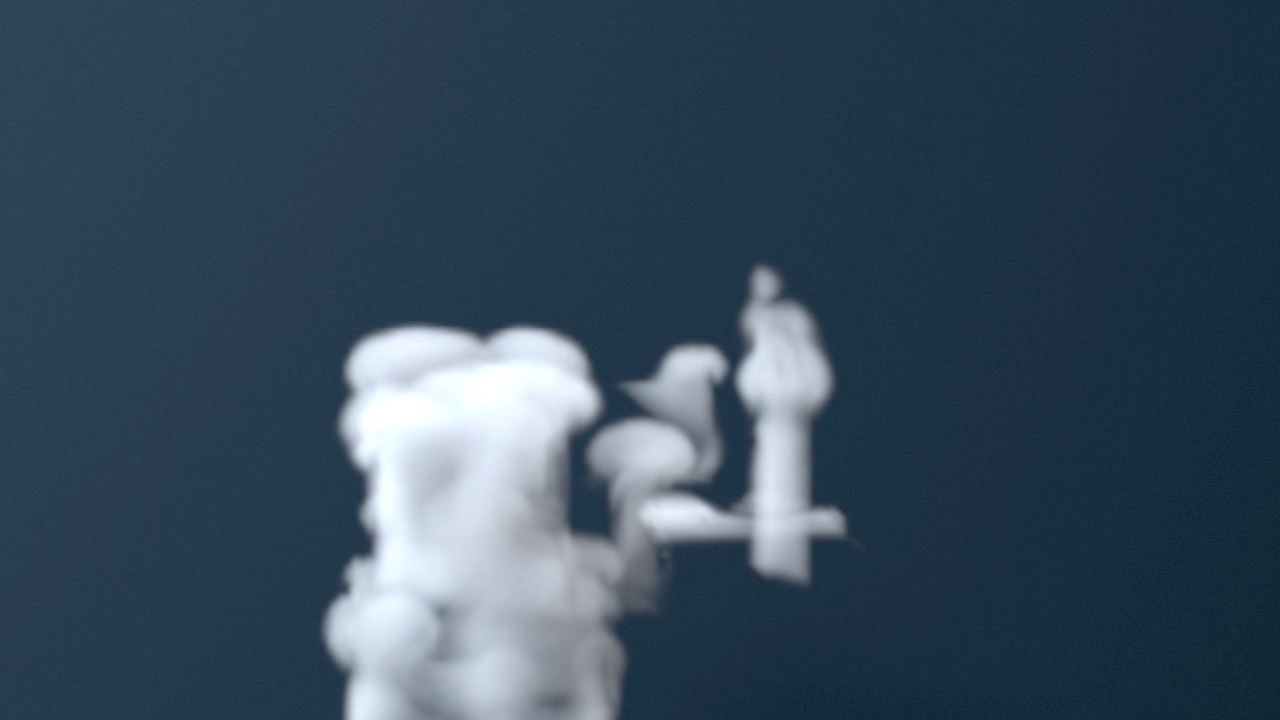} &
		\includegraphics[trim={224px, 0px, 224px, 0px}, clip, width = \myvv\columnwidth]{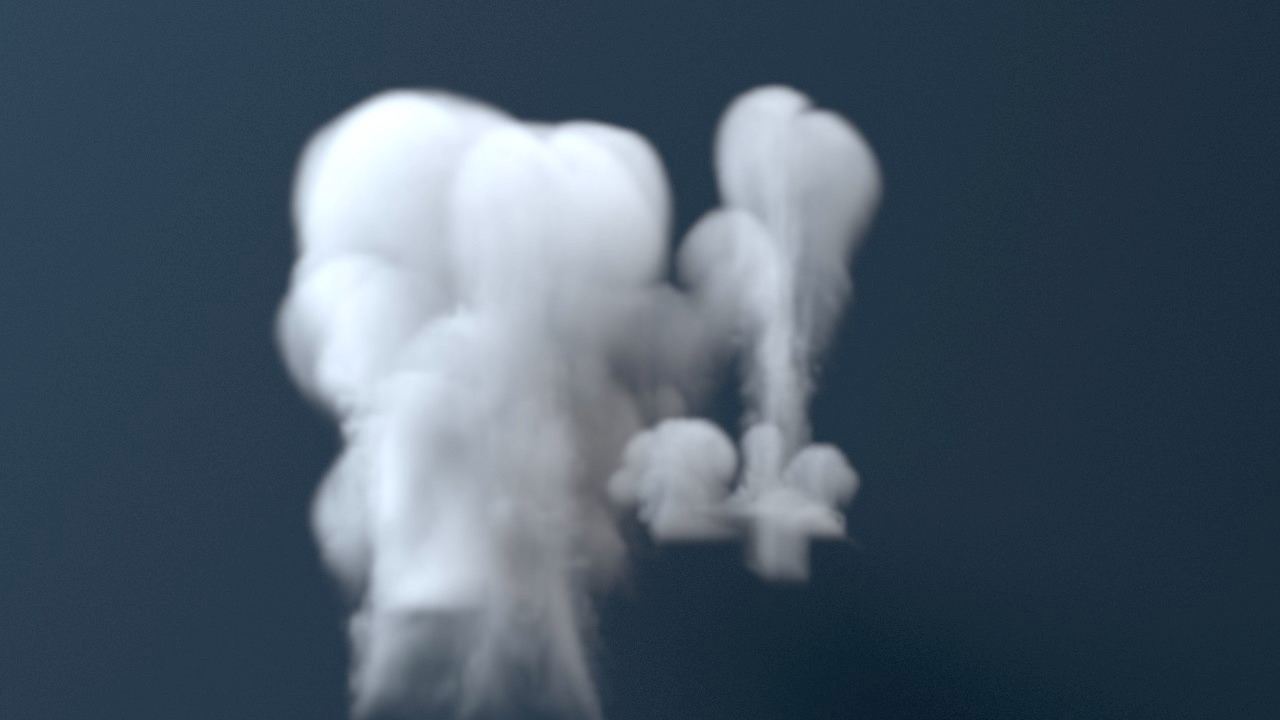} &
		\includegraphics[trim={224px, 0px, 224px, 0px}, clip, width = \myvv\columnwidth]{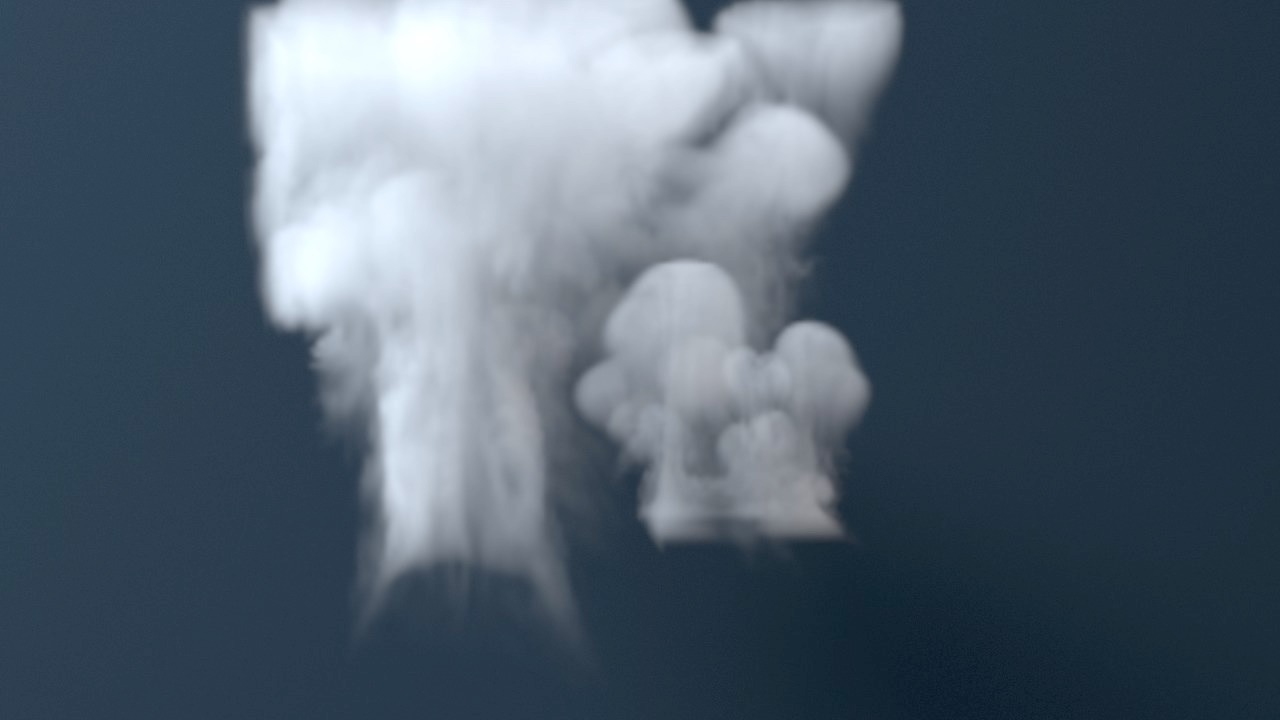} &
		\includegraphics[trim={224px, 0px, 224px, 0px}, clip, width = \myvv\columnwidth]{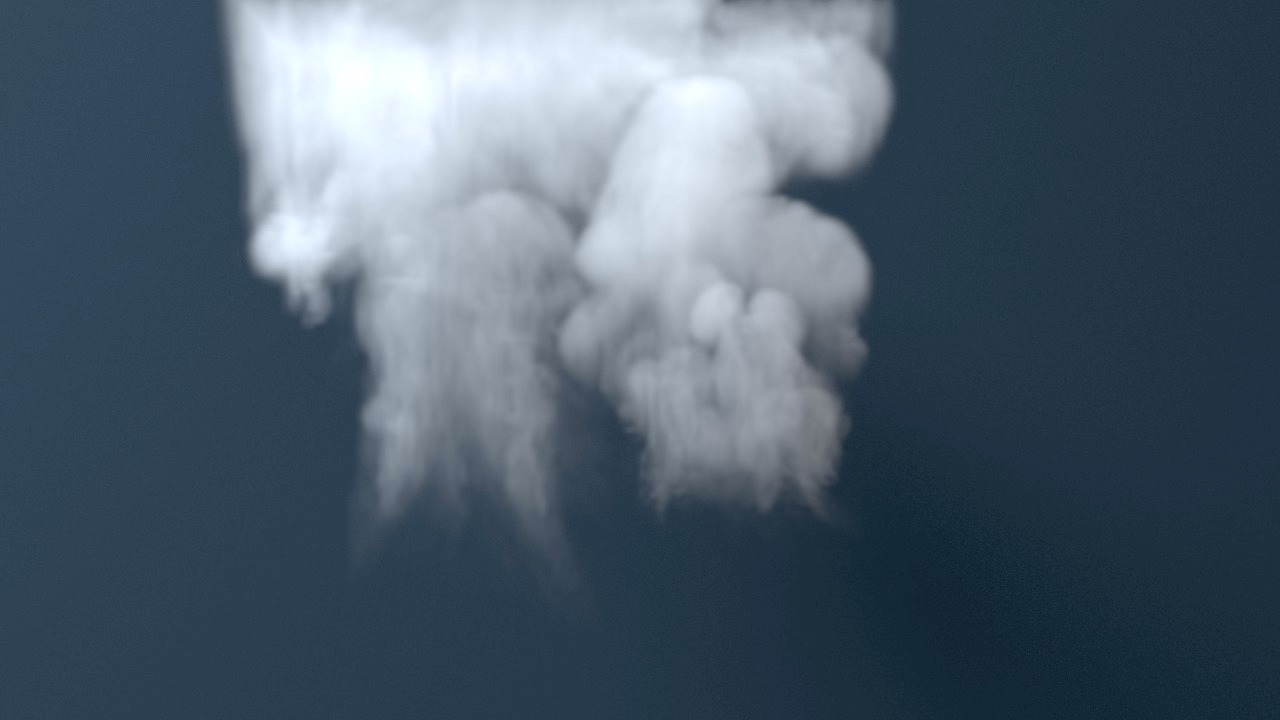} \\
		\subcaptionbox*{$t=0$  }{\includegraphics[trim={224px, 0px, 224px, 0px}, clip, width = \myvv\columnwidth]{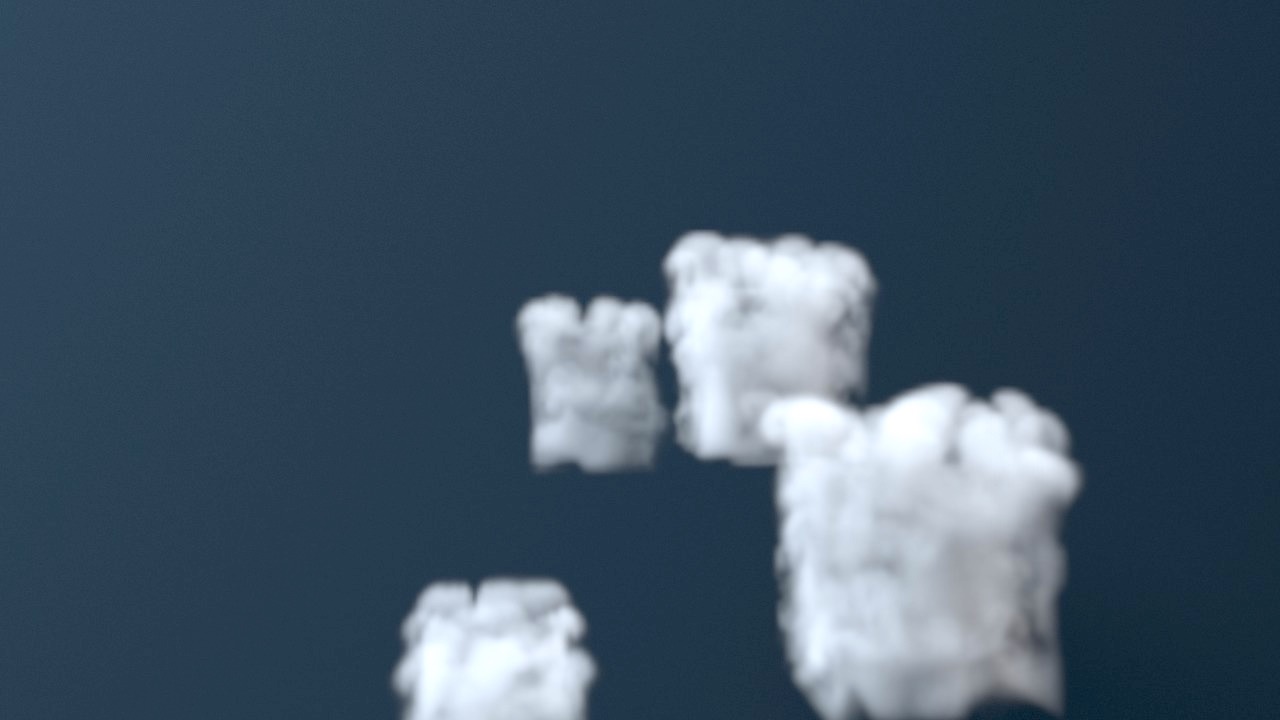}} &
		\subcaptionbox*{$t=33$}{\includegraphics[trim={224px, 0px, 224px, 0px}, clip, width = \myvv\columnwidth]{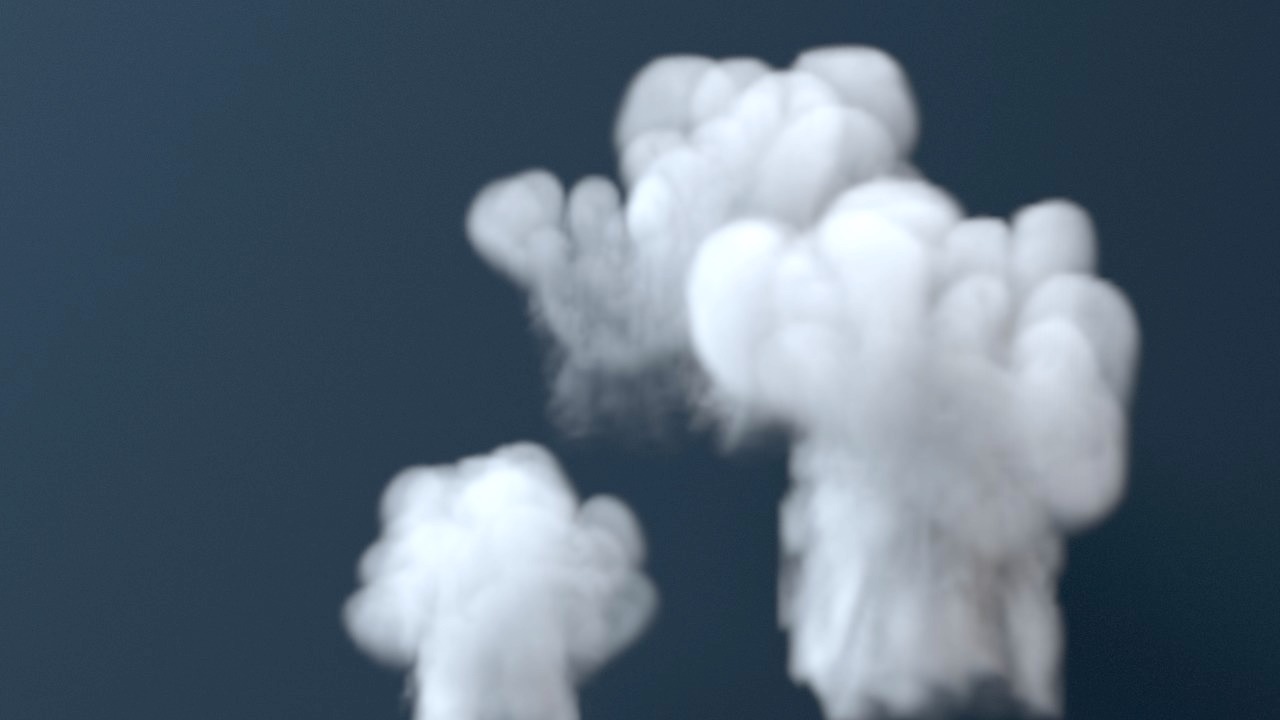}} &
		\subcaptionbox*{$t=66$}{\includegraphics[trim={224px, 0px, 224px, 0px}, clip, width = \myvv\columnwidth]{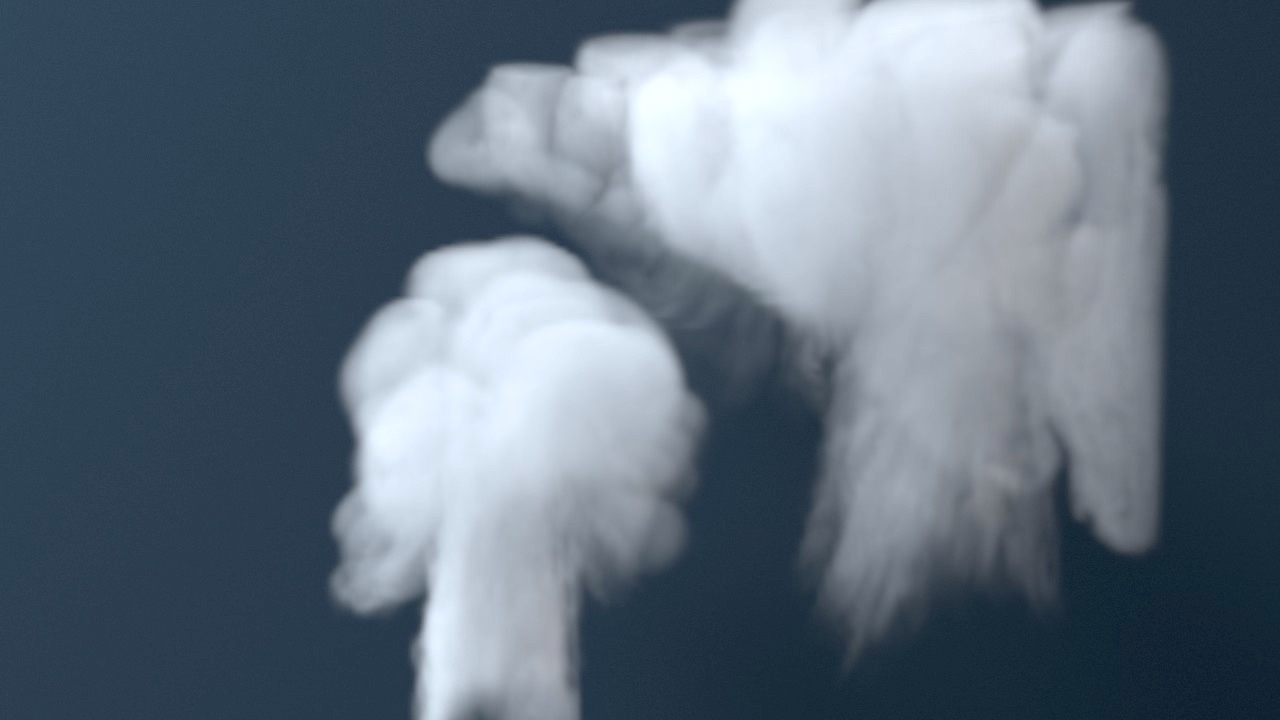}} &
		\subcaptionbox*{$t=99$}{\includegraphics[trim={224px, 0px, 224px, 0px}, clip, width = \myvv\columnwidth]{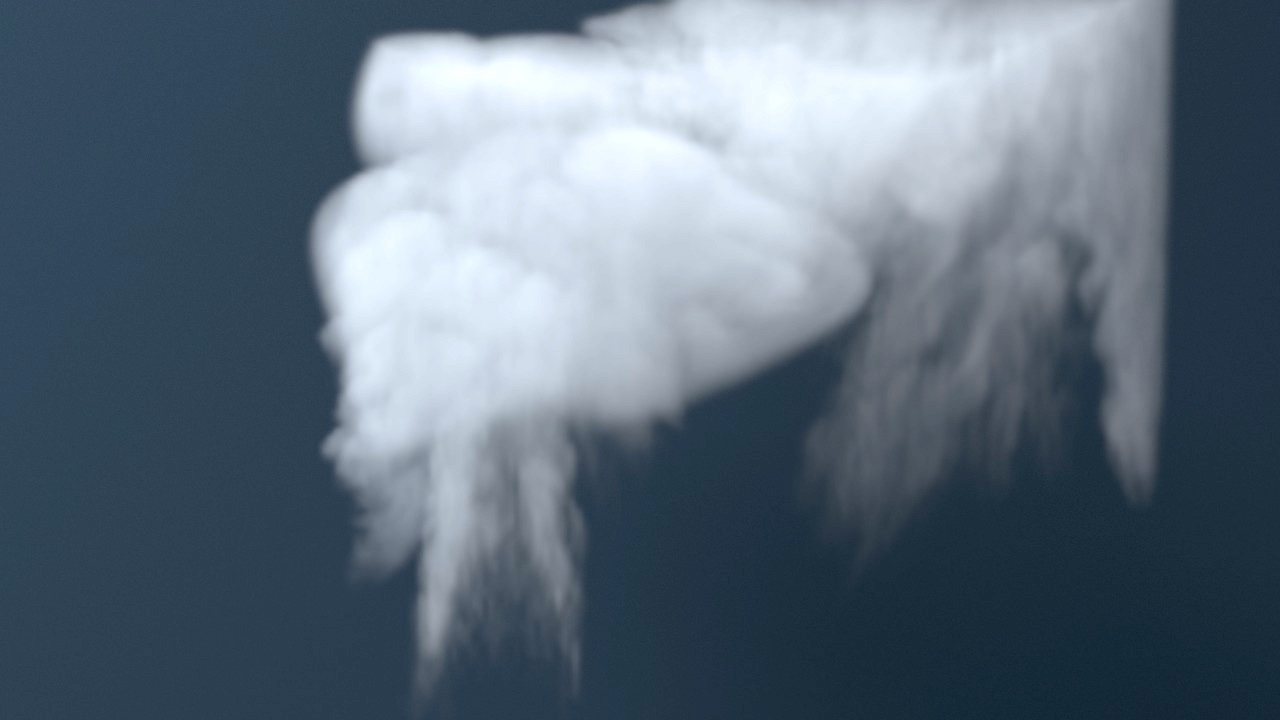}} \\
	\end{tabular}
	\caption{Several examples of $128^3$ smoke scenes predicted with an interval of $i_p=3$ by our LSTM network. }
	\label{fig:smoke_128}
\end{figure}

In \myreffig{fig:prediction_renderingsOld} additional time-steps of the comparison from \myreffig{fig:prediction_renderings} are shown. 
Here, different inferred simulation
quantities can be compared over the course of a simulation for different models.
In addition, \myreffig{fig:transparent_64},
\ref{fig:transparent_128}, and \ref{fig:smoke_128} show more realistic renderings of
our {\em liquid64}, {\em liquid128}, and {\em smoke128} models, respectively.

As our solve indirectly targets divergence, we also measured how well the predicted pressure
fields enforce divergence freeness over time. As a baseline, the numerical solver 
led to a residual divergence of $3.1\cdot10^{-3}$ on average. In contrast,
the pressure field predicted by our LSTM on average introduced a $2.1\cdot10^{-4}$ increase
of divergence per time step. Thus, the per
time step error is well below the accuracy of our reference solver, and especially
in combination with the interval predictions, we did not notice any significant changes
in mass conservation compared to the reference simulations. 

\begin{figure*}
	\vspace{+0.5cm}
	\centering \footnotesize
	{\renewcommand{\arraystretch}{0}
	\newcommand{\mwa}{0.49\textwidth}
	\begin{tabular}{@{}c@{~~~}c@{}c@{}}
\begin{overpic}[width = \mwa]{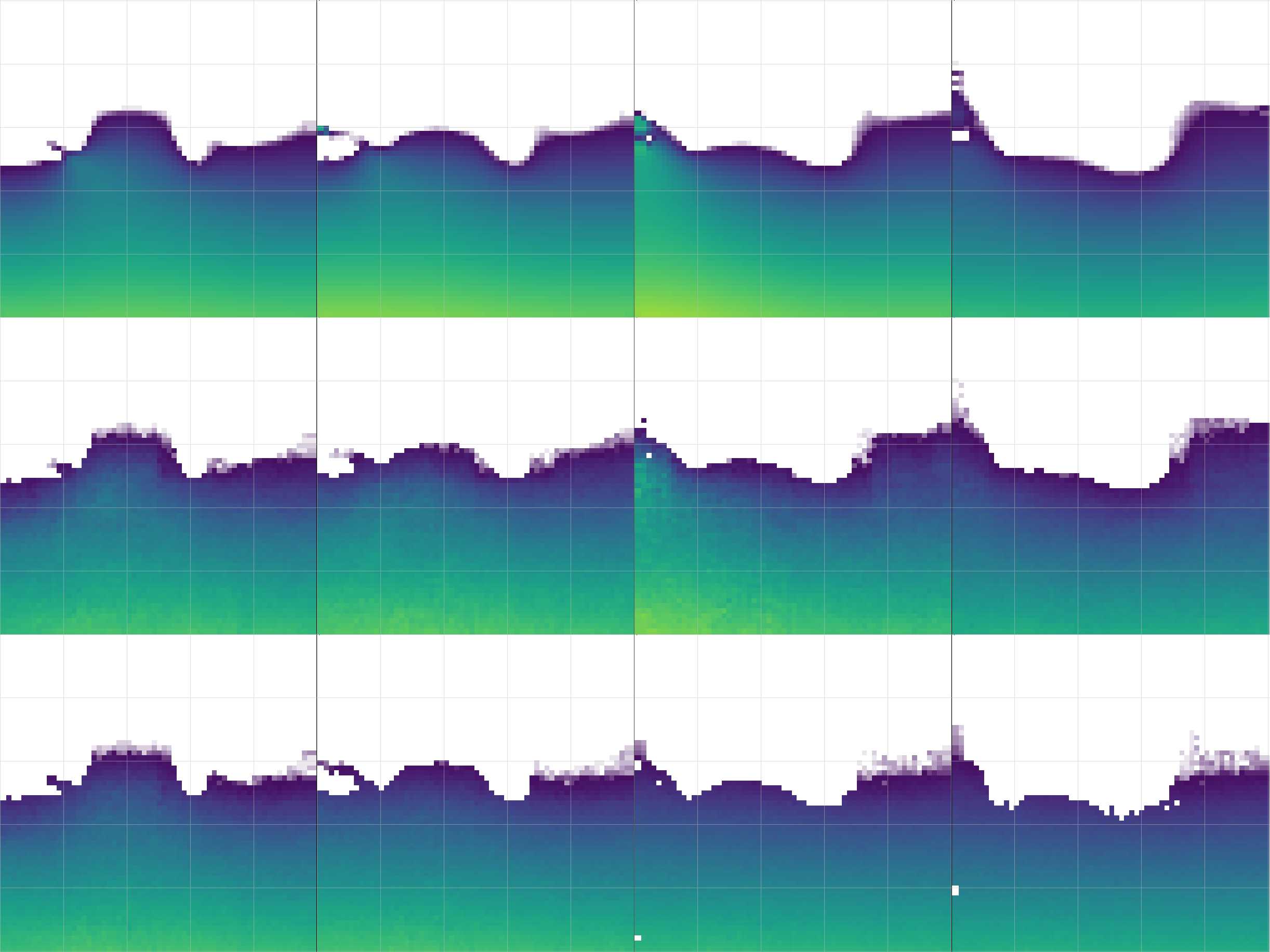}
	\put( 0, 77 ){\scriptsize \color{black}{a)}}
	\put( 1, 70 ){\scriptsize \color{gray}{GT}} 
	\put( 1, 45 ){\scriptsize \color{gray}{AE}} 
	\put( 1, 20 ){\scriptsize \color{gray}{LSTM}} 
\end{overpic}&
\begin{overpic}[width = \mwa]{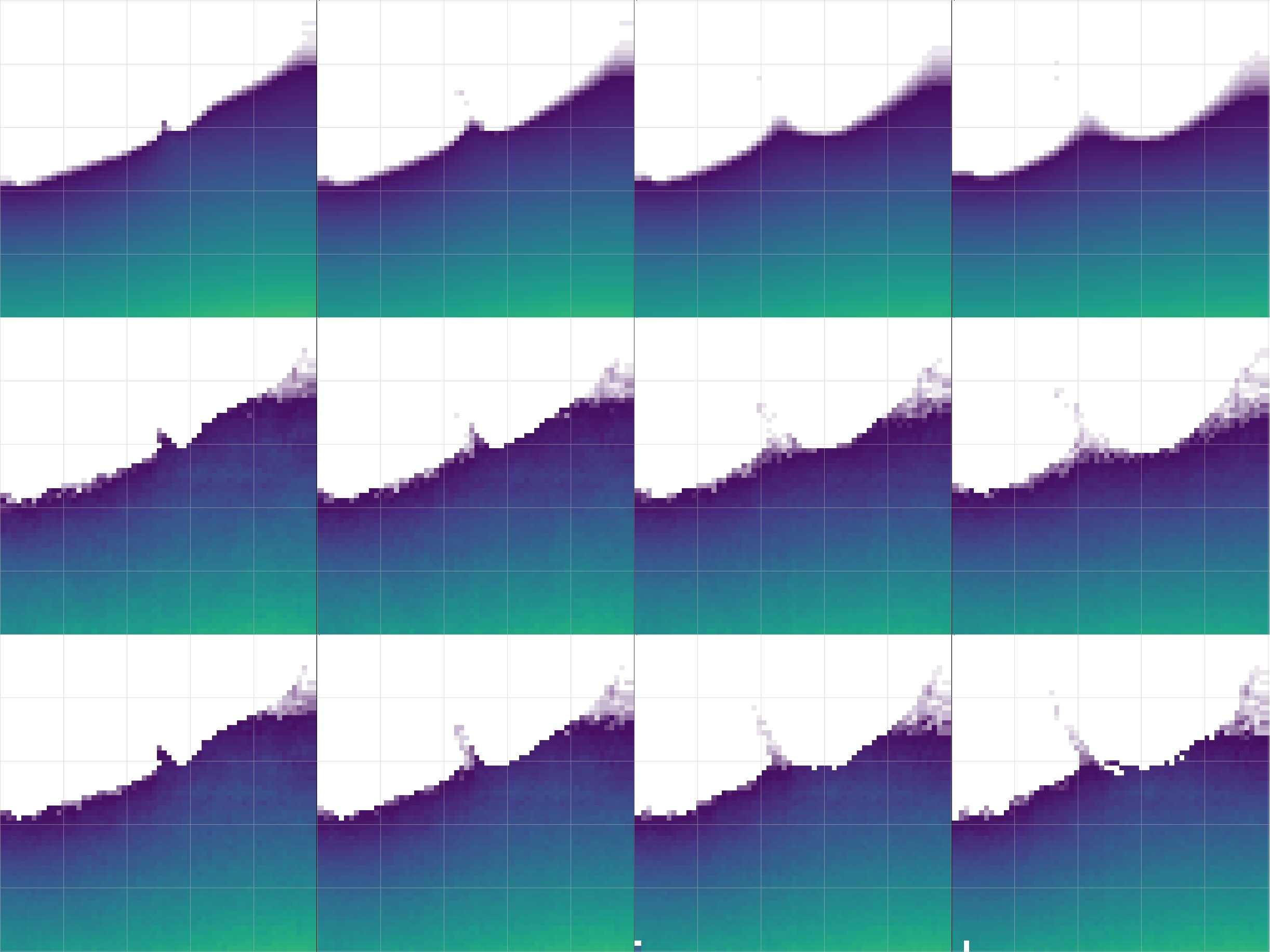}
	\put( 0, 77 ){\scriptsize \color{black}{b)}}
	\put( 1, 70 ){\scriptsize \color{gray}{GT}} 
	\put( 1, 45 ){\scriptsize \color{gray}{AE}} 
	\put( 1, 20 ){\scriptsize \color{gray}{LSTM}} 
\end{overpic}&
\\
&\vspace{0.5cm}&\\
\begin{overpic}[width = \mwa]{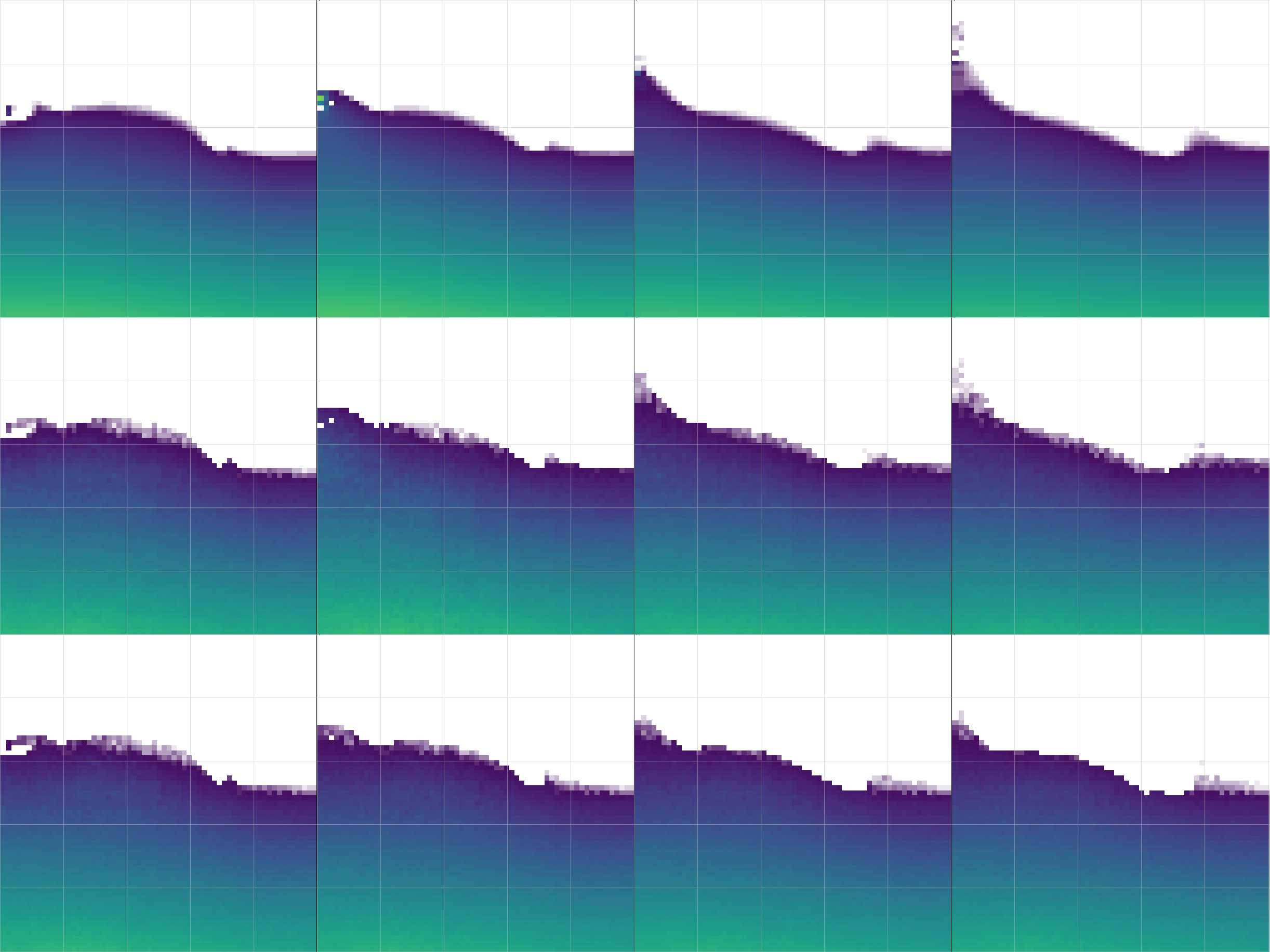}
	\put( 0, 77 ){\scriptsize \color{black}{c)}}
	\put( 1, 70 ){\scriptsize \color{gray}{GT}} 
	\put( 1, 45 ){\scriptsize \color{gray}{AE}} 
	\put( 1, 20 ){\scriptsize \color{gray}{LSTM}} 
\end{overpic}&
\begin{overpic}[width = \mwa]{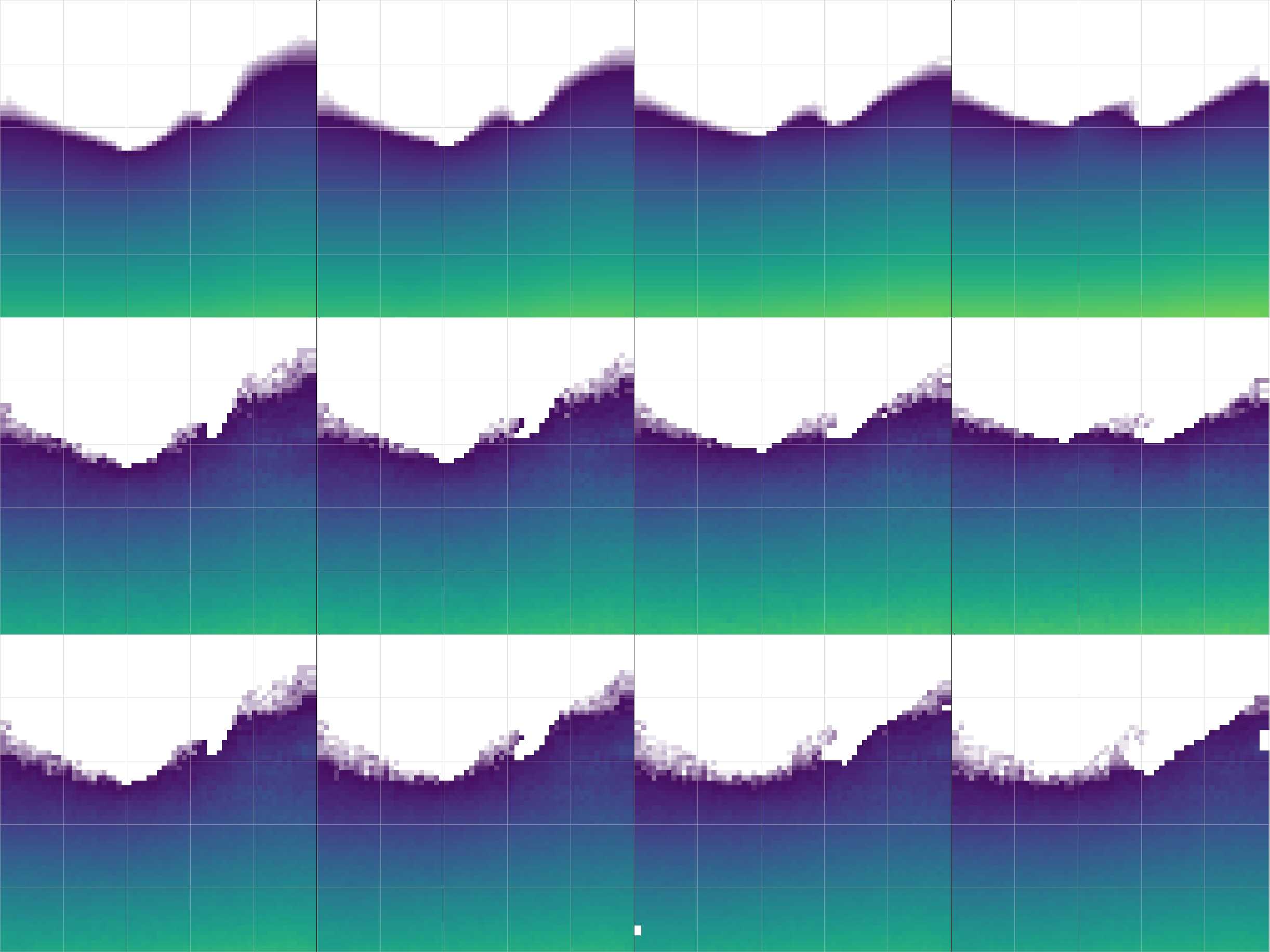}
	\put( 0, 77 ){\scriptsize \color{black}{d)}}
	\put( 1, 70 ){\scriptsize \color{gray}{GT}} 
	\put( 1, 45 ){\scriptsize \color{gray}{AE}} 
	\put( 1, 20 ){\scriptsize \color{gray}{LSTM}} 
\end{overpic}&
\\
&\vspace{0.5cm}&\\
\begin{overpic}[width = \mwa]{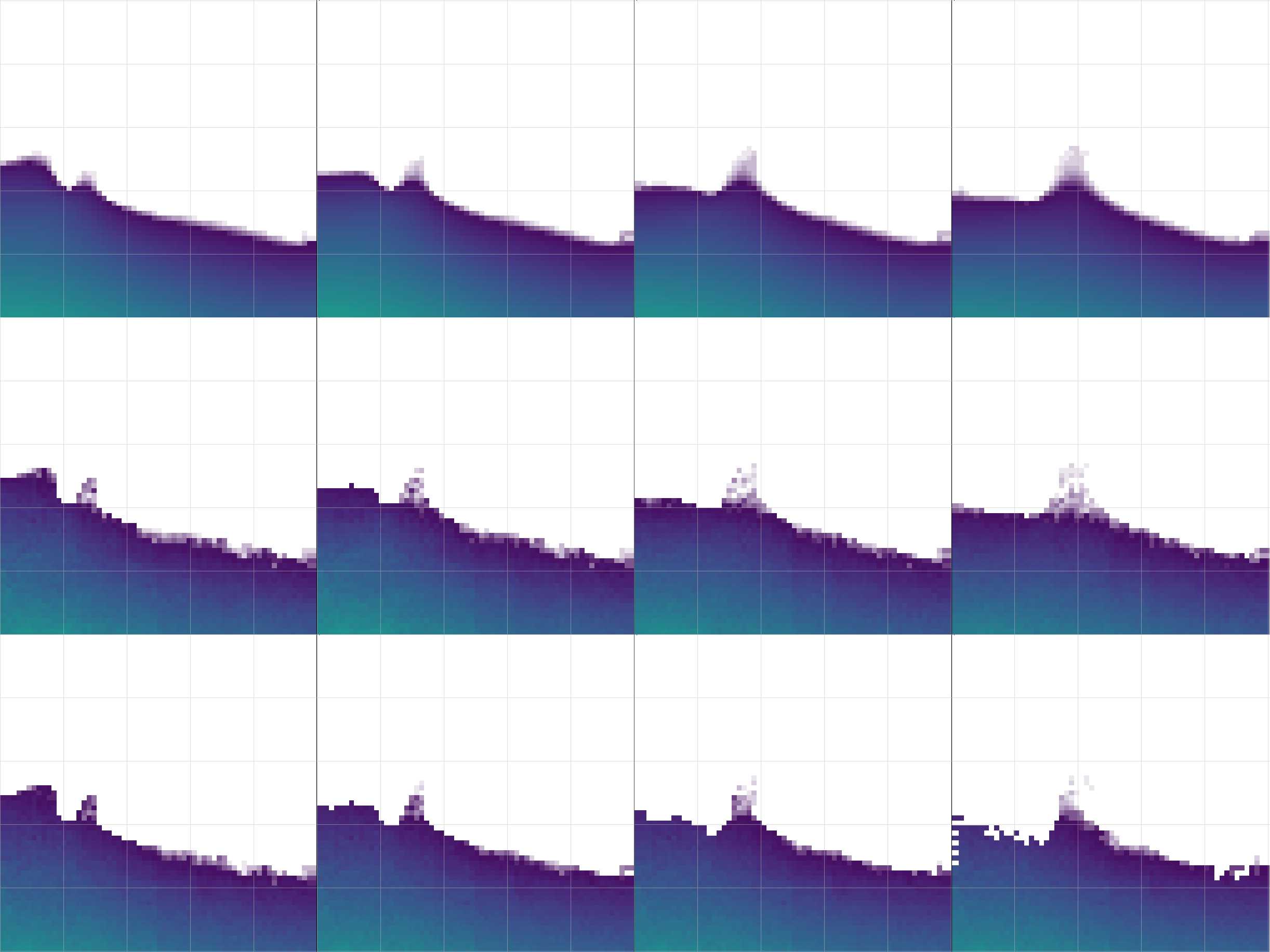}
	\put( 0, 77 ){\scriptsize \color{black}{e)}}
	\put( 1, 70 ){\scriptsize \color{gray}{GT}} 
	\put( 1, 45 ){\scriptsize \color{gray}{AE}} 
	\put( 1, 20 ){\scriptsize \color{gray}{LSTM}} 
\end{overpic}&
\begin{overpic}[width = \mwa]{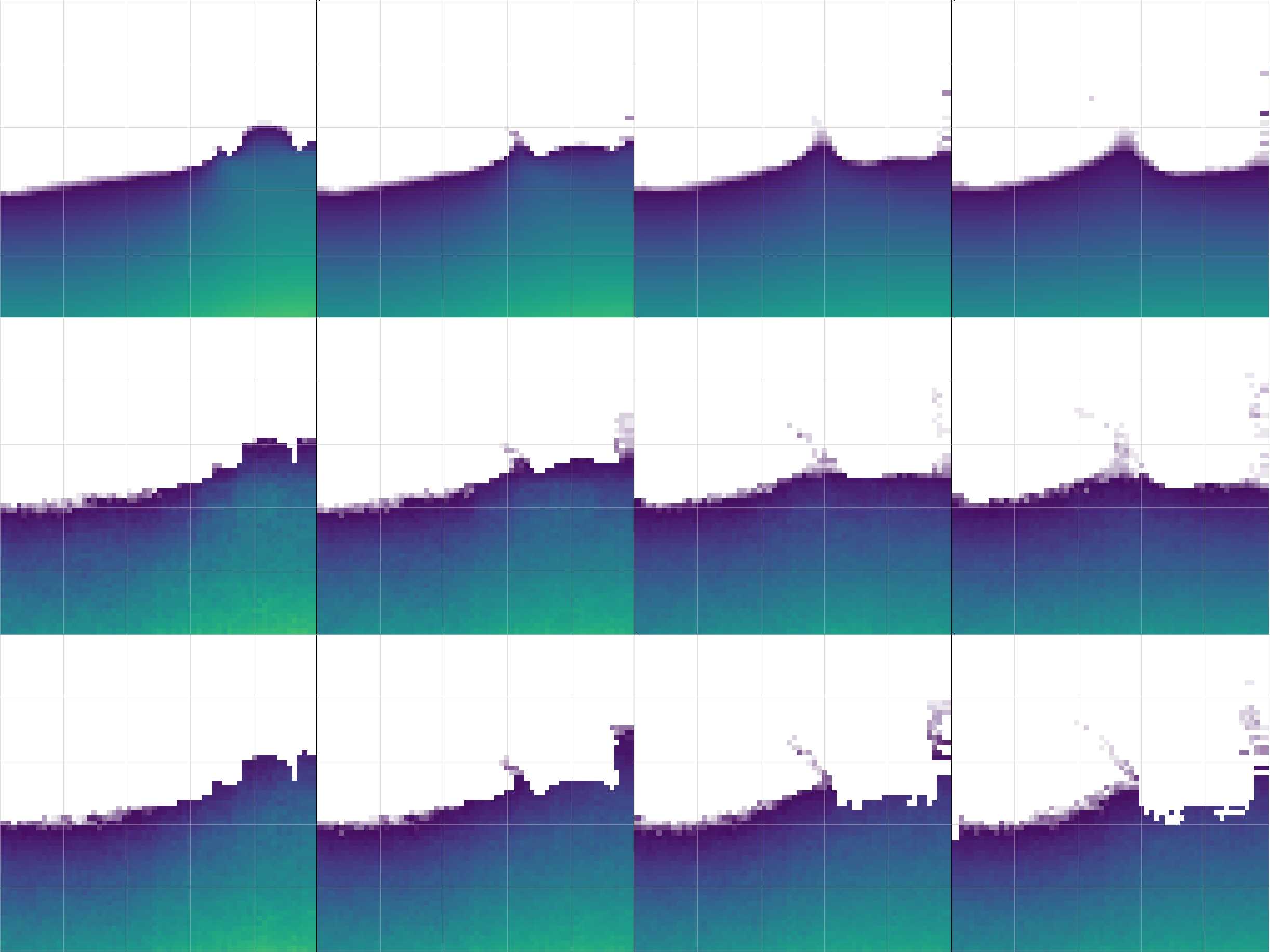}
	\put( 0, 77 ){\scriptsize \color{black}{f)}}
	\put( 1, 70 ){\scriptsize \color{gray}{GT}} 
	\put( 1, 45 ){\scriptsize \color{gray}{AE}} 
	\put( 1, 20 ){\scriptsize \color{gray}{LSTM}} 
\end{overpic}&
\\
	\end{tabular}
	}
	\caption{
	Six additional example sequences of ground truth pressure fields (top), the autoencoder baseline (middle),
	and the LSTM predictions (bottom). 
	All examples have resolutions of $64^2$, 
	and are shown over the course of a long horizon of 30 prediction steps with $i_p=\infty$. 
	The LSTM closely predicts the temporal evolution within the latent space.}
	\label{fig:data_fields_prediction}
\end{figure*}
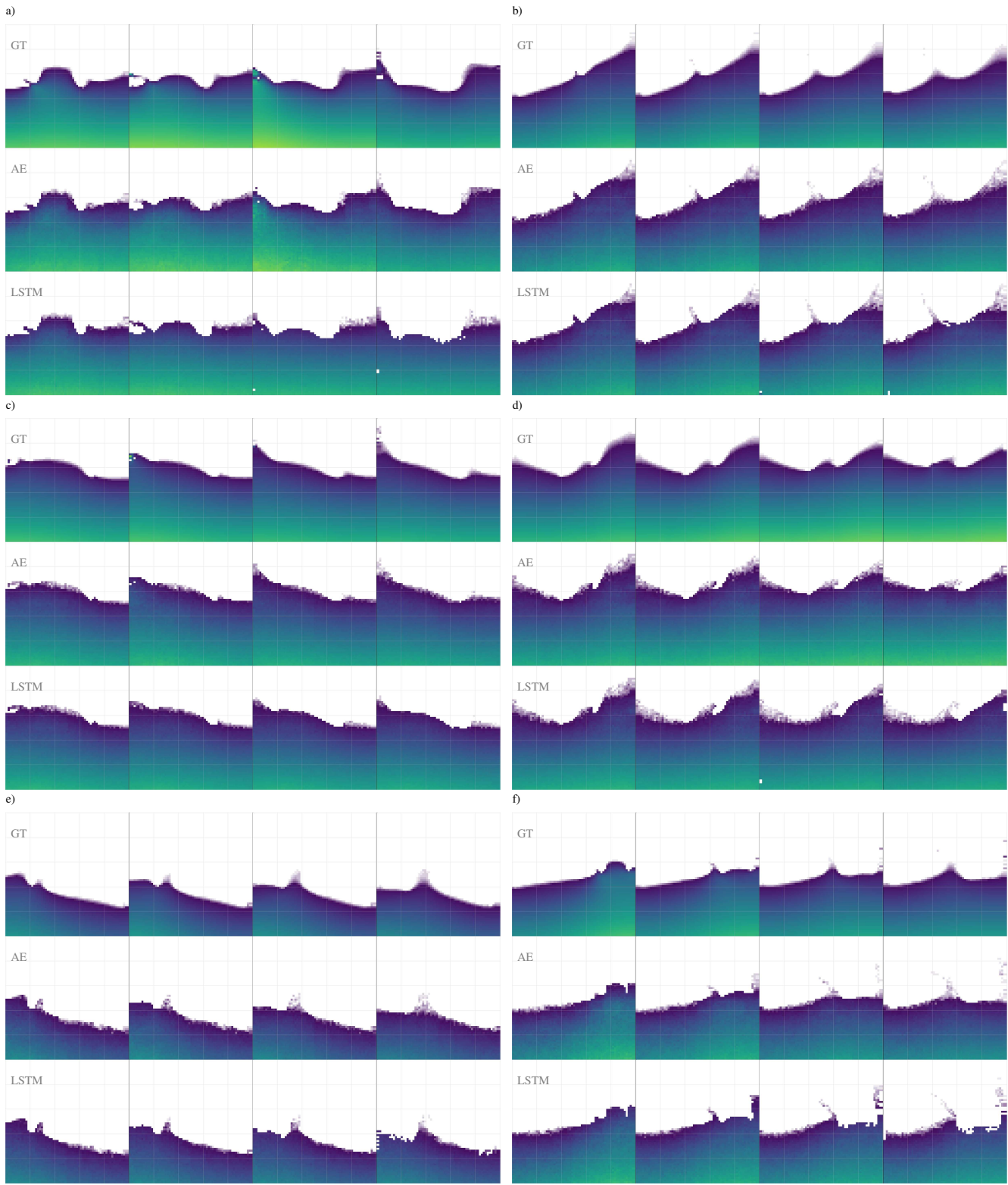

To visualize the temporal prediction capabilities as depicted in \myreffig{fig:data_fields_lstm_noj}, 
the spatial encoding task of the total pressure $p_t$ approach was reduced to two spatial dimensions.
For this purpose a 2D autoencoder network was trained on a dataset of resolution $64^2$.
The temporal prediction network was trained as described in the main document.
Additional sequences of the ground truth, the autoencoder baseline, and the temporal prediction by the LSTM network are shown in \myreffig{fig:data_fields_prediction}.

\section{Fluid Simulation Setup}
\label{app:fluid_sim_setup}

\ \\
\textit{FLIP Simulation}

\begin{figure}
	\centering
	\newcommand{\myw}{0.08} 	{\renewcommand{\arraystretch}{0}
	\begin{tabular}{@{}c@{}c@{}c@{}c@{}c@{}c@{}}
		\includegraphics[width = \myw\textwidth]{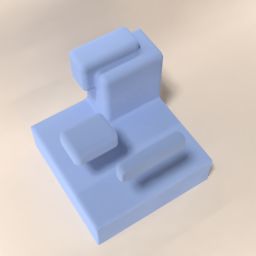} &
		\includegraphics[width = \myw\textwidth]{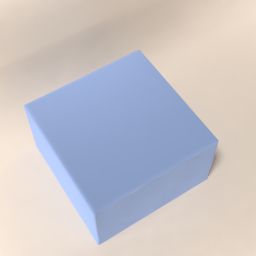} &
		\includegraphics[width = \myw\textwidth]{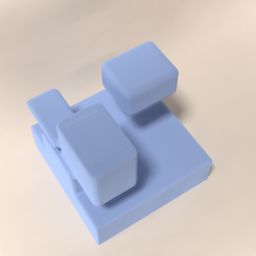} &
		\includegraphics[width = \myw\textwidth]{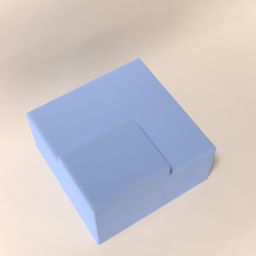} &
		\includegraphics[width = \myw\textwidth]{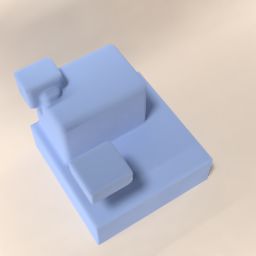} &
		\includegraphics[width = \myw\textwidth]{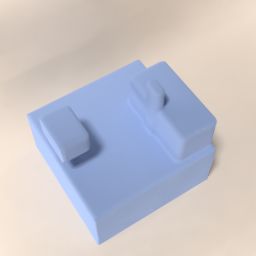} 
		\\
		\includegraphics[width = \myw\textwidth]{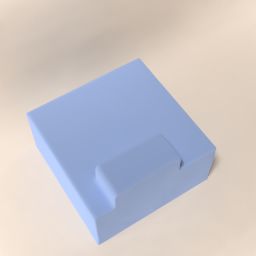} &
		\includegraphics[width = \myw\textwidth]{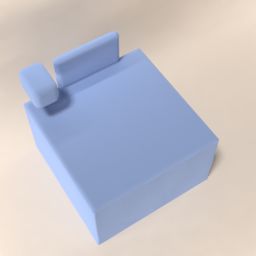} &
		\includegraphics[width = \myw\textwidth]{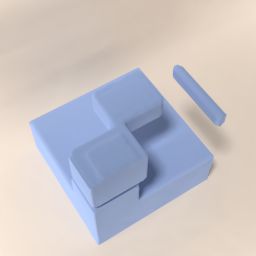} &
		\includegraphics[width = \myw\textwidth]{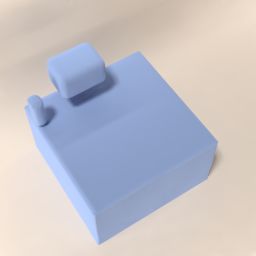} &
		\includegraphics[width = \myw\textwidth]{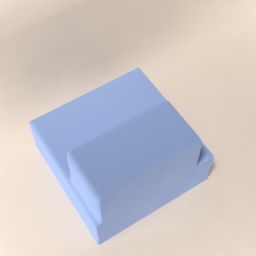} &
		\includegraphics[width = \myw\textwidth]{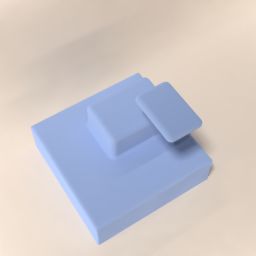}  
	\end{tabular}
	}
	\caption{Examples of initial scene states in the {\em liquid64} data set. }
	\label{fig:dataset_renderings}
\end{figure}

\begin{figure}
	\centering
	\includegraphics[width = 1.00\linewidth]{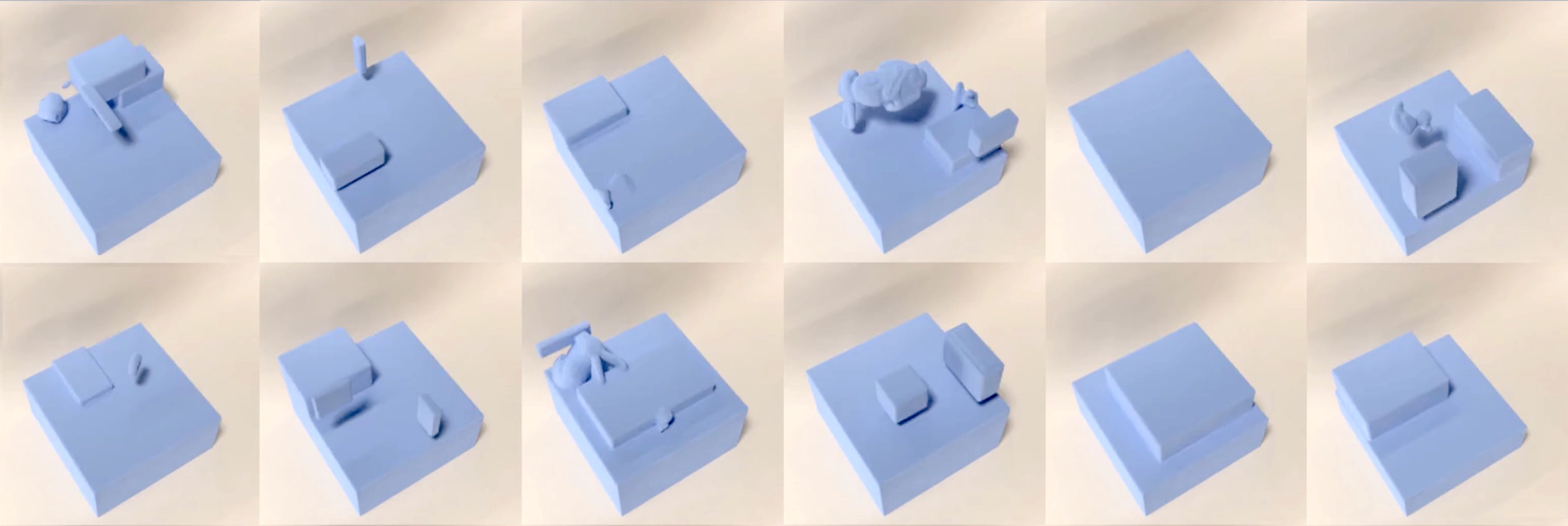} 
	\caption{Examples of initial scene states in the {\em liquid128} data set. The more complex initial shapes are visible in several of these configurations. }
	\label{fig:dataset_liq128initial}
\end{figure}

\begin{figure}
	\centering
	\includegraphics[width = 1.00\linewidth]{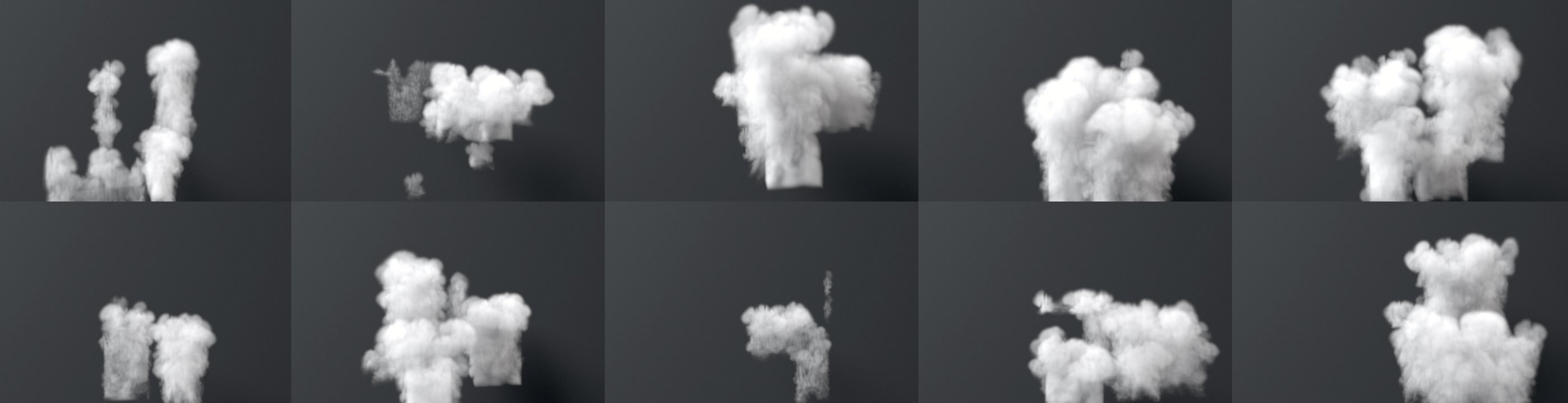} 
	\caption{Examples states of the {\em smoke128} training data set at $t=65$.}
	\label{fig:dataset_smoke}
\end{figure}

In addition to \myrefsec{sec:fluiddata}, we provide more information on the simulation setup in the following.
To generate our liquid datasets we use a classic NS solver~\cite{bridson2015}. The timestep is fixed to $0.1$, and pressure is computed with a conjugent gradient solver accuracy 
of $5\cdot10^{-5}$. The external forces in our setup only consist of a gravity vector of $(0.0, -0.01, 0.0)$ that is 
applied after every velocity advection step. No additional viscosity or surface tension forces are included.

In addition to the central quantities of a fluid solve, flow velocity $\vec{u}$, pressure $p$,
and potentially visible quantities such as the levelset $\phi$, we utilize 
the Fluid Implicit Particle (FLIP) \cite{Zhu2005} method, which represents a grid-particle hybrid.
It is used in this work on the one hand to generate the liquid datasets and on the other to be the base of our 
neural network driven interval prediction simulation.

To give a general overview of how the simulation proceeds, we shortly describe the computations executed for every time step in the following.
In each simulation step we first advect the FLIP particle system $PS$, the levelset $\phi$ and the velocity $\vec{u}$ itself with the current velocity grid.
Afterwards a second levelset containing the particle surface is created based on the current $PS$ configuration and is merged with $\phi$.
The merged levelset is extrapolated within a narrow band region of $3$, as described in the main text. 
After the levelset transformations $\vec{u}$ is updated with the $PS$ velocities and the 
external forces like gravity are applied on the result, followed by the enforcement of the static wall boundary conditions.
Next, a pressure field $p$ is computed via a Poisson solve using the divergence of $\vec{u}$ as right hand side.
After completing the pressure solve, the gradient of the result $\nabla p$ is subtracted from $\vec{u}$ yielding 
an approximation of a divergence free version of $\vec{u}$.
The $PS$ velocities are updated based on the difference between post-advection version of $\vec{u}$ and the latest divergence free one.

\ \\
\textit{Prediction Integration}

The presented LSTM prediction framework supports predictions of different simulation fields from the FLIP simulation presented above.
The supported fields are the final $\vec{u}$ at the end of the simulation loop, the solved pressure $p$ and the decomposed version of $p$ with $p_s$ and $p_d$, i.e. the hydrostatic and dynamic components of the regular pressure field, respectively.
For the prediction of these fields we supply multiple architectures that are compared in the main text. Those are the \emph{total pressure}, \emph{variational split pressure}, \emph{split pressure} and \emph{velocity} versions.
The difference between the variational split pressure and the default split pressure approach is the architecture of the autoencoder, whereas the temporal prediction network stays the same. 

Depending on the prediction architecture, the inferred, decoded predicted field is used instead of executing the 
corresponding numerical approximation step.
When targeting $\vec{u}$ with our method, this means that we can omit velocity advection as well as pressure solve,
while the inference of $p$ by the split or total pressure architecture means that we only omit the pressure solve,
but still need to perform advection and velocity correction with the pressure gradient.
While the latter requires more computations, 
the pressure solve is typically the most time consuming part, with a super-linear complexity, and as such
both options, using either $\vec{u}$ or the $p$ variants, have comparable runtimes.

\section{Hyperparameters}
\label{app:hyperparam}

To find appropriate hyperparameters for the prediction network,
a large number of training runs with varying parameters were executed on a subset of the total training data domain.
The subset consisted of $100$ scenes of the training data set discussed in \myrefsec{sec:data_gen}.

\begin{figure} [htb!]
	\centering
	\renewcommand{\arraystretch}{0}
	\begin{tabular}{@{}c c@{}}
		\subcaptionbox{kernel regularizer and recurrent regularizer\label{fig:hyperparamreglstm}}{\includegraphics[width = 0.49\columnwidth]{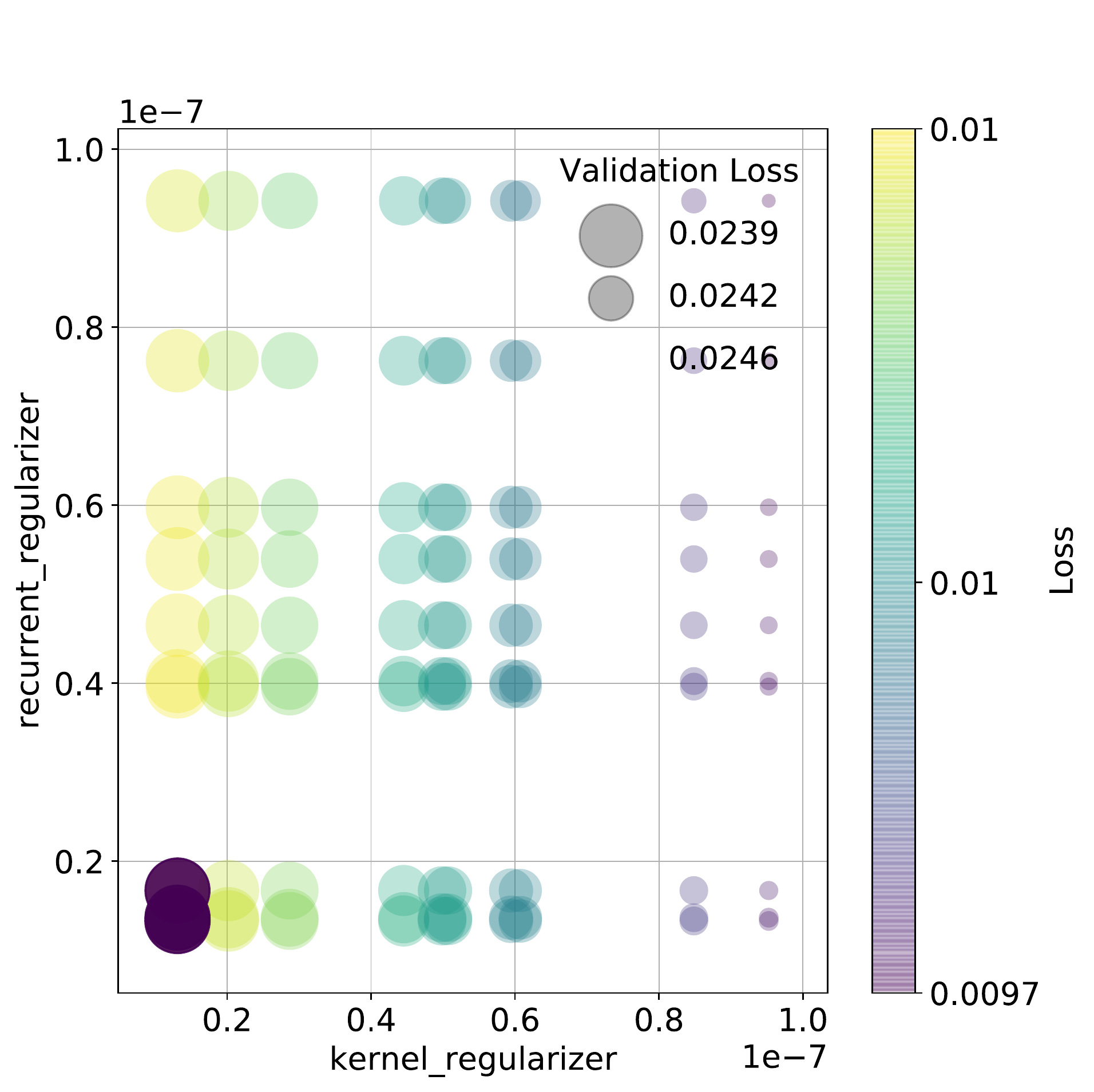}} &
		\subcaptionbox{learning rate and learning rate decay\label{fig:hyperparamlrdecaylstm}} {\includegraphics[width = 0.49\columnwidth]{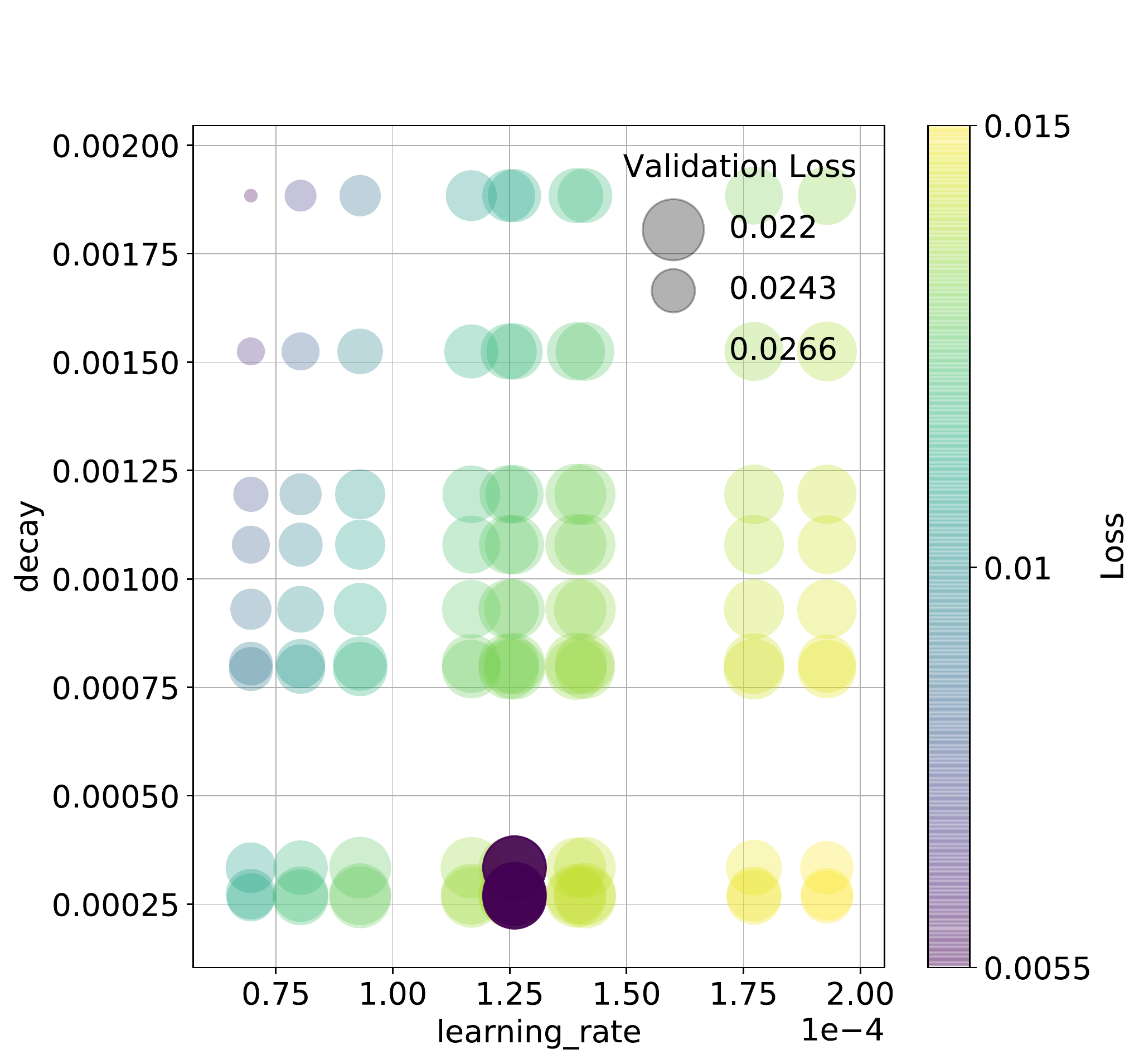}} \\
		\subcaptionbox{learning rate and dropout\label{fig:hyperparamlrdroplstm}}{\includegraphics[width = 0.49\columnwidth]{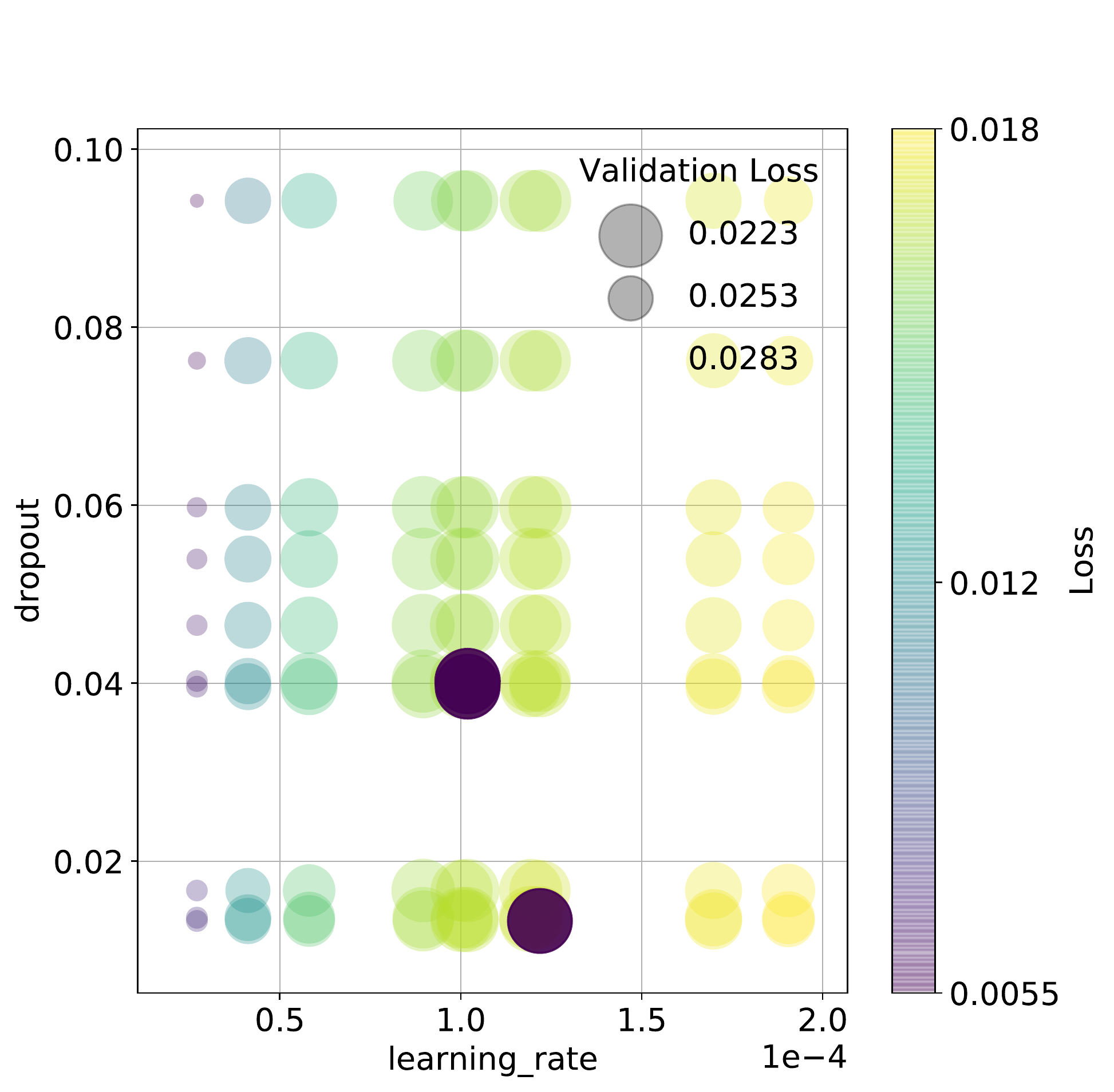}} &
		\subcaptionbox{recurrent dropout and dropout\label{fig:hyperparamrecdropdroplstm}}{\includegraphics[width = 0.49\columnwidth]{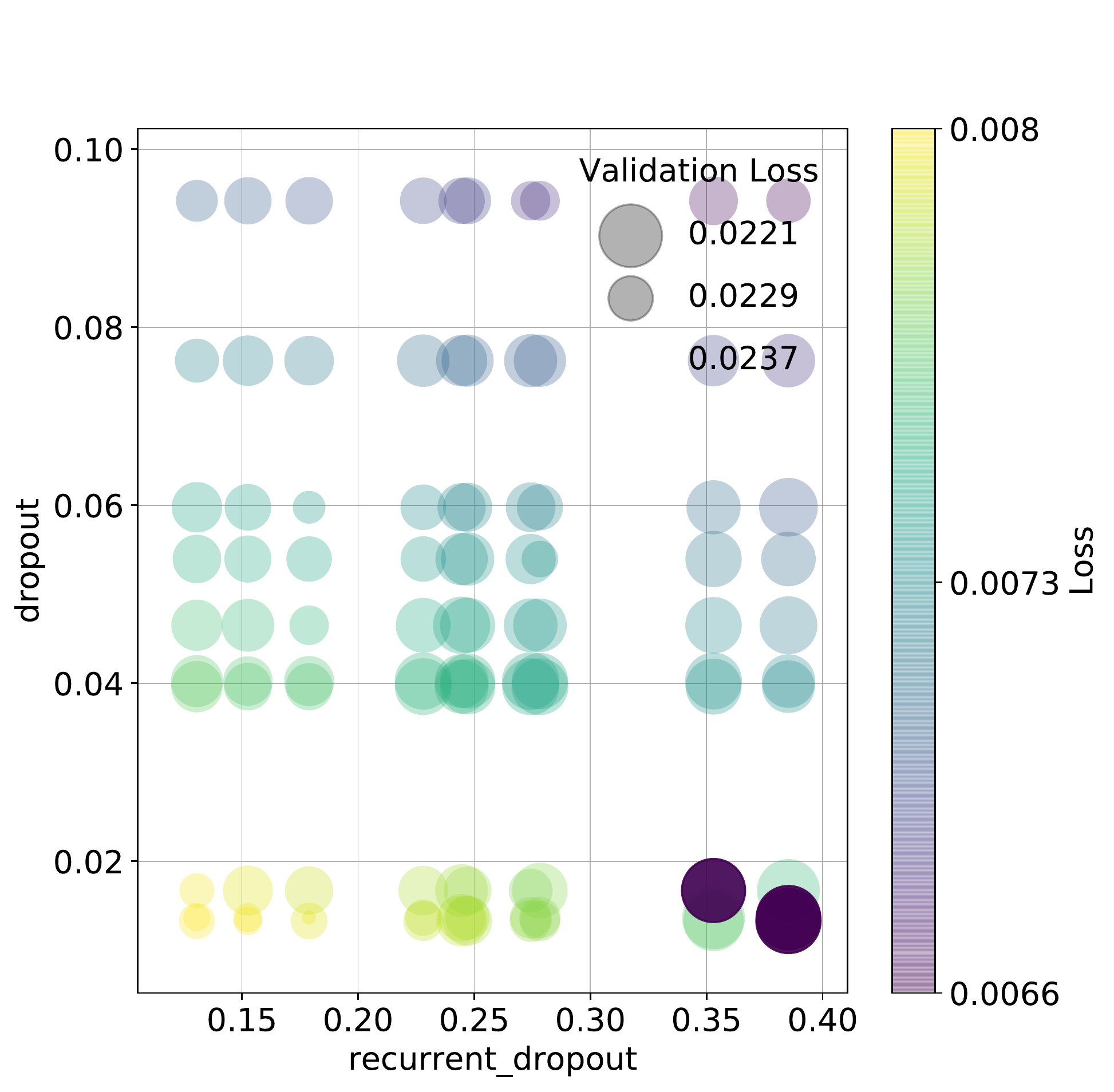}} \\
	\end{tabular}
	\caption{Random search of hyperparameters for the prediction network}
	\label{fig:hyperparam}
\end{figure}

In \myreffig{fig:hyperparam} (a-d),
examples for those searches are shown. Each circle shown in the graphs represents the final result of one complete training run with the parameters given on
the axes. The color represents the mean absolute error of the training error, ranging from purple (the best) to yellow (the worst). The size of the circle 
corresponds to the validation error, i.e., the most important quantity we are interested in. The best two runs are highlighted with a dark coloring.
These searches yield interesting results, e.g. \myreffig{fig:hyperparamreglstm} shows that the network performed best without any weight decay regularization applied.

\begin{figure} [htb!]
	\centering
		\includegraphics[width=0.22\textwidth]{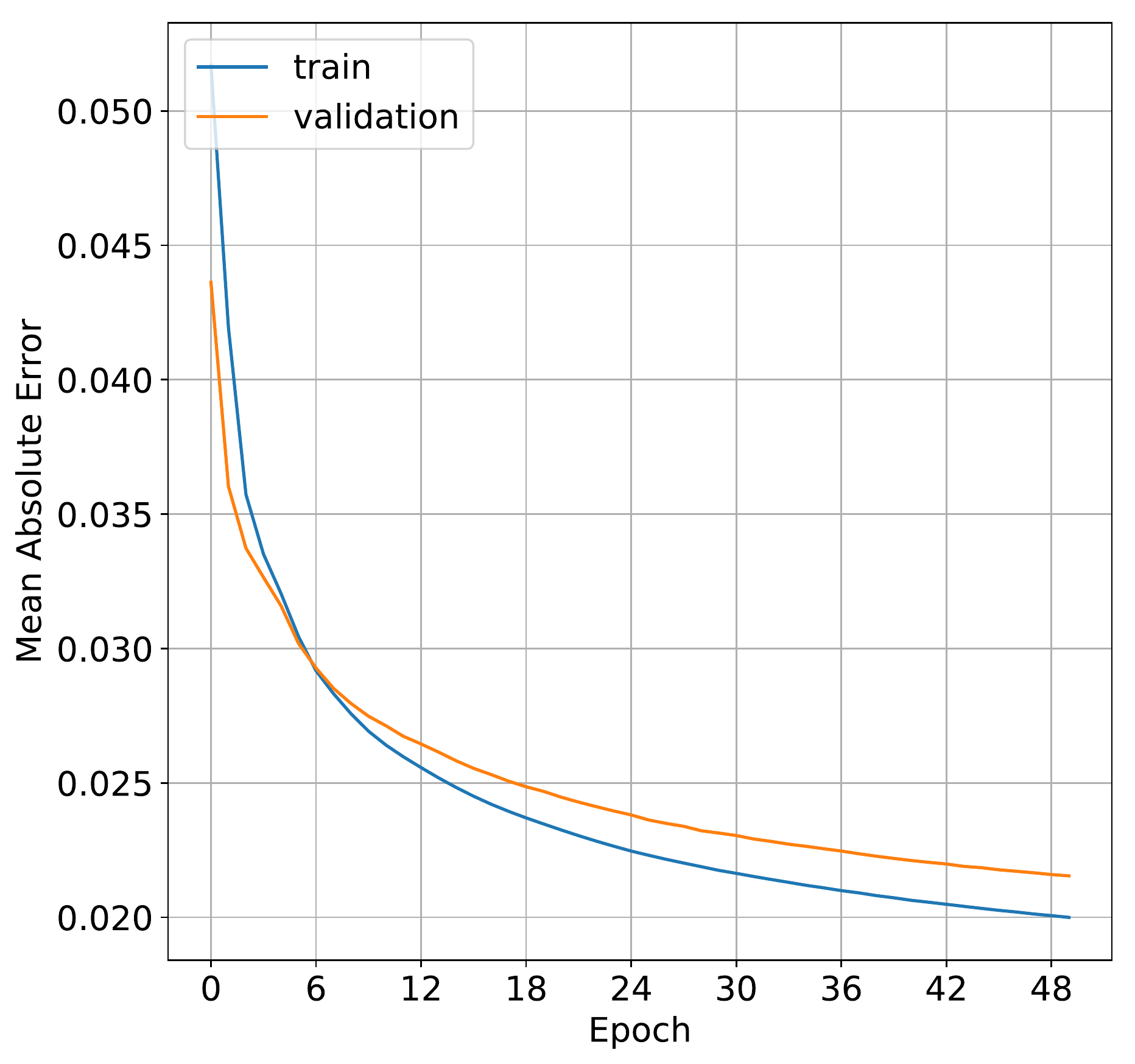}
		\caption{ Training history of the {\em liquid128} $p_t$ network. }
		\label{fig:lstmtrainhist}
\end{figure}

Choosing good parameters leads to robust learning behavior in the training process, an example is shown in \myreffig{fig:lstmtrainhist}.
Note that it is possible for the validation error to be lower than the training error as dropout is not used for computing the validation loss.
The mean absolute error of the prediction on a test set of $40$ scenes, which was generated independently from the training and validation 
data, was $0.0201$ for this case. These results suggest that the network generalizes well with the given hyperparameters.

\end{document}